\pdfoutput=1
\documentclass[10pt]{jcimfastformat}
\usepackage[utf8]{inputenc}
\usepackage[T1]{fontenc}
\graphicspath{{figures/}}

\usepackage{subfig}
\usepackage{amsmath}
\usepackage{siunitx}
\usepackage{multirow}
\usepackage{colortbl}
\usepackage{bm}
\usepackage{eucal}
\usepackage{algorithm2e}
\usepackage{tikz}
\usepackage{arydshln}
\usepackage{placeins}
\usepackage{caption}
\usepackage[percent]{overpic}

\title{Classical structural priors improve neural molecular representations in molecule language models}

\author[1]{Donghyeok Lee}
\author[2]{Sangkwon Park}
\author[3,4,+]{Munyoung Chang}
\author[1,2,+]{Sungroh Yoon}

\affil[1]{Interdisciplinary Program in Artificial Intelligence, Seoul National University, Seoul 08826, Republic of Korea}
\affil[2]{Department of Electrical and Computer Engineering, Seoul National University, Seoul 08826, Republic of Korea}
\affil[3]{Education and Research Program for Future ICT Pioneers, Seoul National University, Seoul 08826, Republic of Korea}
\affil[4]{Department of Otorhinolaryngology-Head \& Neck Surgery, Seoul National University Bundang Hospital, Seongnam 13620, Republic of Korea}

\affil[+]{These authors contributed equally as corresponding authors}

\affil[*]{Correspondence: caduent@gmail.com (M.C.); sryoon@snu.ac.kr (S.Y.)}

\newcommand{\methodname}{SAGA}
\newcommand{\lossname}{SAGA Regularization}

\begin{abstract}
Classical molecular similarity is central to chemical information systems, yet molecule-language models trained on paired structure-text data do not necessarily organize molecular embeddings according to graded structural relationships. Here we introduce Structure-Aware Guided Alignment (SAGA), a training-only regularizer that uses Centered Kernel Alignment to align neural molecular representations with fingerprint-based Tanimoto similarity, without adding parameters or requiring fingerprints at inference. In MolCA and Atomas, SAGA improves molecule--text retrieval, shifts non-exact retrieval toward closer structural analogues, and increases average scaffold-split property-prediction performance, while molecular description and text-to-molecule generation are improved or preserved in single-run ablations. These results show that established chemoinformatics priors can make molecule-language representations more chemically faithful while retaining their multimodal utility.
\end{abstract}
\begin{document}

\raggedbottom
\maketitle
\thispagestyle{empty}

Molecular similarity is a central organizing principle in chemoinformatics and molecular modeling. Compounds that share substructures often exhibit related properties, whereas small structural changes can also alter or abolish activity~\cite{willett1998similarity,vantilborg2022cliffs}. Fingerprints and Tanimoto similarity provide a compact way to encode these relationships and have long supported chemical database searching, analogue retrieval and structure--property analysis~\cite{rogers2010ecfp,bajusz2015tanimoto}. Neural molecular representation learning has expanded this toolkit by learning embeddings for prediction, retrieval and generation~\cite{david2020molrep,atz2021geometric}. In molecule-language models, such embeddings connect chemical structures with natural-language descriptions, enabling molecule--text retrieval~\cite{edwards2021text2mol}, annotation, property prediction~\cite{yang2019chemprop} and molecular generation.

Recent multimodal pre-training frameworks bridge molecular structures and text through paired representation learning~\cite{liu2023moleculestm,zeng2022bridging,edwards2021text2mol}, including MoMu~\cite{su2022momu}, MolCA~\cite{liu2023molca}, Atomas~\cite{atomas2024} and knowledge-guided pre-training methods~\cite{li2023kpgt}. These models are typically trained to match paired molecules and descriptions, or to use auxiliary signals that improve representation quality. However, chemistry is not organized only by exact matches. Binary contrastive objectives~\cite{you2020graphcl,wang2022molclr} treat a structurally related nonmatch and an unrelated molecule in the same negative class, leaving underused the continuum of molecular similarity between exact identity and unrelated compounds.

Classical chemoinformatics provides a direct way to describe this continuum, but incorporating it into neural molecule-language models is not simply a matter of adding descriptor features. Fingerprints and learned embeddings differ in dimensionality, scale and coordinate system, and their value lies in pairwise molecular relationships rather than in one-to-one coordinate correspondence. A natural target is therefore relational alignment: preserving the similarity structure encoded by fingerprints in the neural molecular embedding space. Centered Kernel Alignment (CKA)~\cite{kornblith2019cka} provides a dimension-agnostic measure of similarity between representation spaces, making it possible to compare fingerprint-derived structural relationships with learned molecular representations.

\begin{figure}[ht]
\centering
\includegraphics[width=\textwidth]{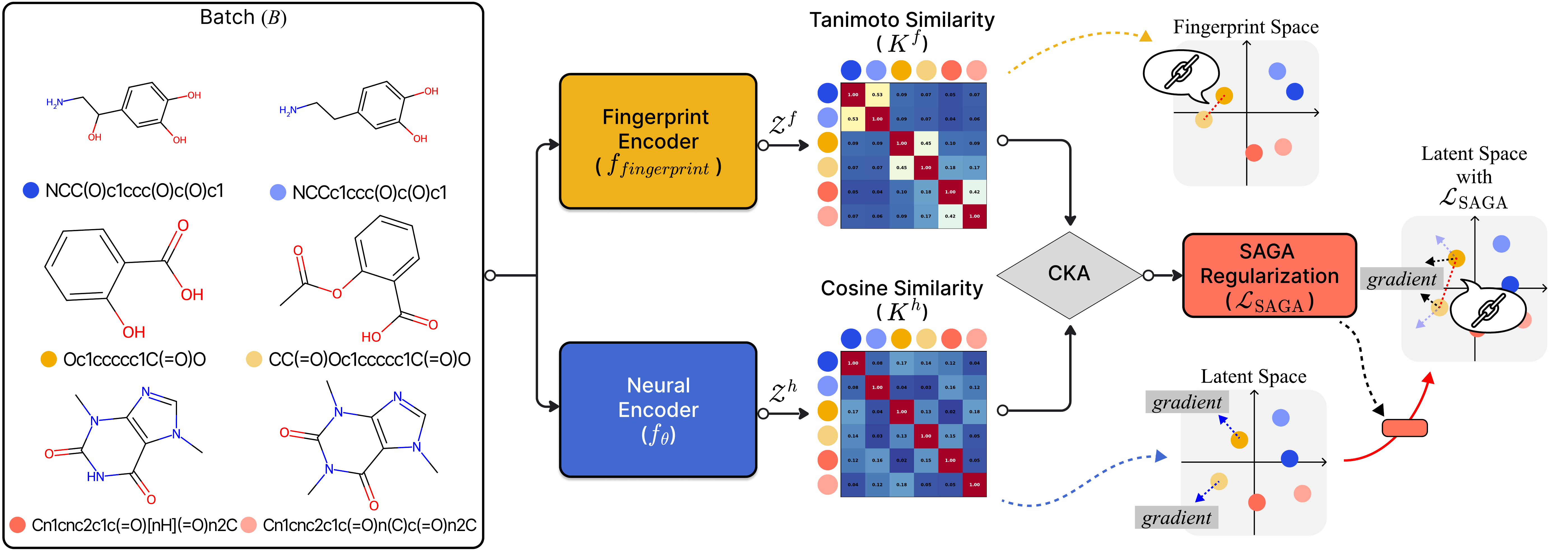}
\caption{\textbf{Overview of \methodname.} A CKA-based auxiliary loss aligns pairwise similarities among neural molecular embeddings with fingerprint-based structural similarities, transferring a classical chemoinformatics prior into molecule-language models without adding trainable parameters or requiring fingerprints at inference.}
\label{fig:method-overview}
\end{figure}

Here we show that this classical structural prior can improve molecule-language representations when used only during training. We introduce \methodname\ (Structure-Aware Guided Alignment), a CKA-based regularizer that aligns neural molecular representations with fingerprint-based Tanimoto similarity without adding trainable parameters or requiring fingerprints at inference (Fig.~\ref{fig:method-overview}). We first test whether fingerprint similarity provides a property-relevant structural signal across ADMET endpoints and then evaluate \methodname\ in MolCA and Atomas backbones. The resulting representations improve molecule--text retrieval, shift non-exact retrieval toward closer structural analogues, and yield positive average gains in scaffold-split property prediction. Molecular description and text-to-molecule generation ablations further show that performance is improved or preserved, while representation analyses indicate that the training-time structural prior reorganizes molecule-language representations around chemically meaningful similarity.

\section{Results}

\subsection{Fingerprint similarity as a structural prior}

Before introducing \methodname, we first examined whether the structural signal used as its training prior captures chemically relevant relationships, and whether CKA can compare those relationships across representation spaces.

As an external readout of chemical relevance, we analyzed 20 TDC ADMET benchmark datasets~\cite{huang2021tdc} (7 regression, 13 binary classification), computing pairwise Tanimoto similarity with Morgan fingerprints (radius=2, 2048 bits) under stratified sampling across 10 similarity bins (details in Methods).

\begin{figure*}[!ht]
\centering
\begin{minipage}[t]{0.45\textwidth}
\raggedright
\textbf{(a)}\\[0.5em]
\centering
\includegraphics[width=0.95\textwidth]{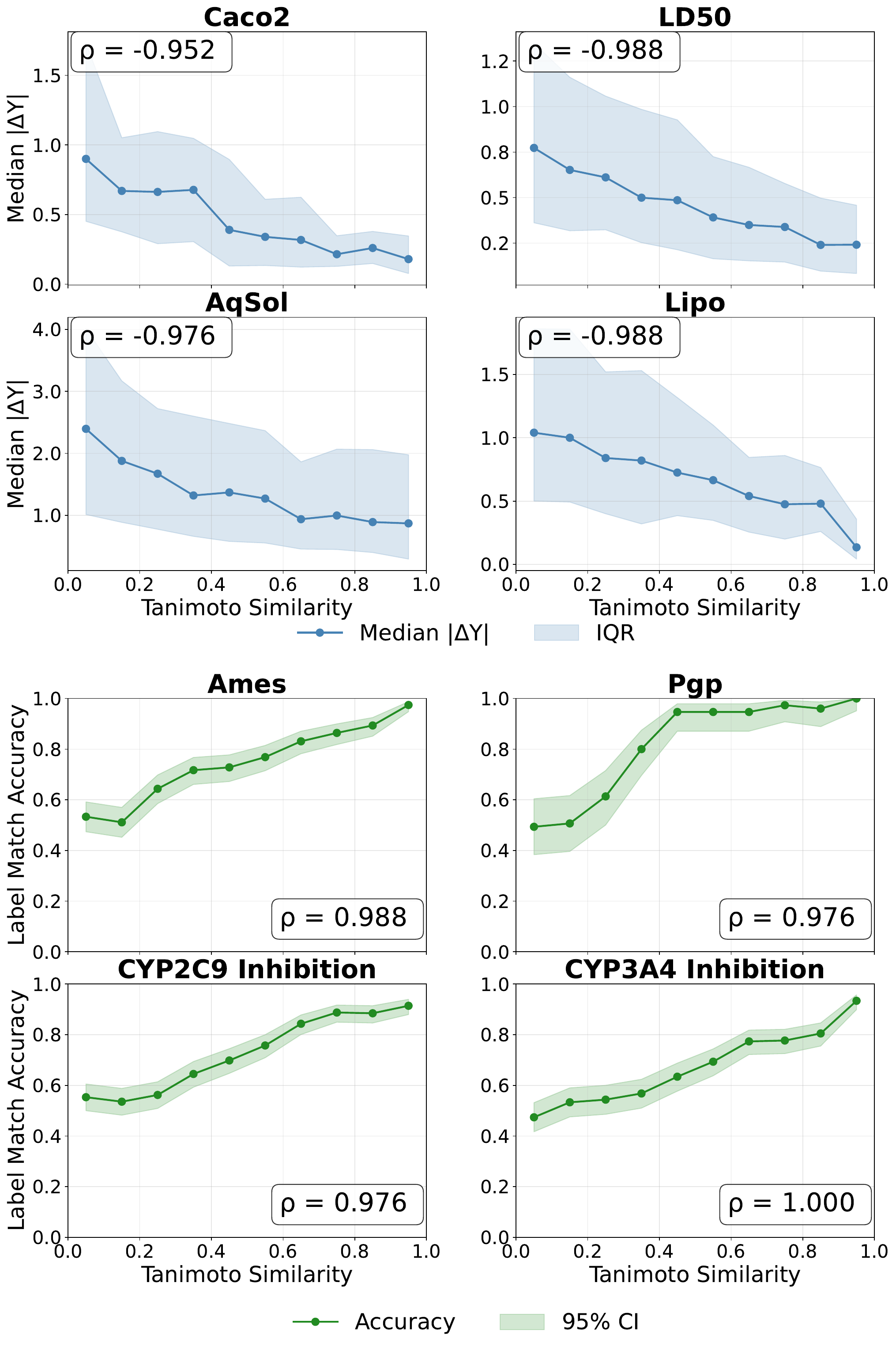}
\end{minipage}%
\hfill
\begin{minipage}[t]{0.52\textwidth}
\raggedright
\textbf{(b)}\\[0.3em]
\centering
{\footnotesize
\begin{tabular*}{\textwidth}{@{\extracolsep{\fill}}lrccc@{}}
\toprule
\textbf{Dataset} & \textbf{$N$} & \textbf{$\rho$} & \textbf{$p$} & \textbf{Effect ($\delta$/$h$)} \\
\midrule
\multicolumn{5}{l}{\textit{Regression}} \\
\midrule
Caco2       & 637 & $-0.95$ & $<.001$ & $0.76$ \\
Lipo        & 2,940 & $-0.99$ & $<.001$ & $0.79$ \\
AqSol       & 6,986 & $-0.98$ & $<.001$ & $0.46$ \\
PPBR        & 1,952 & $-0.96$ & $<.001$ & $0.56$ \\
VDss        & 791 & $-0.86$ & $.014$ & $0.46$ \\
Half\_Life  & 465 & $-0.78$ & $.041$ & $0.41$ \\
LD50        & 5,168 & $-0.99$ & $<.001$ & $0.59$ \\
\midrule
\multicolumn{5}{l}{\textit{Binary Classification}} \\
\midrule
Pgp         & 851 & $0.98$ & $<.001$ & $0.85$ \\
BBB         & 1,421 & $0.88$ & $<.001$ & $0.32$ \\
CYP2C9\_Inh & 8,463 & $0.98$ & $<.001$ & $0.60$ \\
CYP2D6\_Inh & 9,191 & $0.99$ & $<.001$ & $0.41$ \\
CYP3A4\_Inh & 8,628 & $1.00$ & $<.001$ & $0.53$ \\
CYP2D6\_Sub & 465 & $0.94$ & $.005$ & $0.24$ \\
CYP3A4\_Sub & 468 & $0.89$ & $.019$ & $0.38$ \\
Ames        & 5,093 & $0.99$ & $<.001$ & $0.57$ \\
HIA         & 403 & $0.64$ & $.173$ & $0.31$ \\
Bioav       & 448 & $0.62$ & $.191$ & $0.11$ \\
CYP2C9\_Sub & 467 & $0.73$ & $.103$ & $0.35$ \\
hERG        & 457 & $0.51$ & $.131$ & $0.26$ \\
DILI        & 331 & $0.73$ & $.103$ & $0.84$ \\
\bottomrule
\end{tabular*}}

\vspace{0.5em}
\raggedright
\textbf{(c)}\\[0.3em]
\centering
\small
\begin{tabular*}{\textwidth}{@{\extracolsep{\fill}}lccc@{}}
\toprule
\textbf{$n$} & \textbf{Original} & \textbf{Shuffled} & \textbf{$\Delta$} \\
\midrule
100 & $0.504 \pm 0.047$ & $0.001 \pm 0.018$ & $+0.50$ \\
200 & $0.474 \pm 0.029$ & $0.000 \pm 0.004$ & $+0.47$ \\
500 & $0.501 \pm 0.009$ & $-0.001 \pm 0.003$ & $+0.50$ \\
\bottomrule
\end{tabular*}
\end{minipage}

\caption{\textbf{Structural similarity and relational alignment.} \textbf{(a)} Structure-property relationship showing median property difference (regression) and label agreement rate (classification) across Tanimoto similarity bins. \textbf{(b)} Statistical analysis on 20 TDC ADMET datasets. $\rho$: Spearman correlation; Effect: Cliff's $\delta$ (regression) / Cohen's $h$ (classification). \textbf{(c)} CKA comparison of matched and shuffled structure-text pairs.}
\label{fig:preliminary-analysis}
\end{figure*}

Fig.~\ref{fig:preliminary-analysis}a illustrates this structure-property relationship on representative regression and classification tasks: property differences decrease with structural similarity for regression, while label agreement increases for classification. Across all 20 datasets (Fig.~\ref{fig:preliminary-analysis}b; individual plots in Supplementary Fig.~1), 15/20 datasets (75\%) exhibit statistically significant monotonic relationships ($p < 0.05$). All 7 regression datasets yield strong negative Spearman correlations ($\rho = -0.78$ to $-0.99$) with consistently large Cliff's $\delta$ ($0.41$--$0.79$), and 8/13 classification datasets show significant positive correlations ($\rho = 0.88$--$1.00$). The remaining five classification datasets show positive but non-significant correlations, indicating that the structural prior is broadly informative but not uniformly significant across all ADMET endpoints.

We next tested whether CKA can detect relational correspondence between molecular structure and language on ChEBI-20~\cite{edwards2021text2mol} (Fig.~\ref{fig:preliminary-analysis}c). Matched structure-text pairs yield CKA values near $0.5$ ($0.504 \pm 0.047$ at $n=100$), whereas shuffled pairs stay near zero ($0.001 \pm 0.018$), and this $\Delta \approx 0.5$ gap is stable across $n \in \{100, 200, 500\}$. This separation supports the use of CKA as a relational alignment statistic for comparing fingerprint-derived structural relationships with learned molecule-language representations.

\subsection{SAGA aligns neural representations with fingerprint similarity}

Building on these findings, we used fingerprint similarity as a training-time structural prior for molecule-language representation learning. We refer to this objective as \textbf{SAGA} (\textbf{S}tructure-\textbf{A}ware \textbf{G}uided \textbf{A}lignment). \methodname\ adds a CKA-based auxiliary loss that aligns pairwise relationships among neural molecular embeddings with the corresponding relationships among fingerprint representations. This guides the learned molecular space toward structural organization while requiring no fingerprint features at inference time.

We integrated this relational prior into two molecule-language models with different molecular encoders and alignment mechanisms. MolCA~\cite{liu2023molca} uses a Q-Former~\cite{li2023blip2} to fuse molecular graphs with language through learned query tokens and cross-attention, and is trained on mini-batches. Atomas~\cite{atomas2024} directly projects SMILES representations into a shared embedding space using MoCo-style~\cite{he2020momentum} momentum queues; we evaluated both its Base (Atomas-B) and Large (Atomas-L) variants. This comparison tests whether a fingerprint-derived relational prior can be incorporated into both graph- and sequence-based molecule-language representations.

Because the two models construct training batches differently, the same structural prior is implemented with model-specific sampling. For MolCA, CKA is computed directly over mini-batches. For Atomas, we maintain a parallel fingerprint queue alongside the existing feature queue, allowing CKA to operate on a larger sample set without increasing the batch size. Full loss formulations are in Methods.

We then evaluated whether this training-only prior improves chemically meaningful use of the learned representation space. Cross-modal retrieval on ChEBI-20~\cite{edwards2021text2mol} tests molecule--text alignment, exclude-self retrieval asks whether non-exact hits become closer structural analogues, MoleculeNet property prediction~\cite{wu2018moleculenet} tests transfer to scaffold-held-out molecules, and molecular generation and description examine whether the same prior introduces trade-offs in decoder-mediated tasks.

\subsection{Molecule--text retrieval with structural priors}

We first evaluated molecule--text retrieval on ChEBI-20, where the learned representation space is used directly for nearest-neighbour matching between molecular structures and textual descriptions. \methodname\ improved every reported retrieval metric for all three backbones and in both retrieval directions (Table~\ref{tab:retrieval}). In Text-to-Molecule retrieval, R@1 increased from 31.48 to 35.82 for MolCA, from 33.56 to 40.44 for Atomas-B, and from 43.15 to 50.97 for Atomas-L. The reverse Molecule-to-Text direction showed the same pattern, with R@1 gains of +2.40, +5.60 and +4.84 percentage points, respectively.

These bidirectional gains are informative because the CKA term is applied to the molecular representations; it does not directly regularize the text encoder. The improvement in Molecule-to-Text retrieval therefore suggests that organizing molecular embeddings by structural relationships can indirectly strengthen language alignment through the contrastive objective. This effect was observed in both the graph-based MolCA model and the SMILES-based Atomas variants, indicating that the training-time structural prior is not tied to a single encoder design.

\begin{table}[t]
\centering
\caption{Molecule--text retrieval performance on the ChEBI-20 validation split. R@K denotes Recall at K and MRR denotes Mean Reciprocal Rank. $\Delta$: absolute difference. Atomas-B and Atomas-L denote Base and Large variants, respectively. Best results for each base model are shown in \textbf{bold}.}
\label{tab:retrieval}
\begin{tabular}{lcccccccc}
\toprule
& \multicolumn{4}{c}{\textbf{Text $\rightarrow$ Molecule}} & \multicolumn{4}{c}{\textbf{Molecule $\rightarrow$ Text}} \\
\cmidrule(lr){2-5} \cmidrule(lr){6-9}
\textbf{Model} & R@1 & R@5 & R@10 & MRR & R@1 & R@5 & R@10 & MRR \\
\midrule
MolCA & 31.48 & 48.30 & 54.09 & 39.31 & 29.15 & 44.33 & 48.73 & 36.32 \\
+ \methodname & \textbf{35.82} & \textbf{57.55} & \textbf{65.42} & \textbf{45.88} & \textbf{31.55} & \textbf{49.73} & \textbf{56.73} & \textbf{40.29} \\
\rowcolor{gray!10} \quad $\Delta$ & +4.34 & +9.25 & +11.33 & +6.57 & +2.40 & +5.40 & +8.00 & +3.97 \\
\rowcolor{gray!10} \quad (\%) & (+13.8\%) & (+19.2\%) & (+21.0\%) & (+16.7\%) & (+8.2\%) & (+12.2\%) & (+16.4\%) & (+10.9\%) \\
\midrule
Atomas-B & 33.56 & 58.62 & 68.85 & 45.33 & 31.05 & 55.75 & 65.19 & 42.90 \\
+ \methodname & \textbf{40.44} & \textbf{67.70} & \textbf{76.39} & \textbf{52.80} & \textbf{36.65} & \textbf{62.11} & \textbf{71.19} & \textbf{48.50} \\
\rowcolor{gray!10} \quad $\Delta$ & +6.88 & +9.08 & +7.54 & +7.47 & +5.60 & +6.36 & +6.00 & +5.60 \\
\rowcolor{gray!10} \quad (\%) & (+20.5\%) & (+15.5\%) & (+11.0\%) & (+16.5\%) & (+18.0\%) & (+11.4\%) & (+9.2\%) & (+13.1\%) \\
\midrule
Atomas-L & 43.15 & 69.21 & 77.96 & 55.23 & 41.56 & 65.01 & 73.70 & 52.49 \\
+ \methodname & \textbf{50.97} & \textbf{76.48} & \textbf{83.47} & \textbf{62.49} & \textbf{46.40} & \textbf{71.70} & \textbf{78.30} & \textbf{57.74} \\
\rowcolor{gray!10} \quad $\Delta$ & +7.82 & +7.27 & +5.51 & +7.26 & +4.84 & +6.69 & +4.60 & +5.25 \\
\rowcolor{gray!10} \quad (\%) & (+18.1\%) & (+10.5\%) & (+7.1\%) & (+13.1\%) & (+11.6\%) & (+10.3\%) & (+6.2\%) & (+10.0\%) \\
\bottomrule
\end{tabular}
\end{table}

\subsection{Structure-aware retrieval of chemical analogues}

Exact-match retrieval does not by itself show whether a model retrieves chemically meaningful alternatives when the paired target is absent. We therefore removed each exact target on the ChEBI-20 validation split and measured Tanimoto similarity among the next-ranked molecules (Fig.~\ref{fig:casestudy-retrieval}).

\begin{center}
\begin{minipage}{\textwidth}
\centering
\includegraphics[width=0.72\textwidth]{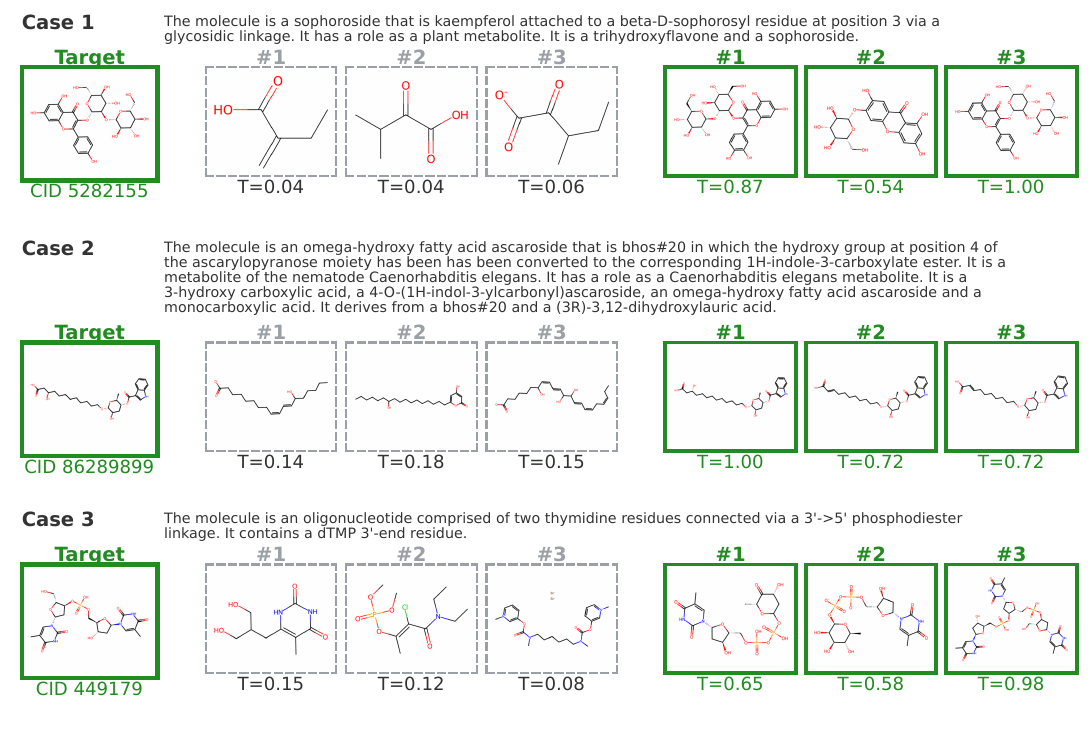}
\captionsetup{type=figure,hypcap=false,font=small}
\caption{\textbf{\methodname\ aligns cross-modal retrieval to chemical structure.}
Three representative Text$\rightarrow$Molecule cases spanning distinct chemotypes: a flavonol glycoside, an ascaroside, and a dinucleotide. From left to right, each row shows the target molecule, the top-3 non-exact molecules retrieved by MolCA, and the top-3 non-exact molecules retrieved by MolCA$+$\methodname. Non-exact denotes exclusion of the paired target entry; a separate database entry may therefore have $T=1.00$ under the Morgan fingerprint. Retrieved molecules are annotated by their Tanimoto similarity $T$ to the target (Morgan fingerprints, radius 2, 2{,}048 bits); green outlines mark structural analogues ($T\geq0.40$).}
\label{fig:casestudy-retrieval}
\end{minipage}
\end{center}

\begin{figure*}[!t]
\centering
\noindent
\begin{minipage}[t]{0.675\textwidth}
{\normalsize\textbf{(a)}}\par\vspace{0.2em}
\includegraphics[width=\linewidth]{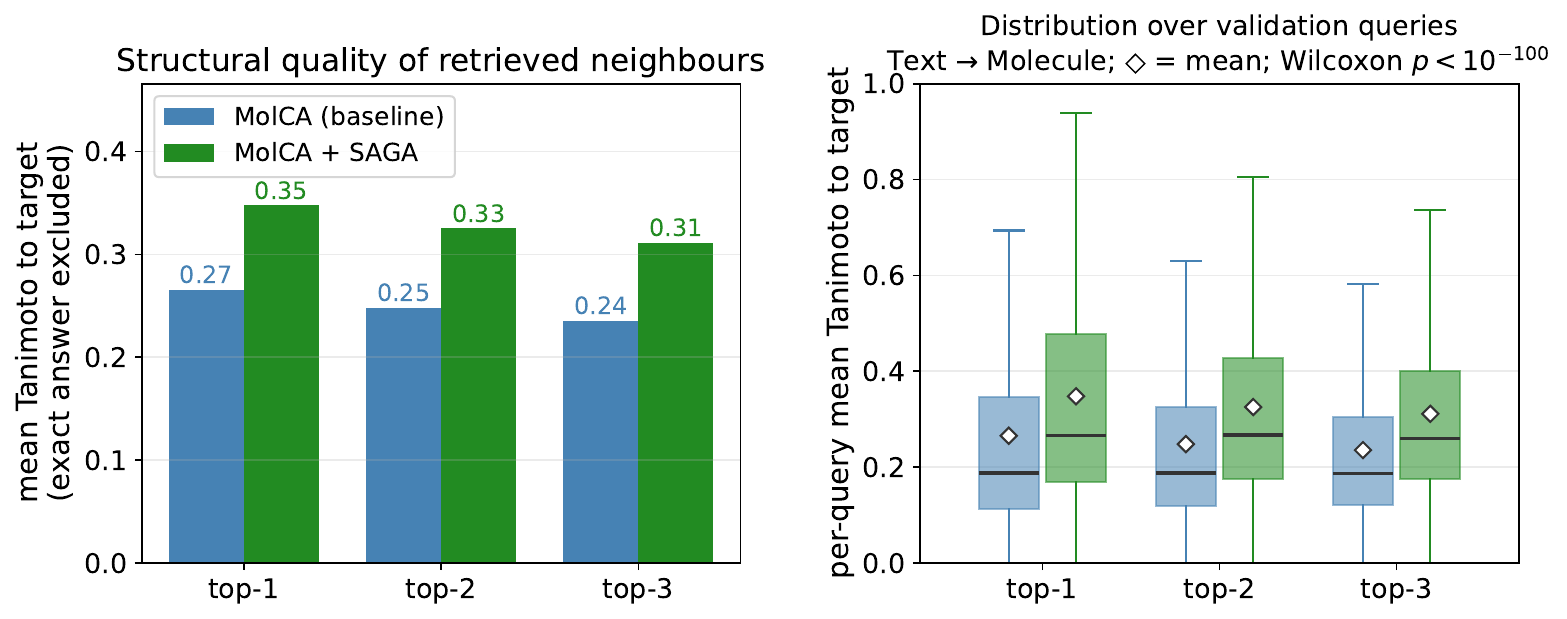}
\end{minipage}\hfill
\begin{minipage}[t]{0.305\textwidth}
{\normalsize\textbf{(b)}}\par\vspace{0.2em}
\includegraphics[width=\linewidth]{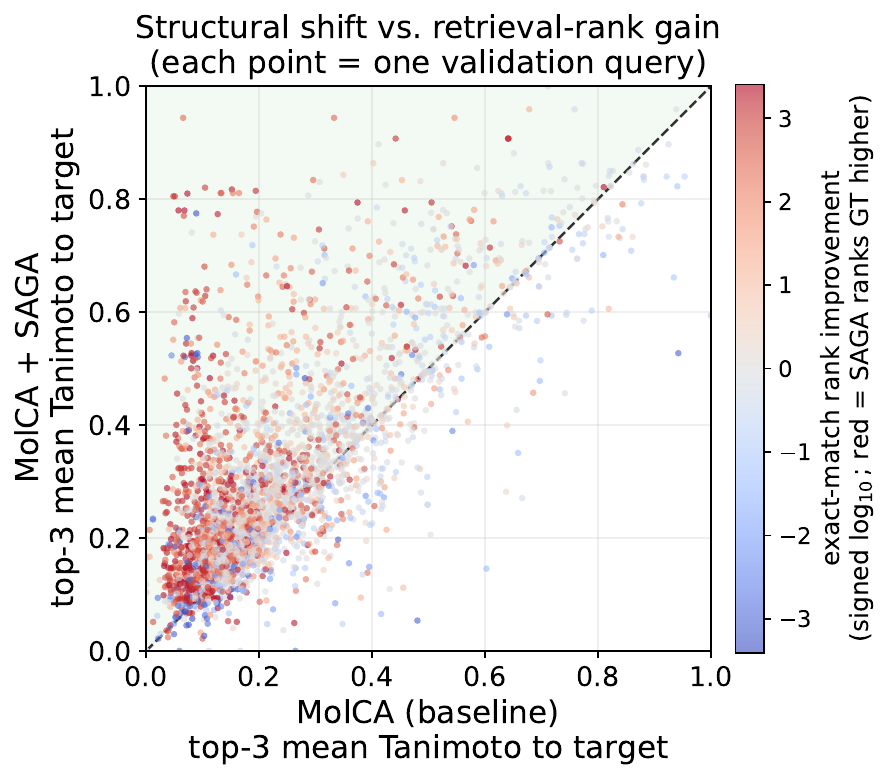}
\end{minipage}
\caption{\textbf{\methodname\ improves the structural quality of retrieved neighbours.}
\textbf{(a)} Exclude-self structural neighbourhoods for MolCA Text$\rightarrow$Molecule retrieval on the ChEBI-20 validation split ($n=3{,}301$). For each query, the exact target was removed from the candidate pool and the Tanimoto similarity between the target and the next retrieved molecules was measured using Morgan fingerprints (radius 2, 2{,}048 bits). Bars summarize mean Tanimoto values for the top-1, top-2, and top-3 non-exact neighbours, and box plots show the per-query distributions. Differences are significant by one-sided Wilcoxon signed-rank test ($p<10^{-100}$).
\textbf{(b)} Per-query structural shift versus exact-match rank gain in Text$\rightarrow$Molecule retrieval. Points above the diagonal indicate queries for which MolCA$+$\methodname\ retrieves structurally closer non-exact neighbours, and colour denotes the signed log-rank improvement of the exact target.}
\label{fig:casestudy-structural}
\end{figure*}

In MolCA, \methodname\ retrieved substantially closer structural analogues than the baseline (Fig.~\ref{fig:casestudy-structural}a). For Text-to-Molecule retrieval, the mean Tanimoto similarity of the top-3 non-exact neighbours increased from 0.235 to 0.311, and \methodname\ retrieved structurally closer neighbours for 76.3\% of validation queries (one-sided Wilcoxon signed-rank test, $p<10^{-100}$). The same pattern held in the reverse direction, where the Molecule-to-Text top-3 neighbourhood improved from 0.211 to 0.272 and was closer for 73.8\% of queries, and extended to both Atomas-B and Atomas-L in both retrieval directions (Supplementary Figs.~8--12). In representative examples, the descriptions associated with \methodname-retrieved neighbours also shared structural-class terms with the query descriptions, providing a text-side view of the same analogue-retrieval effect (Supplementary Figs.~13--15). A per-query scatter analysis further shows that queries with the largest exact-match rank gains tend to move above the structural-similarity diagonal, linking improved retrieval rank to improved chemical neighbourhood quality (Fig.~\ref{fig:casestudy-structural}b). These results indicate that the retrieval gain is not only an improvement in paired molecule--text ranking, but also a shift toward chemically coherent analogue retrieval.

\FloatBarrier

\subsection{Scaffold-split molecular property prediction}

We next asked whether the structurally reorganized embeddings support property prediction for molecules outside the training scaffold distribution. For each MoleculeNet task, we froze the pre-trained molecular encoder and trained only a two-layer prediction head (Methods), so performance reflects information available in the learned representation rather than full end-to-end adaptation. All evaluations use scaffold splits, so the test molecules contain structural frameworks held out from training rather than random compounds from the same scaffold distribution.

Across eight classification tasks, \methodname\ increased average ROC-AUC for all three backbones: MolCA (+4.1\%), Atomas-B (+2.0\%) and Atomas-L (+1.4\%; Table~\ref{tab:property}). The gains were not uniform across endpoints, but they were broadly positive, with improvements on 6/8, 7/8 and 7/8 tasks, respectively. MolCA showed the largest average effect, driven by MUV (+21.5\%) and BACE (+7.0\%), whereas Atomas variants showed smaller but mostly positive transfer. Isolated drops, such as HIV for Atomas-B and MUV for Atomas-L, indicate that the structural prior is not universally beneficial for every endpoint. Nevertheless, the positive average gains under scaffold splits suggest that the training-time prior improves representations in settings that require transfer to structurally distinct molecules, complementing the analogue-retrieval results above.

\begin{table}[t]
\centering
\caption{Molecular property prediction performance (ROC-AUC) on scaffold-split MoleculeNet classification tasks. Results are reported as mean $\pm$ standard deviation over multiple runs. Atomas-B and Atomas-L denote Base and Large variants, respectively. Best results for each base model are shown in \textbf{bold}.}
\label{tab:property}
\begin{tabular}{lccccccccc}
\toprule
\textbf{Model} & \textbf{BBBP} & \textbf{Tox21} & \textbf{ToxCast} & \textbf{SIDER} & \textbf{ClinTox} & \textbf{MUV} & \textbf{HIV} & \textbf{BACE} & \textbf{Avg} \\
\midrule
MolCA & 0.681{\tiny$\pm$0.008} & \textbf{0.704}{\tiny$\pm$0.005} & 0.687{\tiny$\pm$0.003} & 0.628{\tiny$\pm$0.007} & \textbf{0.672}{\tiny$\pm$0.030} & 0.536{\tiny$\pm$0.030} & 0.749{\tiny$\pm$0.013} & 0.756{\tiny$\pm$0.008} & 0.677 \\
+ \methodname & \textbf{0.688}{\tiny$\pm$0.013} & 0.701{\tiny$\pm$0.007} & \textbf{0.715}{\tiny$\pm$0.007} & \textbf{0.652}{\tiny$\pm$0.006} & 0.669{\tiny$\pm$0.042} & \textbf{0.651}{\tiny$\pm$0.009} & \textbf{0.754}{\tiny$\pm$0.007} & \textbf{0.809}{\tiny$\pm$0.010} & \textbf{0.705} \\
\midrule
Atomas-B & 0.703{\tiny$\pm$0.003} & 0.659{\tiny$\pm$0.004} & 0.670{\tiny$\pm$0.005} & 0.584{\tiny$\pm$0.015} & 0.955{\tiny$\pm$0.003} & 0.579{\tiny$\pm$0.011} & \textbf{0.739}{\tiny$\pm$0.009} & 0.742{\tiny$\pm$0.016} & 0.704 \\
+ \methodname & \textbf{0.704}{\tiny$\pm$0.007} & \textbf{0.676}{\tiny$\pm$0.002} & \textbf{0.677}{\tiny$\pm$0.007} & \textbf{0.621}{\tiny$\pm$0.009} & \textbf{0.994}{\tiny$\pm$0.001} & \textbf{0.605}{\tiny$\pm$0.007} & 0.714{\tiny$\pm$0.009} & \textbf{0.753}{\tiny$\pm$0.009} & \textbf{0.718} \\
\midrule
Atomas-L & 0.726{\tiny$\pm$0.007} & 0.683{\tiny$\pm$0.006} & 0.661{\tiny$\pm$0.008} & 0.604{\tiny$\pm$0.003} & 0.959{\tiny$\pm$0.002} & \textbf{0.632}{\tiny$\pm$0.015} & 0.716{\tiny$\pm$0.022} & 0.741{\tiny$\pm$0.015} & 0.715 \\
+ \methodname & \textbf{0.735}{\tiny$\pm$0.012} & \textbf{0.690}{\tiny$\pm$0.006} & \textbf{0.677}{\tiny$\pm$0.004} & \textbf{0.614}{\tiny$\pm$0.005} & \textbf{0.978}{\tiny$\pm$0.014} & 0.591{\tiny$\pm$0.025} & \textbf{0.741}{\tiny$\pm$0.006} & \textbf{0.771}{\tiny$\pm$0.012} & \textbf{0.725} \\
\bottomrule
\end{tabular}
\end{table}

\subsection{Molecular generation and description}

Because \methodname\ reorganizes the embedding space, we next asked whether this change disrupts tasks whose outputs are produced through a decoder rather than read directly from nearest-neighbour geometry. We evaluated the \methodname\ ablation on two generation settings supported by the underlying backbones: de novo molecular generation from text in Atomas, and molecular description generation from structure in MolCA and Atomas.

\begin{table}[t]
\centering
\caption{\textbf{Generation and description performance on ChEBI-20.}
Rows compare each baseline with its \methodname\ counterpart. Values are single-run results. Bounded metrics are multiplied by 100, whereas Levenshtein is reported as raw edit distance. Text$\rightarrow$Molecule generation is evaluated for Atomas on the validation split. Molecule$\rightarrow$Text description is evaluated on the MolCA test split and the Atomas validation split. Best results within each model pair are shown in \textbf{bold}.}
\label{tab:generation-description}
\small
\textbf{a. Text$\rightarrow$Molecule de novo molecular generation}

\vspace{0.3em}
\begin{tabular}{lccccccc}
\toprule
\textbf{Model} & \textbf{BLEU$\uparrow$} & \textbf{Exact$\uparrow$} & \textbf{Lev.$\downarrow$} & \textbf{MACCS$\uparrow$} & \textbf{RDK$\uparrow$} & \textbf{Morgan$\uparrow$} & \textbf{Valid.$\uparrow$} \\
\midrule
Atomas-B & 79.20 & \textbf{17.20} & 20.86 & 85.70 & 74.60 & 68.30 & \textbf{92.50} \\
+ \methodname & \textbf{79.40} & 17.10 & \textbf{20.83} & \textbf{86.10} & \textbf{75.10} & \textbf{68.70} & 92.30 \\
\midrule
Atomas-L & \textbf{82.80} & \textbf{24.10} & \textbf{17.24} & 89.00 & 80.10 & \textbf{74.50} & \textbf{94.50} \\
+ \methodname & 82.60 & \textbf{24.10} & 17.25 & \textbf{89.10} & \textbf{80.30} & 74.40 & \textbf{94.50} \\
\bottomrule
\end{tabular}

\vspace{0.8em}
\textbf{b. Molecule$\rightarrow$Text description generation}

\vspace{0.3em}
\begin{tabular}{lccccc}
\toprule
\textbf{Model} & \textbf{BLEU-2$\uparrow$} & \textbf{BLEU-4$\uparrow$} & \textbf{ROUGE-1$\uparrow$} & \textbf{ROUGE-2$\uparrow$} & \textbf{ROUGE-L$\uparrow$} \\
\midrule
MolCA & 57.59 & 48.75 & 65.65 & 50.80 & 59.48 \\
+ \methodname & \textbf{58.00} & \textbf{49.18} & \textbf{65.77} & \textbf{50.98} & \textbf{59.57} \\
\midrule
Atomas-B & \textbf{61.10} & \textbf{53.60} & \textbf{67.30} & \textbf{54.20} & \textbf{62.20} \\
+ \methodname & \textbf{61.10} & 53.50 & 67.20 & 53.90 & 62.00 \\
\midrule
Atomas-L & 67.60 & 61.10 & 71.90 & 60.40 & 67.40 \\
+ \methodname & \textbf{68.40} & \textbf{62.30} & \textbf{72.90} & \textbf{61.80} & \textbf{68.50} \\
\bottomrule
\end{tabular}
\end{table}

In these single-run ablations, the effect is task-dependent but shows no material trade-off (Table~\ref{tab:generation-description}). Description generation improves for MolCA and Atomas-L, while Atomas-B remains essentially unchanged, consistent with the fact that decoder-mediated outputs can absorb part of the representational change introduced by \methodname. In text-to-molecule generation, \methodname\ slightly improves BLEU and fingerprint similarity for Atomas-B and preserves the stronger Atomas-L generator, including unchanged validity. Thus, the largest gains occur when downstream tasks directly consume the reorganized embedding geometry, while decoder-mediated generation is improved or largely maintained rather than degraded.

\FloatBarrier
\subsection{Sensitivity of structural prior integration}

We next examined how the structural prior depends on three design choices: regularization weight ($\lambda$), fingerprint type and CKA sample size ($n$). Full validation tables and training curves are provided in Supplementary Note~1, Supplementary Tables~1--3 and Supplementary Figs.~2--4.

Regularization weight produced a capacity-dependent response. MolCA improved up to $\lambda=100$ (+11.9\% R@1) but decreased at $\lambda=1000$, whereas Atomas-B also peaked at $\lambda=100$ (+20.5\% R@1) and Atomas-L remained robust at larger weights. We therefore use $\lambda=100$ as the default setting, which gives strong gains across all three backbones while avoiding the high-weight regime that destabilizes the smaller or more constrained models.

Fingerprint type showed that the best structural target depends on the molecular encoder. Morgan fingerprints performed best for the graph-based MolCA model, whereas MACCS keys were better for the SMILES-based Atomas variants (Atomas-L, 53.32\% versus 50.97\% R@1 with Morgan; Atomas-B, 41.92\% versus 40.44\%). RDKit topological fingerprints were consistently weaker and degraded MolCA, suggesting that path-based similarity provides a less compatible relational signal for these molecule-language objectives. This model-dependent behaviour indicates that the structural prior is not an arbitrary auxiliary target; its benefit depends on compatibility between the fingerprint relation and the molecular encoder.

The CKA sample size also showed an intermediate optimum. MolCA gains were stable across batch sizes despite a lower large-batch baseline, while Atomas-B and Atomas-L performed best around $n=128$ and lost benefit at very large sample sizes. This pattern suggests that the CKA target should be large enough to estimate structural relationships reliably but not so large that it imposes overly detailed pairwise constraints during training.

\FloatBarrier
\subsection{Representation-space interpretation}

\begin{figure}[p]
\centering
\includegraphics[width=\textwidth]{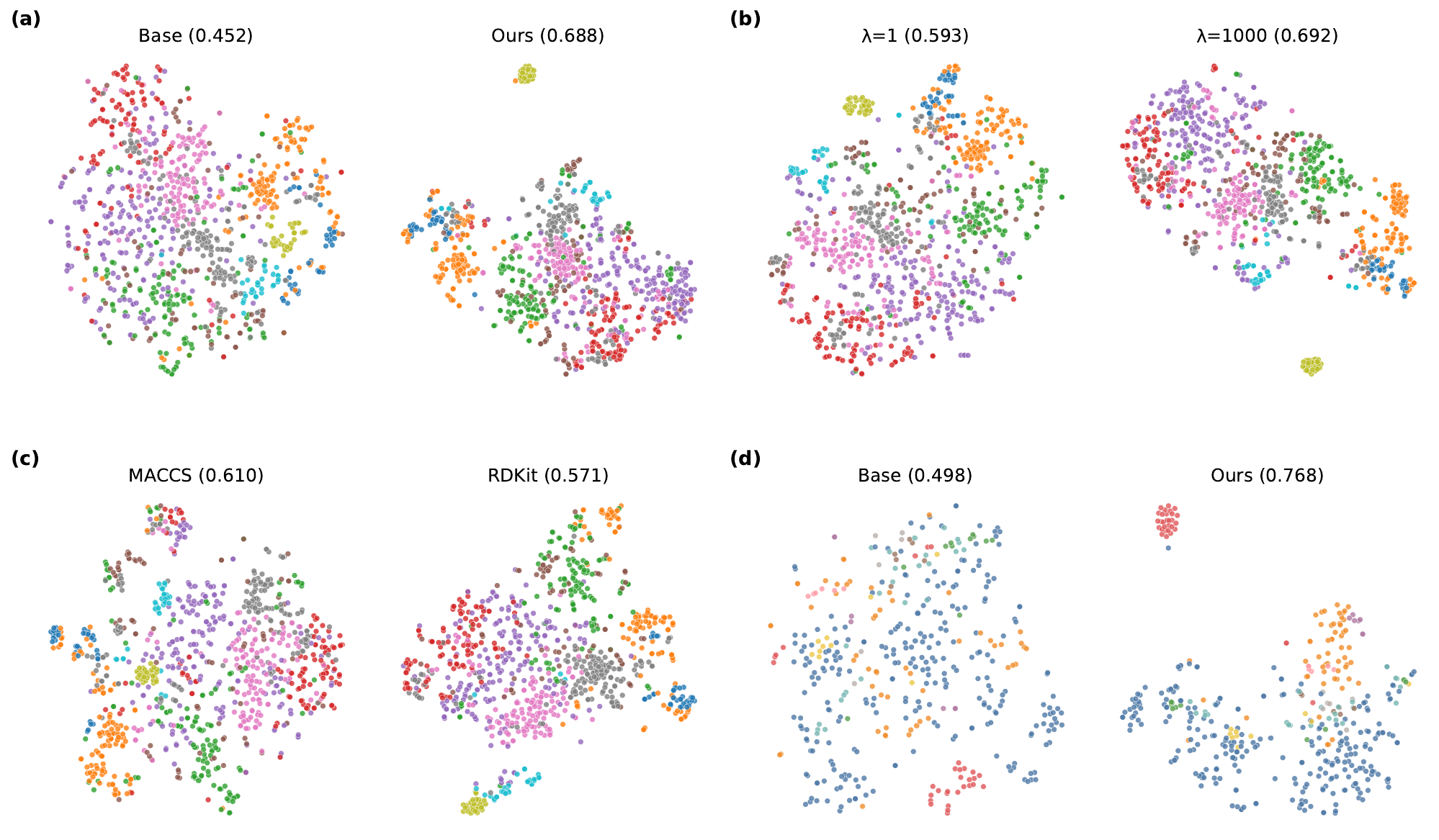}

\vspace{0.5em}
\makebox[\textwidth][l]{\hspace{0.01\textwidth}\textbf{(e)}}\\[0.3em]
\footnotesize
\begin{tabular*}{\textwidth}{l@{\extracolsep{\fill}}ccccccc}
\toprule
& \multicolumn{3}{c}{\textbf{CKA Alignment}} & \multicolumn{2}{c}{\textbf{Mol Clustering}} & \multicolumn{2}{c}{\textbf{Text Clustering}} \\
\cmidrule(lr){2-4} \cmidrule(lr){5-6} \cmidrule(lr){7-8}
\textbf{Model} & Mol$\leftrightarrow$FP & Mol$\leftrightarrow$Text & Text$\leftrightarrow$FP & Sil & CH & Sil & CH \\
\midrule
\rowcolor{gray!10} \multicolumn{8}{l}{\textit{MolCA (Graph Embedding)}} \\
Base & 0.452 & 0.281 & 0.249 & $-$0.004 & 15.6 & $-$0.000 & 3.5 \\
$\lambda=1$ & 0.593 & 0.283 & 0.254 & $-$0.011 & 20.9 & 0.000 & 3.3 \\
$\lambda=100$ (Ours) & 0.688 & 0.323 & 0.458 & 0.016 & \textbf{44.1} & 0.018 & 9.2 \\
$\lambda=1000$ & \textbf{0.692} & \textbf{0.375} & \textbf{0.528} & \textbf{0.036} & 42.3 & \textbf{0.028} & \textbf{13.1} \\
\cdashline{1-8}
MACCS & 0.610 & 0.337 & 0.349 & $-$0.039 & 25.6 & 0.012 & 7.2 \\
RDKit & 0.571 & 0.278 & 0.328 & 0.013 & 31.5 & 0.011 & 7.0 \\
\midrule
\rowcolor{gray!10} \multicolumn{8}{l}{\textit{Atomas-B (SMILES Embedding)}} \\
Base & 0.595 & \textbf{0.569} & 0.478 & 0.008 & 13.4 & 0.008 & 9.1 \\
$\lambda=1$ & 0.628 & 0.550 & 0.476 & 0.019 & 15.0 & 0.008 & 8.1 \\
$\lambda=100$ (Ours) & 0.784 & 0.553 & 0.504 & 0.030 & 18.6 & \textbf{0.011} & 8.7 \\
$\lambda=1000$ & \textbf{0.843} & 0.514 & \textbf{0.519} & \textbf{0.038} & 19.0 & 0.009 & 7.7 \\
\cdashline{1-8}
MACCS & 0.770 & 0.567 & 0.489 & 0.035 & 19.5 & 0.009 & \textbf{9.2} \\
RDKit & 0.715 & 0.552 & 0.422 & 0.036 & \textbf{20.6} & 0.009 & 8.7 \\
\midrule
\rowcolor{gray!10} \multicolumn{8}{l}{\textit{Atomas-L (SMILES Embedding)}} \\
Base & 0.581 & 0.580 & 0.465 & 0.013 & 13.7 & 0.009 & 9.4 \\
$\lambda=1$ & 0.628 & 0.635 & 0.499 & 0.017 & 15.6 & 0.012 & 10.9 \\
$\lambda=100$ (Ours) & 0.788 & 0.590 & 0.514 & 0.032 & 18.8 & 0.012 & 9.4 \\
$\lambda=1000$ & \textbf{0.854} & 0.571 & \textbf{0.560} & \textbf{0.044} & 18.8 & 0.012 & 8.8 \\
\cdashline{1-8}
MACCS & 0.804 & \textbf{0.642} & 0.533 & 0.040 & \textbf{21.1} & 0.012 & \textbf{11.1} \\
RDKit & 0.717 & 0.616 & 0.436 & 0.036 & 20.6 & 0.008 & 10.0 \\
\bottomrule
\end{tabular*}

\caption{\textbf{Representation space analysis.} t-SNE of MolCA graph embeddings (a--d) and quantitative metrics for all models (e). \textbf{(a)} Base vs \methodname\ ($\lambda=100$, Morgan), colored by Tanimoto-based clusters. \textbf{(b)} Effect of regularization weight $\lambda$. \textbf{(c)} Fingerprint type comparison. \textbf{(d)} Base vs \methodname\ colored by Murcko scaffold identity. Parenthetical values in (a--d) denote molecule--fingerprint CKA. \textbf{(e)} CKA alignment and clustering metrics (Silhouette, Calinski-Harabasz) across all models; bold marks the column maximum within each model section. Atomas and text-side embeddings are shown in Supplementary Figs.~5--7.}
\label{fig:tsne}
\end{figure}

Finally, we asked whether the downstream improvements are accompanied by a measurable reorganization of representation space. We analyzed molecule and text embeddings using CKA, clustering metrics and t-SNE visualization (Fig.~\ref{fig:tsne}); additional model-specific visualizations and scaffold-clustering metrics are provided in Supplementary Note~2, Supplementary Figs.~5--7 and Supplementary Table~4.

The primary effect of \methodname\ was to align molecular embeddings with fingerprint structure. Molecule--fingerprint CKA increased for all backbones (MolCA, $0.452 \rightarrow 0.688$; Atomas-B, $0.595 \rightarrow 0.784$; Atomas-L, $0.581 \rightarrow 0.788$), and MolCA showed a marked improvement in Tanimoto-cluster separation (Calinski--Harabasz, $15.6 \rightarrow 44.1$). Weight and fingerprint analyses in representation space mirrored the sensitivity results above: stronger structural alignment was beneficial only within a model-dependent regime, and fingerprint preferences followed the MolCA/Atomas encoder split.

The reorganization also propagated to the text side through contrastive training. Text--fingerprint CKA increased for MolCA, Atomas-B and Atomas-L, despite the CKA loss being applied to molecular representations. Molecule--text CKA remained stable or improved for MolCA and Atomas-L and only marginally decreased for Atomas-B, indicating that structural organization did not disrupt cross-modal alignment.

The same trend was visible under structural criteria not used directly by the training loss. When embeddings were grouped by Murcko scaffold identity, Calinski--Harabasz scores improved across all models (MolCA +341\%, Atomas-B +70\%, Atomas-L +73\%), and Silhouette became positive for every backbone. Together, these analyses suggest that \methodname\ does not merely imprint the exact fingerprint-similarity target, but induces a broader chemically organized representation space that helps explain the retrieval and property-prediction gains reported above.

\section{Discussion}

We have shown that a classical chemoinformatics prior can improve molecule--language representation learning when used as a training-time relational target. \methodname\ transfers fingerprint-based structural similarity into neural molecular embeddings through CKA regularization, without adding trainable parameters or requiring fingerprints at inference. Across MolCA and Atomas backbones, this prior improved molecule--text retrieval, shifted non-exact retrieval toward closer chemical analogues, increased average scaffold-split property performance, and improved or preserved performance in decoder-mediated generation ablations.

The main implication for chemical information modeling is that fingerprint similarity can be used as more than an external descriptor or post hoc retrieval score. By aligning pairwise relationships rather than concatenating descriptor features, \methodname\ preserves the role of fingerprints as an interpretable structural prior while allowing the deployed model to operate only with neural embeddings. This distinction matters for molecule--language systems: the same representation can support paired molecule--text retrieval and, when the exact target is removed, retrieve structurally closer analogues. The scaffold-split property results further suggest that this structural organization is not confined to ChEBI-20 retrieval, but can transfer to held-out molecular frameworks in downstream property prediction.

The results also clarify that structural-prior integration is a compatibility problem, not a single fixed recipe. Morgan fingerprints were most effective for the graph-based MolCA model, whereas MACCS keys were stronger for the SMILES-based Atomas variants, consistent with their different sensitivities to circular substructures and curated functional-group patterns~\cite{meyer2025reverse}. The regularization strength and CKA sample size were likewise model-dependent, with intermediate or capacity-adjusted settings performing best. Representation analyses supported this interpretation: molecule--fingerprint CKA increased across backbones, structural organization propagated to the text side through contrastive training, and scaffold-based clustering improved even though scaffold labels were not used by the loss. These observations suggest that the gains reflect a broader reorganization of molecule--language representations around chemically meaningful similarity, rather than simple fitting to one fingerprint table.

The current study focuses on two-dimensional fingerprint-derived priors in molecule--language models evaluated on retrieval, property prediction and generation benchmarks. The results do not establish that the same prior will be optimal for tasks dominated by three-dimensional conformation, target-specific binding mechanisms or prospective experimental validation. Generation results should also be interpreted as robustness evidence: the largest gains occur when downstream tasks directly consume the embedding geometry, whereas decoder-mediated tasks are improved or largely maintained rather than uniformly enhanced. Future work should therefore examine how to select or combine relational priors for different molecular encoders and chemical tasks, including priors that capture three-dimensional or target-aware relationships when those mechanisms are central to the application.

\FloatBarrier
\section{Methods}

\subsection{Overview of \methodname}

Standard contrastive learning for molecular representations relies on binary supervision: matched pairs are pulled together while unmatched pairs are pushed apart. However, this objective ignores the continuous spectrum of molecular similarity---two molecules may share significant substructural overlap yet be treated identically to completely unrelated pairs. \methodname\ (Structure-Aware Guided Alignment) addresses this by aligning \emph{relational structures} rather than individual representations. Using Centered Kernel Alignment (CKA), we measure how well the pairwise similarity patterns in the neural embedding space match those in the fingerprint space. This alignment is enforced through \lossname, an auxiliary regularization term added to the base contrastive loss at training time; no architectural changes to the underlying model are required, and \methodname\ introduces no additional trainable parameters. As a result, it can be applied as a drop-in to existing molecule-language frameworks regardless of their encoder type.

An overview of the proposed method is illustrated in Fig.~\ref{fig:method-overview}. Given a sample of molecules, we construct Gram matrices capturing pairwise similarities in both spaces and encourage their structural alignment through CKA-based regularization. The remainder of this section is organized as follows. Section~\ref{sec:method-data-curation} summarizes dataset curation, split usage and leakage-control procedures. Section~\ref{sec:method-representation-spaces} formalizes the two representation spaces and Section~\ref{sec:method-relational-alignment} introduces the CKA-based relational alignment objective that together define \methodname. Section~\ref{sec:method-preliminary} presents the preliminary analyses that validate its underlying assumptions. Section~\ref{sec:method-training-integration} describes the overall training objective and its integration with the two baseline models (MolCA and Atomas). Section~\ref{sec:method-baseline-pretraining} details the baseline pre-training setup, and Sections~\ref{sec:method-retrieval-eval}--\ref{sec:method-repr-analysis} describe the evaluation protocols used for cross-modal retrieval, structural-neighbour retrieval analysis, molecular property prediction, molecular generation and description, and representation space analysis.

\subsection{Data curation and dataset splits}
\label{sec:method-data-curation}

All analyses use public molecular datasets or source corpora. The molecule-language pre-training corpus was assembled from PubChem following the Atomas/MoleculeSTM data-collection pipeline. Briefly, raw SMILES-text pairs were merged by PubChem CID, filtered by minimum description length, and screened to remove SMILES that overlapped with ChEBI-20 and PCdes before pre-training. The resulting curated PubChem-derived corpus was used under matched conditions for each baseline and its \methodname\ counterpart, so comparisons isolate the effect of the structural-prior regularizer rather than differences in training data.

ChEBI-20 was used for molecule--text retrieval, structural-neighbour retrieval, molecular description generation and text-to-molecule generation. We retained the published train/validation/test split. Cross-modal retrieval, exclude-self structural-neighbour retrieval, Atomas text-to-molecule generation and Atomas description-generation metrics were evaluated on the validation split, whereas MolCA description generation was evaluated on the ChEBI-20 test split following the native MolCA evaluation recipe. In the exclude-self retrieval analysis, the paired ground-truth entry was removed from the candidate ranking before computing non-exact neighbours, but the same validation query set and model scores were otherwise used.

Property prediction used eight MoleculeNet classification datasets with benchmark scaffold splits. Molecular encoders were frozen, and only the prediction head was trained on property labels; no MoleculeNet labels were used during molecule-language pre-training or during \methodname\ regularization. The TDC ADMET datasets were used only for the diagnostic structure-property analysis of fingerprint similarity and were not used to train or tune the molecule-language models. SMILES strings, labels and split definitions were used as provided by the corresponding benchmark or by the native baseline preprocessing pipeline, and molecular fingerprints, Tanimoto similarities and validity checks were computed with RDKit as described below.

\subsection{Molecular representation spaces}
\label{sec:method-representation-spaces}

Given a sample of $n$ molecules $\mathcal{S} = \{s_i\}_{i=1}^{n}$, where each $s_i$ is a SMILES string, our method operates on two representation spaces. We refer to $n$ as the \emph{CKA sample size}; the concrete construction of $\mathcal{S}$ depends on the base model and is described in Section~\ref{sec:method-training-integration}.

\paragraph{Latent space ($\mathcal{Z}^{(h)}$).}
A neural encoder $f_\theta$ maps each molecule to a $d$-dimensional continuous embedding $\mathbf{z}^{(h)}_i \in \mathbb{R}^d$, forming a matrix $Z^{(h)} \in \mathbb{R}^{n \times d}$. This space captures task-adaptive neural representations learned through contrastive or multi-modal objectives.

\paragraph{Fingerprint space ($\mathcal{Z}^{(f)}$).}
Each molecule is also represented by a fixed, domain-informed binary fingerprint $\mathbf{z}^{(f)}_i \in \{0, 1\}^D$, forming a matrix $Z^{(f)} \in \{0,1\}^{n \times D}$. This space encodes structural priors grounded in classical chemoinformatics. We primarily use Morgan (extended-connectivity) fingerprints~\cite{rogers2010ecfp} with radius $r=2$ and bit length $D=2048$, computed via RDKit's~\cite{landrum_rdkit} \texttt{GetMorganGenerator}. In ablation studies we additionally consider two alternative fingerprint types: MACCS keys~\cite{durant2002mdl} ($D=167$), a set of 166 predefined structural patterns plus a reserved bit, generated via \texttt{MACCSkeys.GenMACCSKeys}; and RDKit topological fingerprints ($D=2048$) with path lengths from 1 to 7, generated via \texttt{GetRDKitFPGenerator}. These three fingerprint families capture complementary aspects of molecular structure---circular substructures, predefined pharmacophoric patterns, and linear topological paths, respectively.

\subsection{Relational structure alignment}
\label{sec:method-relational-alignment}

Rather than enforcing point-wise correspondence between representations, we preserve pairwise similarity relationships within the CKA sample set $\mathcal{S}$.

\paragraph{Gram matrix construction.}
In the fingerprint space, we utilize Tanimoto similarity:
\begin{equation}
\label{eq:tanimoto-kernel}
    K^{(f)}_{ij} = \frac{\mathbf{z}^{(f)}_i \cdot \mathbf{z}^{(f)}_j}{\|\mathbf{z}^{(f)}_i\|_1 + \|\mathbf{z}^{(f)}_j\|_1 - \mathbf{z}^{(f)}_i \cdot \mathbf{z}^{(f)}_j}
\end{equation}
In the latent space, we compute a linear Gram matrix using $L_2$-normalized embeddings, which is equivalent to cosine similarity between the two embeddings:
\begin{equation}
    K^{(h)}_{ij} = \tilde{\mathbf{z}}^{(h)\top}_i \tilde{\mathbf{z}}^{(h)}_j, \quad \text{where } \tilde{\mathbf{z}}^{(h)}_i = \frac{\mathbf{z}^{(h)}_i}{\|\mathbf{z}^{(h)}_i\|_2}.
\end{equation}

\paragraph{Structural alignment via CKA.}
We employ Centered Kernel Alignment (CKA)~\cite{kornblith2019cka} to measure alignment between the two Gram matrices. Intuitively, CKA evaluates how consistently the pairwise similarity pattern in one representation space matches that in the other, and is invariant to orthogonal transformations and isotropic scaling of the embeddings. To obtain reliable alignment estimates at finite sample sizes, we adopt the unbiased estimator~\cite{song2012feature}. For a Gram matrix $K \in \mathbb{R}^{n \times n}$, the unbiased centering operation is:
\begin{equation}
\label{eq:unbiased-centering}
    \tilde{K}_{ij} =
    \begin{cases}
        K_{ij} - \frac{1}{n-2}\sum_{k}K_{ik} - \frac{1}{n-2}\sum_{k}K_{kj} + \frac{1}{(n-1)(n-2)}\sum_{k,l}K_{kl} & \text{if } i \neq j \\
        0 & \text{if } i = j
    \end{cases}
\end{equation}
The \lossname\ is then defined as:
\begin{equation}
    \mathcal{L}_{\text{SAGA}} = 1 - \frac{\langle \tilde{K}^{(h)}, \tilde{K}^{(f)} \rangle_F}{\|\tilde{K}^{(h)}\|_F \cdot \|\tilde{K}^{(f)}\|_F}
\end{equation}
where $\langle \cdot, \cdot \rangle_F$ denotes the Frobenius inner product.

\subsection{Validation analyses for structural priors and CKA}
\label{sec:method-preliminary}

\paragraph{Structure-property alignment.}
For each of the 20 TDC ADMET datasets, we generated structurally stratified molecule pairs as follows. Let $\text{sim}(i,j) = \text{Tanimoto}(\mathbf{z}^{(f)}_i, \mathbf{z}^{(f)}_j)$ denote the pairwise Tanimoto similarity computed on Morgan fingerprints (radius~=~2, 2048 bits). We partitioned pairs into $K=10$ equal-width similarity bins,
\begin{equation}
    \mathcal{B}_k = \bigl\{(i,j) : i<j,\ \text{sim}(i,j) \in \bigl[\tfrac{k-1}{K},\tfrac{k}{K}\bigr)\bigr\}, \quad k = 1,\ldots,K,
\end{equation}
and sampled up to 500 pairs per bin to counteract the heavy skew toward low-similarity pairs. For each bin we aggregated the alignment signal as
\begin{equation}
    \bar{d}_k = \operatorname{median}\!\bigl\{|y_i - y_j|:(i,j)\in\mathcal{B}_k\bigr\} \quad \text{(regression)},
\end{equation}
\begin{equation}
    p_k = \frac{1}{|\mathcal{B}_k|}\sum_{(i,j)\in\mathcal{B}_k}\mathbb{1}[y_i=y_j] \quad \text{(classification)},
\end{equation}
and computed the Spearman rank correlation $\rho$ between bin centers $c_k=\tfrac{k-0.5}{K}$ and the aggregate ($\bar{d}_k$ or $p_k$). Effect magnitudes were quantified by contrasting high- and low-similarity subpopulations using Cliff's $\delta$ for regression,
\begin{equation}
    \delta = \frac{\#\{(x,y):x>y\} - \#\{(x,y):x<y\}}{n_1 n_2},
\end{equation}
and Cohen's $h$ for classification,
\begin{equation}
    h = 2\!\left(\arcsin\!\sqrt{p_1} - \arcsin\!\sqrt{p_2}\right),
\end{equation}
where $x,y$ are property differences drawn from the two populations and $p_1,p_2$ the corresponding label-agreement proportions.

\paragraph{CKA validity.}
To verify that CKA meaningfully reflects structure-text alignment, we applied the unbiased estimator of Eq.~\eqref{eq:unbiased-centering} to matched and shuffled ChEBI-20 pairs. Molecular Gram matrices $K^{(f)}$ were built with the Tanimoto kernel in Eq.~\eqref{eq:tanimoto-kernel} on Morgan fingerprints, while text Gram matrices used a linear kernel, $K^{(t)}_{ij} = \mathbf{t}_i^\top \mathbf{t}_j$, over description embeddings $\mathbf{t}_i \in \mathbb{R}^{1536}$ obtained from OpenAI's \texttt{text-embedding-3-small} model~\cite{neelakantan2022text}. CKA values were evaluated at sample sizes $n \in \{100, 200, 500\}$ and averaged across 4 random seeds, comparing the \emph{Original} (correctly matched) and \emph{Shuffled} (randomly permuted descriptions) conditions.

\subsection{Training objective and model integration}
\label{sec:method-training-integration}

The final training objective integrates the base contrastive loss $\mathcal{L}_{\text{base}}$ with the proposed regularization:
\begin{equation}
    \mathcal{L}_{\text{total}} = \mathcal{L}_{\text{base}} + \lambda \cdot \mathcal{L}_{\text{SAGA}}
\end{equation}
where $\lambda$ controls the trade-off between instance-level discriminative learning and manifold-level structural alignment. The concrete construction of the CKA sample set $\mathcal{S}$, however, differs between the two baseline models due to their underlying training architectures.

\paragraph{MolCA.}
MolCA performs standard mini-batch training without a memory queue. At each training step, the CKA sample set $\mathcal{S}$ is simply the current mini-batch $\mathcal{B}$, so the CKA sample size equals the batch size ($n = |\mathcal{B}|$) and $\mathcal{L}_{\text{SAGA}}$ is computed directly within the batch.

\paragraph{Atomas.}
Atomas employs MoCo-style~\cite{he2020momentum} contrastive learning with a momentum-updated queue for negative sampling. Because large models such as Atomas cannot afford arbitrarily large training batches, we decouple the CKA sample set from the training batch: we maintain a separate molecular queue that stores fingerprint and corresponding latent representations alongside the existing feature queue, and compute $\mathcal{L}_{\text{SAGA}}$ over $n$ samples drawn from this queue. This allows the CKA sample size $n$ to be tuned independently of the training batch size, which remains fixed at 16 following the original implementation.

\subsection{Baseline models and pre-training}
\label{sec:method-baseline-pretraining}

\paragraph{Baseline models.}
We evaluate \methodname\ on two representative molecular representation learning frameworks. MolCA~\cite{liu2023molca} is a graph-language modeling framework that employs a Q-Former~\cite{li2023blip2} to bridge 2D molecular graphs with language models. Atomas~\cite{atomas2024} is a hierarchical adaptive alignment framework that unifies SMILES and text representations through a shared encoder architecture; we evaluate both Base and Large variants.

\paragraph{Pre-training dataset.}
We assemble a unified pre-training corpus directly from PubChem by following the data collection pipeline of Atomas~\cite{atomas2024}, which itself adapts the MoleculeSTM framework~\cite{liu2023moleculestm}. Starting from 408{,}250 raw SMILES-text pairs, we merge entries sharing the same PubChem CID, apply a minimum description length filter (18 characters), and remove all molecules whose SMILES overlap with the downstream evaluation datasets (ChEBI-20 and PCdes) to prevent data leakage. The resulting dataset contains 34{,}365 unique molecule-text pairs and is used to pre-train both MolCA and Atomas, under both their original training objective (baseline) and with the added \methodname\ regularizer, so that any performance difference can be attributed solely to \methodname.

\paragraph{Training configuration.}
All pre-training runs are executed using the official open-source implementations of MolCA~\cite{liu2023molca} and Atomas~\cite{atomas2024}, with the \methodname\ loss integrated into each training pipeline. For the main retrieval, property-prediction and representation analyses, \methodname\ uses Morgan fingerprints with $\lambda=100$ unless otherwise specified. MolCA computes CKA over mini-batches of 64 molecules, so the CKA sample size is $n=64$. Atomas uses a training batch size of 16 and computes CKA over a separate fingerprint queue of $n=128$ samples. Generation-specific checkpoint provenance is described in Section~\ref{sec:method-generation-eval}, and ablation experiments vary $\lambda$, fingerprint type and CKA sample size as reported in Supplementary Note~1. We otherwise follow the default hyperparameter settings of each baseline (optimizer, learning rate schedule, and base contrastive loss).

\subsection{Cross-modal retrieval evaluation}
\label{sec:method-retrieval-eval}

\paragraph{Dataset.}
We evaluate cross-modal retrieval on ChEBI-20~\cite{edwards2021text2mol}, which contains approximately 33{,}000 molecule-description pairs under an 80\%/10\%/10\% train/validation/test split. Retrieval is performed over the validation split, yielding a candidate pool of approximately 3{,}300 pairs.

\paragraph{Retrieval protocol.}
We follow the native scoring protocol of each baseline: molecular and text embeddings are extracted from the respective encoders, and a similarity matrix is constructed between all query-candidate pairs over the full validation pool. For Atomas, embeddings are $L_2$-normalized and scored by dot product, equivalent to cosine similarity. For MolCA, each text embedding is scored against the set of eight Q-Former query tokens of a molecule via dot product, and the maximum response across query tokens is taken as the final match score. Retrieval is evaluated in both directions: \emph{Molecule-to-Text} uses each molecule as the query to rank all text candidates, and \emph{Text-to-Molecule} uses each text as the query to rank all molecule candidates. We report standard retrieval metrics,
\begin{equation}
    \text{R@}K = \frac{1}{|\mathcal{Q}|}\sum_{q \in \mathcal{Q}} \mathbb{1}\!\left[\text{rank}(q) \leq K\right], \qquad \text{MRR} = \frac{1}{|\mathcal{Q}|}\sum_{q \in \mathcal{Q}} \frac{1}{\text{rank}(q)},
\end{equation}
where $\mathcal{Q}$ is the set of queries and $\text{rank}(q)$ is the position of the ground-truth target of query $q$ in the sorted candidate list. We report R@$K$ for $K \in \{1, 5, 10\}$.

\subsection{Structural-neighbour retrieval analysis}
\label{sec:method-structural-neighbour}

To test whether retrieval gains correspond to chemically closer non-exact neighbours, we performed an exclude-self structural-neighbour analysis on the ChEBI-20 validation split using the same model scores as in the retrieval evaluation. For each query, the paired ground-truth candidate was removed from the ranking, and the next top-ranked candidates were treated as non-exact neighbours. In Text-to-Molecule retrieval, Tanimoto similarity was computed between the molecule paired with the query text and each retrieved molecule. In Molecule-to-Text retrieval, each retrieved text was mapped back to its paired molecule and compared with the query molecule.

All structural similarities in this analysis used Morgan fingerprints with radius 2 and 2,048 bits. For each query and cutoff $K \in \{1,2,3\}$, we computed the mean Tanimoto similarity of the top-$K$ non-exact neighbours, and aggregate displays report the mean and per-query distributions of these values. Paired differences between \methodname\ and the baseline were assessed with a one-sided Wilcoxon signed-rank test, using the alternative hypothesis that \methodname\ retrieves structurally closer neighbours. We also report the fraction of queries for which the \methodname\ top-3 mean Tanimoto similarity exceeds the corresponding baseline value.

For the per-query scatter analysis, each point corresponds to one validation query, with the horizontal and vertical coordinates given by the baseline and \methodname\ top-3 mean Tanimoto similarities, respectively. Point colour encodes the signed, log-scaled improvement in the exact target rank. Representative retrieval examples were selected from queries in which \methodname\ retrieved high-Tanimoto neighbours while the baseline retrieved low-Tanimoto neighbours, and are used only as illustrative examples. Description-level text-overlap displays in the Supplementary Information use the same top-3 non-exact neighbours. Red annotations mark structural-class terms shared with the query description; these annotations are qualitative aids and are not used in the quantitative Tanimoto or statistical analyses.

\subsection{Molecular property prediction evaluation}
\label{sec:method-property-eval}

\paragraph{Datasets.}
We use eight classification datasets from the MoleculeNet benchmark~\cite{wu2018moleculenet}: BBBP, HIV, and BACE are single-label binary classification, while Tox21 (12 labels), ToxCast (617), SIDER (27), ClinTox (2), and MUV (17) are multi-label tasks in which each label corresponds to an independent bioassay endpoint. All datasets follow the scaffold split recommended by the benchmark, which provides a more realistic assessment of out-of-distribution generalization than random splits.

\paragraph{Embedding extraction.}
To feed a single fixed-dimensional vector into the probe, we extract a pooled molecular embedding from each frozen encoder following the pooling convention of the original model. For MolCA, we mean-pool over the eight Q-Former query tokens to obtain a 768-dimensional graph embedding. For Atomas, a learned linear layer aggregates the SMILES token sequence into a single vector which is then $L_2$-normalized, yielding a 768-dimensional embedding for Atomas-B and a 1024-dimensional embedding for Atomas-L (matching the native hidden size of the underlying MolT5~\cite{edwards2022molt5} encoder).

\paragraph{Linear probing protocol.}
To evaluate the quality of the learned representations independently of the pre-training objective, we freeze the molecular encoder and train only a lightweight 2-layer MLP classification head on top of the pooled embedding. The head applies a linear projection to a 256-dimensional hidden representation, followed by LayerNorm, a GELU activation, dropout with $p=0.1$, and a final linear layer mapping to the task output. No part of the encoder is fine-tuned; only the MLP parameters are updated.

We train for 25 epochs using AdamW (initial learning rate $10^{-3}$, weight decay $0.01$) with a \texttt{ReduceLROnPlateau} scheduler (factor 0.5, patience 5, minimum learning rate $10^{-4}$) that monitors validation ROC-AUC. The batch size is 32. We use binary cross-entropy with logits as the loss for both single-label and multi-label tasks, computed independently per label. Performance is reported as mean $\pm$ standard deviation of ROC-AUC over 5 random seeds; for multi-label datasets, the per-label AUROCs are averaged to obtain the reported score.

\subsection{Molecular generation and description evaluation}
\label{sec:method-generation-eval}

Generation experiments use the native fine-tuning and decoding protocols of the corresponding backbone. Text-to-molecule generation is evaluated only for Atomas, whose original framework supports SMILES generation from text, whereas molecule-to-text description generation is evaluated for MolCA and Atomas. In Atomas, the \texttt{genmol} and \texttt{gentext} tasks are fine-tuned separately from the corresponding task-specific baseline or \methodname-pretrained initialization. The values reported in Table~\ref{tab:generation-description} are single-run results.

For the MolCA description-generation runs, we use the ChEBI-20 train/validation/test split and fine-tune the Galactica-1.3B language model with LoRA adapters following the original MolCA stage-2 recipe. The run reported in Table~\ref{tab:generation-description} uses micro-batches of 16 with gradient accumulation of 4, corresponding to an effective batch size of 64, for 100 fine-tuning epochs. Captions are decoded on the ChEBI-20 test split (3,300 molecule-description pairs) with beam search using five beams and no sampling. We report BLEU-2, BLEU-4, ROUGE-1, ROUGE-2, and ROUGE-L, multiplied by 100, using the original MolCA caption-evaluation implementation.

For the Atomas description-generation runs, we fine-tune the original Atomas Base and Large models on ChEBI-20 with the \texttt{gentext} objective, using the SMILES encoder output as the conditioning signal for text generation. Fine-tuning itself follows the native Atomas recipe and does not add a fingerprint-based CKA loss; the \methodname\ condition is instead initialized from the corresponding fingerprint-CKA-pretrained checkpoint. The description-generation \methodname\ checkpoint uses Morgan fingerprints with a fingerprint queue size of 128 and $\lambda=1000$ during pre-training. Atomas caption metrics are computed on the ChEBI-20 validation split with the original Atomas text-translation evaluator, using SciBERT tokenization for BLEU-2/4 and ROUGE-1/2/L F-measures. These metrics are multiplied by 100 in Table~\ref{tab:generation-description}.

For Atomas text-to-molecule generation, Base and Large variants are fine-tuned separately with the native \texttt{genmol} objective and evaluated on the ChEBI-20 validation split. The \methodname-pretrained initialization uses Morgan fingerprints with a fingerprint queue size of 128 and $\lambda=100$; this setting is distinct from the Atomas description-generation checkpoint. We report BLEU, exact-match rate, Levenshtein distance, MACCS, RDKit and Morgan fingerprint Tanimoto similarity (FTS), and chemical validity. BLEU is computed over character-tokenized SMILES strings. Exact match is computed after conversion of generated and reference SMILES to InChI strings, Levenshtein is the mean SMILES edit distance, chemical validity is the fraction of generated SMILES parsed by RDKit, and FTS metrics are mean Tanimoto similarities between generated and reference molecules using the corresponding fingerprint family. Bounded metrics are multiplied by 100 in Table~\ref{tab:generation-description}, whereas Levenshtein distance is reported on its raw scale.

\subsection{Representation space analysis}
\label{sec:method-repr-analysis}

To assess how \methodname\ reshapes the learned embedding space, we employ two complementary clustering schemes that capture different aspects of molecular structure. For \emph{Tanimoto-based clustering}, we compute pairwise Tanimoto distances from Morgan fingerprints and apply agglomerative hierarchical clustering with Ward linkage, cutting the dendrogram to yield 10 clusters. This data-driven grouping reflects the same fingerprint similarity structure used during training and thus serves as a direct measure of how well the learned representations preserve the intended structural prior. For \emph{scaffold-based clustering}, we extract Murcko scaffolds~\cite{bemis1996properties}---core ring systems obtained by removing side chains---and retain the 10 most frequent scaffolds as cluster labels, excluding molecules belonging to rarer scaffolds. Because scaffold identity is not used during training, this grouping provides an independent assessment of whether structural organization generalizes beyond the training signal. Scaffold frequencies in the evaluation set are heavily long-tailed, with several of the 10 retained scaffolds containing fewer than 20 molecules; such small clusters introduce instability in Silhouette and Calinski--Harabasz estimation. For quantitative scaffold-based metrics (Supplementary Table~4) we therefore retain only scaffolds with at least 20 samples, while the t-SNE visualization in Fig.~\ref{fig:tsne}d continues to display the top~10 scaffolds for completeness. Embeddings are projected to two dimensions via t-SNE~\cite{van2008visualizing} (perplexity~=~30) for visualization only; all quantitative metrics are computed in the original high-dimensional embedding space.

We quantify the representation space along two axes: alignment between representation spaces via CKA, and clustering quality under a given label assignment via Silhouette and Calinski--Harabasz indices. CKA is computed with the same Gram matrix construction as in the training objective (Section~\ref{sec:method-relational-alignment}): a Tanimoto kernel for fingerprints (Eq.~\ref{eq:tanimoto-kernel}) and a linear kernel over $L_2$-normalized embeddings for neural representations. For post-hoc analysis we use the biased HSIC estimator rather than the unbiased variant of Eq.~\eqref{eq:unbiased-centering}, as biased centering is the standard convention for CKA reporting~\cite{kornblith2019cka} and small-sample variance is less of a concern in the offline evaluation regime.

\paragraph{Silhouette coefficient.}
For each sample $i$ with cluster label $c_i$, let $a(i)$ denote the mean distance to all other samples in the same cluster and $b(i)$ the mean distance to all samples in the nearest neighboring cluster:
\begin{equation}
s(i) = \frac{b(i) - a(i)}{\max\{a(i),\, b(i)\}}, \quad s(i) \in [-1, 1].
\end{equation}
The Silhouette score is the average $\bar{s} = \frac{1}{n}\sum_{i=1}^n s(i)$; positive values indicate that samples lie closer to their own cluster than to neighboring clusters, while negative values indicate potential misassignment.

\paragraph{Calinski--Harabasz index.}
The Calinski--Harabasz index measures the ratio of between-cluster to within-cluster dispersion for $k$ clusters over $n$ samples:
\begin{equation}
\mathrm{CH} = \frac{\operatorname{tr}(\bm{B}_k)}{\operatorname{tr}(\bm{W}_k)} \cdot \frac{n - k}{k - 1},
\end{equation}
where $\bm{B}_k$ and $\bm{W}_k$ are the between- and within-cluster scatter matrices. Higher values indicate denser, better-separated clusters.

\section*{Supporting Information}

Supporting Information: Supplementary methods, sensitivity analyses, representation-space analyses, molecular generation and description details, retrieval case-study displays, and supplementary figures and tables (PDF).

\section*{Acknowledgments}

This research was supported by a grant of the Korea Machine Learning Ledger Orchestration for Drug Discovery Project (K-MELLODDY), funded by the Ministry of Health \& Welfare and Ministry of Science and ICT, Republic of Korea (grant number: RS-2024-00459964).

AI-assisted tools were used for language editing and translation support during manuscript preparation. The authors reviewed and are responsible for all manuscript content.

\section*{Author contributions}

All authors contributed to the conceptualization of the study, data analysis, and manuscript writing. D.L. implemented the method and conducted the experiments. S.P. participated in project discussions and provided feedback. M.C. and S.Y. jointly supervised the research. All authors reviewed and approved the final manuscript.

\section*{Data Availability Statement}

All source datasets analysed in this study are publicly available. The 20 ADMET benchmark datasets were obtained from the Therapeutics Data Commons (TDC, \url{https://tdcommons.ai})\cite{huang2021tdc}. The pre-training corpus was assembled from PubChem following the data collection pipeline of Atomas\cite{atomas2024}; processing details are described in the Methods. Cross-modal retrieval and generation experiments used ChEBI-20\cite{edwards2021text2mol}, and molecular property prediction used MoleculeNet\cite{wu2018moleculenet}. Text embeddings for the CKA analysis were obtained from OpenAI's \texttt{text-embedding-3-small} API\cite{neelakantan2022text} (accessed December 2025). Processed outputs and source data underlying the figures and tables will be made available to editors and reviewers through a private reviewer-access repository at \url{https://github.com/February24-Lee/SAGA.git} during peer review and will be deposited in a public archive upon publication.

\section*{Code and Software Availability}

The custom code used to train models, run evaluations and reproduce the analysis figures and tables will be made available to editors and reviewers through a private reviewer-access repository at \url{https://github.com/February24-Lee/SAGA.git} during peer review. Upon publication, the repository will be made publicly available on GitHub and archived on Zenodo with a permanent DOI. The implementation builds on the official open-source releases of MolCA\cite{liu2023molca} and Atomas\cite{atomas2024}.

\section*{Conflicts of Interest}

The authors declare no competing financial interest.

\bibliography{references}

\providecommand{\latin}[1]{#1}
\makeatletter
\providecommand{\doi}
  {\begingroup\let\do\@makeother\dospecials
  \catcode`\{=1 \catcode`\}=2 \doi@aux}
\providecommand{\doi@aux}[1]{\endgroup\texttt{#1}}
\makeatother
\providecommand*\mcitethebibliography{\thebibliography}
\csname @ifundefined\endcsname{endmcitethebibliography}
  {\let\endmcitethebibliography\endthebibliography}{}
\begin{mcitethebibliography}{29}
\providecommand*\natexlab[1]{#1}
\providecommand*\mciteSetBstSublistMode[1]{}
\providecommand*\mciteSetBstMaxWidthForm[2]{}
\providecommand*\mciteBstWouldAddEndPuncttrue
  {\def\EndOfBibitem{\unskip.}}
\providecommand*\mciteBstWouldAddEndPunctfalse
  {\let\EndOfBibitem\relax}
\providecommand*\mciteSetBstMidEndSepPunct[3]{}
\providecommand*\mciteSetBstSublistLabelBeginEnd[3]{}
\providecommand*\EndOfBibitem{}
\mciteSetBstSublistMode{f}
\mciteSetBstMaxWidthForm{subitem}{(\alph{mcitesubitemcount})}
\mciteSetBstSublistLabelBeginEnd
  {\mcitemaxwidthsubitemform\space}
  {\relax}
  {\relax}

\bibitem[Willett \latin{et~al.}(1998)Willett, Barnard, and
  Downs]{willett1998similarity}
Willett,~P.; Barnard,~J.~M.; Downs,~G.~M. Chemical Similarity Searching.
  \emph{Journal of Chemical Information and Computer Sciences} \textbf{1998},
  \emph{38}, 983--996, DOI: \doi{10.1021/ci9800211}\relax
\mciteBstWouldAddEndPuncttrue
\mciteSetBstMidEndSepPunct{\mcitedefaultmidpunct}
{\mcitedefaultendpunct}{\mcitedefaultseppunct}\relax
\EndOfBibitem
\bibitem[van Tilborg \latin{et~al.}(2022)van Tilborg, Alenicheva, and
  Grisoni]{vantilborg2022cliffs}
van Tilborg,~D.; Alenicheva,~A.; Grisoni,~F. Exposing the Limitations of
  Molecular Machine Learning with Activity Cliffs. \emph{Journal of Chemical
  Information and Modeling} \textbf{2022}, \emph{62}, 5938--5951, DOI:
  \doi{10.1021/acs.jcim.2c01073}\relax
\mciteBstWouldAddEndPuncttrue
\mciteSetBstMidEndSepPunct{\mcitedefaultmidpunct}
{\mcitedefaultendpunct}{\mcitedefaultseppunct}\relax
\EndOfBibitem
\bibitem[Rogers and Hahn(2010)Rogers, and Hahn]{rogers2010ecfp}
Rogers,~D.; Hahn,~M. Extended-Connectivity Fingerprints. \emph{Journal of
  Chemical Information and Modeling} \textbf{2010}, \emph{50}, 742--754, DOI:
  \doi{10.1021/ci100050t}\relax
\mciteBstWouldAddEndPuncttrue
\mciteSetBstMidEndSepPunct{\mcitedefaultmidpunct}
{\mcitedefaultendpunct}{\mcitedefaultseppunct}\relax
\EndOfBibitem
\bibitem[Bajusz \latin{et~al.}(2015)Bajusz, R{\'a}cz, and
  H{\'e}berger]{bajusz2015tanimoto}
Bajusz,~D.; R{\'a}cz,~A.; H{\'e}berger,~K. Why is Tanimoto index an appropriate
  choice for fingerprint-based similarity calculations? \emph{Journal of
  Cheminformatics} \textbf{2015}, \emph{7}, 20, DOI:
  \doi{10.1186/s13321-015-0069-3}\relax
\mciteBstWouldAddEndPuncttrue
\mciteSetBstMidEndSepPunct{\mcitedefaultmidpunct}
{\mcitedefaultendpunct}{\mcitedefaultseppunct}\relax
\EndOfBibitem
\bibitem[David \latin{et~al.}(2020)David, Thakkar, Mercado, and
  Engkvist]{david2020molrep}
David,~L.; Thakkar,~A.; Mercado,~R.; Engkvist,~O. Molecular representations in
  {AI}-driven drug discovery: a review and practical guide. \emph{Journal of
  Cheminformatics} \textbf{2020}, \emph{12}, 56, DOI:
  \doi{10.1186/s13321-020-00460-5}\relax
\mciteBstWouldAddEndPuncttrue
\mciteSetBstMidEndSepPunct{\mcitedefaultmidpunct}
{\mcitedefaultendpunct}{\mcitedefaultseppunct}\relax
\EndOfBibitem
\bibitem[Atz \latin{et~al.}(2021)Atz, Grisoni, and Schneider]{atz2021geometric}
Atz,~K.; Grisoni,~F.; Schneider,~G. Geometric deep learning on molecular
  representations. \emph{Nature Machine Intelligence} \textbf{2021}, \emph{3},
  1023--1032, DOI: \doi{10.1038/s42256-021-00418-8}\relax
\mciteBstWouldAddEndPuncttrue
\mciteSetBstMidEndSepPunct{\mcitedefaultmidpunct}
{\mcitedefaultendpunct}{\mcitedefaultseppunct}\relax
\EndOfBibitem
\bibitem[Edwards \latin{et~al.}(2021)Edwards, Zhai, and
  Ji]{edwards2021text2mol}
Edwards,~C.; Zhai,~C.; Ji,~H. Text2Mol: Cross-Modal Molecule Retrieval with
  Natural Language Queries. Proceedings of the 2021 Conference on Empirical
  Methods in Natural Language Processing. 2021; pp 595--607\relax
\mciteBstWouldAddEndPuncttrue
\mciteSetBstMidEndSepPunct{\mcitedefaultmidpunct}
{\mcitedefaultendpunct}{\mcitedefaultseppunct}\relax
\EndOfBibitem
\bibitem[Yang \latin{et~al.}(2019)Yang, Swanson, Jin, Coley, Eiden, Gao,
  Guzman-Perez, Hopper, Kelley, Mathea, Palmer, Settels, Jaakkola, Jensen, and
  Barzilay]{yang2019chemprop}
Yang,~K. \latin{et~al.}  Analyzing Learned Molecular Representations for
  Property Prediction. \emph{Journal of Chemical Information and Modeling}
  \textbf{2019}, \emph{59}, 3370--3388, DOI:
  \doi{10.1021/acs.jcim.9b00237}\relax
\mciteBstWouldAddEndPuncttrue
\mciteSetBstMidEndSepPunct{\mcitedefaultmidpunct}
{\mcitedefaultendpunct}{\mcitedefaultseppunct}\relax
\EndOfBibitem
\bibitem[Liu \latin{et~al.}(2023)Liu, Nie, Wang, Lu, Qiao, Liu, Tang, Xiao, and
  Anandkumar]{liu2023moleculestm}
Liu,~S.; Nie,~W.; Wang,~C.; Lu,~J.; Qiao,~Z.; Liu,~L.; Tang,~J.; Xiao,~C.;
  Anandkumar,~A. Multi-modal molecule structure-text model for text-based
  retrieval and editing. \emph{Nature Machine Intelligence} \textbf{2023},
  \emph{5}, 1447--1457, DOI: \doi{10.1038/s42256-023-00759-6}\relax
\mciteBstWouldAddEndPuncttrue
\mciteSetBstMidEndSepPunct{\mcitedefaultmidpunct}
{\mcitedefaultendpunct}{\mcitedefaultseppunct}\relax
\EndOfBibitem
\bibitem[Zeng \latin{et~al.}(2022)Zeng, Yao, Liu, and Sun]{zeng2022bridging}
Zeng,~Z.; Yao,~Y.; Liu,~Z.; Sun,~M. A deep-learning system bridging molecule
  structure and biomedical text with comprehension comparable to human
  professionals. \emph{Nature Communications} \textbf{2022}, \emph{13}, 862,
  DOI: \doi{10.1038/s41467-022-28494-3}\relax
\mciteBstWouldAddEndPuncttrue
\mciteSetBstMidEndSepPunct{\mcitedefaultmidpunct}
{\mcitedefaultendpunct}{\mcitedefaultseppunct}\relax
\EndOfBibitem
\bibitem[Su \latin{et~al.}(2022)Su, Du, Yang, Zhou, Li, Rao, Sun, Lu, and
  Wen]{su2022momu}
Su,~B.; Du,~D.; Yang,~Z.; Zhou,~Y.; Li,~J.; Rao,~A.; Sun,~H.; Lu,~Z.;
  Wen,~J.-R. A Molecular Multimodal Foundation Model Associating Molecule
  Graphs with Natural Language. \emph{Preprint at
  https://arxiv.org/abs/2209.05481} \textbf{2022}, \relax
\mciteBstWouldAddEndPunctfalse
\mciteSetBstMidEndSepPunct{\mcitedefaultmidpunct}
{}{\mcitedefaultseppunct}\relax
\EndOfBibitem
\bibitem[Liu \latin{et~al.}(2023)Liu, Li, Luo, Fei, Cao, Kawaguchi, Wang, and
  Chua]{liu2023molca}
Liu,~Z.; Li,~S.; Luo,~Y.; Fei,~H.; Cao,~Y.; Kawaguchi,~K.; Wang,~X.;
  Chua,~T.-S. {MolCA}: Molecular Graph-Language Modeling with Cross-Modal
  Projector and Uni-Modal Adapter. Proceedings of the 2023 Conference on
  Empirical Methods in Natural Language Processing. 2023; pp 15623--15638, DOI:
  \doi{10.18653/v1/2023.emnlp-main.966}\relax
\mciteBstWouldAddEndPuncttrue
\mciteSetBstMidEndSepPunct{\mcitedefaultmidpunct}
{\mcitedefaultendpunct}{\mcitedefaultseppunct}\relax
\EndOfBibitem
\bibitem[Zhang \latin{et~al.}(2025)Zhang, Ye, Yuan, Han, Huang, Yao, Liu, and
  Rong]{atomas2024}
Zhang,~Y.; Ye,~G.; Yuan,~C.; Han,~B.; Huang,~L.-K.; Yao,~J.; Liu,~W.; Rong,~Y.
  Atomas: Hierarchical Adaptive Alignment on Molecule-Text for Unified Molecule
  Understanding and Generation. International Conference on Learning
  Representations (ICLR). 2025\relax
\mciteBstWouldAddEndPuncttrue
\mciteSetBstMidEndSepPunct{\mcitedefaultmidpunct}
{\mcitedefaultendpunct}{\mcitedefaultseppunct}\relax
\EndOfBibitem
\bibitem[Li \latin{et~al.}(2023)Li, Zhang, Min, Ma, Zhao, and Zeng]{li2023kpgt}
Li,~H.; Zhang,~R.; Min,~Y.; Ma,~D.; Zhao,~D.; Zeng,~J. A knowledge-guided
  pre-training framework for improving molecular representation learning.
  \emph{Nature Communications} \textbf{2023}, \emph{14}, 7568, DOI:
  \doi{10.1038/s41467-023-43214-1}\relax
\mciteBstWouldAddEndPuncttrue
\mciteSetBstMidEndSepPunct{\mcitedefaultmidpunct}
{\mcitedefaultendpunct}{\mcitedefaultseppunct}\relax
\EndOfBibitem
\bibitem[You \latin{et~al.}(2020)You, Chen, Sui, Chen, Wang, and
  Shen]{you2020graphcl}
You,~Y.; Chen,~T.; Sui,~Y.; Chen,~T.; Wang,~Z.; Shen,~Y. Graph Contrastive
  Learning with Augmentations. Advances in Neural Information Processing
  Systems. 2020\relax
\mciteBstWouldAddEndPuncttrue
\mciteSetBstMidEndSepPunct{\mcitedefaultmidpunct}
{\mcitedefaultendpunct}{\mcitedefaultseppunct}\relax
\EndOfBibitem
\bibitem[Wang \latin{et~al.}(2022)Wang, Wang, Cao, and
  Barati~Farimani]{wang2022molclr}
Wang,~Y.; Wang,~J.; Cao,~Z.; Barati~Farimani,~A. Molecular contrastive learning
  of representations via graph neural networks. \emph{Nature Machine
  Intelligence} \textbf{2022}, \emph{4}, 279--287, DOI:
  \doi{10.1038/s42256-022-00447-x}\relax
\mciteBstWouldAddEndPuncttrue
\mciteSetBstMidEndSepPunct{\mcitedefaultmidpunct}
{\mcitedefaultendpunct}{\mcitedefaultseppunct}\relax
\EndOfBibitem
\bibitem[Kornblith \latin{et~al.}(2019)Kornblith, Norouzi, Lee, and
  Hinton]{kornblith2019cka}
Kornblith,~S.; Norouzi,~M.; Lee,~H.; Hinton,~G. Similarity of Neural Network
  Representations Revisited. Proceedings of the 36th International Conference
  on Machine Learning. 2019; pp 3519--3529\relax
\mciteBstWouldAddEndPuncttrue
\mciteSetBstMidEndSepPunct{\mcitedefaultmidpunct}
{\mcitedefaultendpunct}{\mcitedefaultseppunct}\relax
\EndOfBibitem
\bibitem[Huang \latin{et~al.}(2021)Huang, Fu, Gao, Zhao, Roohani, Leskovec,
  Coley, Xiao, Sun, and Zitnik]{huang2021tdc}
Huang,~K.; Fu,~T.; Gao,~W.; Zhao,~Y.; Roohani,~Y.; Leskovec,~J.; Coley,~C.~W.;
  Xiao,~C.; Sun,~J.; Zitnik,~M. Therapeutics Data Commons: Machine Learning
  Datasets and Tasks for Drug Discovery and Development. Proceedings of Neural
  Information Processing Systems, NeurIPS Datasets and Benchmarks. 2021\relax
\mciteBstWouldAddEndPuncttrue
\mciteSetBstMidEndSepPunct{\mcitedefaultmidpunct}
{\mcitedefaultendpunct}{\mcitedefaultseppunct}\relax
\EndOfBibitem
\bibitem[Li \latin{et~al.}(2023)Li, Li, Savarese, and Hoi]{li2023blip2}
Li,~J.; Li,~D.; Savarese,~S.; Hoi,~S. BLIP-2: Bootstrapping Language-Image
  Pre-training with Frozen Image Encoders and Large Language Models.
  Proceedings of the 40th International Conference on Machine Learning (ICML).
  2023\relax
\mciteBstWouldAddEndPuncttrue
\mciteSetBstMidEndSepPunct{\mcitedefaultmidpunct}
{\mcitedefaultendpunct}{\mcitedefaultseppunct}\relax
\EndOfBibitem
\bibitem[He \latin{et~al.}(2020)He, Fan, Wu, Xie, and Girshick]{he2020momentum}
He,~K.; Fan,~H.; Wu,~Y.; Xie,~S.; Girshick,~R. Momentum Contrast for
  Unsupervised Visual Representation Learning. Proceedings of the IEEE/CVF
  Conference on Computer Vision and Pattern Recognition (CVPR). 2020; pp
  9729--9738\relax
\mciteBstWouldAddEndPuncttrue
\mciteSetBstMidEndSepPunct{\mcitedefaultmidpunct}
{\mcitedefaultendpunct}{\mcitedefaultseppunct}\relax
\EndOfBibitem
\bibitem[Wu \latin{et~al.}(2018)Wu, Ramsundar, Feinberg, Gomes, Geniesse,
  Pappu, Leswing, and Pande]{wu2018moleculenet}
Wu,~Z.; Ramsundar,~B.; Feinberg,~E.~N.; Gomes,~J.; Geniesse,~C.; Pappu,~A.~S.;
  Leswing,~K.; Pande,~V. {MoleculeNet}: a benchmark for molecular machine
  learning. \emph{Chemical Science} \textbf{2018}, \emph{9}, 513--530\relax
\mciteBstWouldAddEndPuncttrue
\mciteSetBstMidEndSepPunct{\mcitedefaultmidpunct}
{\mcitedefaultendpunct}{\mcitedefaultseppunct}\relax
\EndOfBibitem
\bibitem[Meyer \latin{et~al.}(2025)Meyer, Duigou, Gricourt, and
  Faulon]{meyer2025reverse}
Meyer,~P.; Duigou,~T.; Gricourt,~G.; Faulon,~J.-L. Reverse engineering
  molecules from fingerprints through deterministic enumeration and generative
  models. \emph{Journal of Cheminformatics} \textbf{2025}, \emph{17}, 157, DOI:
  \doi{10.1186/s13321-025-01074-5}\relax
\mciteBstWouldAddEndPuncttrue
\mciteSetBstMidEndSepPunct{\mcitedefaultmidpunct}
{\mcitedefaultendpunct}{\mcitedefaultseppunct}\relax
\EndOfBibitem
\bibitem[Landrum()]{landrum_rdkit}
Landrum,~G. {RDKit}: Open-source cheminformatics. \url{https://www.rdkit.org},
  Accessed: 2026\relax
\mciteBstWouldAddEndPuncttrue
\mciteSetBstMidEndSepPunct{\mcitedefaultmidpunct}
{\mcitedefaultendpunct}{\mcitedefaultseppunct}\relax
\EndOfBibitem
\bibitem[Durant \latin{et~al.}(2002)Durant, Leland, Henry, and
  Nourse]{durant2002mdl}
Durant,~J.~L.; Leland,~B.~A.; Henry,~D.~R.; Nourse,~J.~G. Reoptimization of
  {MDL} keys for use in drug discovery. \emph{Journal of Chemical Information
  and Computer Sciences} \textbf{2002}, \emph{42}, 1273--1280, DOI:
  \doi{10.1021/ci010132r}\relax
\mciteBstWouldAddEndPuncttrue
\mciteSetBstMidEndSepPunct{\mcitedefaultmidpunct}
{\mcitedefaultendpunct}{\mcitedefaultseppunct}\relax
\EndOfBibitem
\bibitem[Song \latin{et~al.}(2012)Song, Smola, Gretton, Bedo, and
  Borgwardt]{song2012feature}
Song,~L.; Smola,~A.; Gretton,~A.; Bedo,~J.; Borgwardt,~K. Feature selection via
  dependence maximization. \emph{Journal of Machine Learning Research}
  \textbf{2012}, \emph{13}, 1393--1434\relax
\mciteBstWouldAddEndPuncttrue
\mciteSetBstMidEndSepPunct{\mcitedefaultmidpunct}
{\mcitedefaultendpunct}{\mcitedefaultseppunct}\relax
\EndOfBibitem
\bibitem[Neelakantan \latin{et~al.}(2022)Neelakantan, Xu, Puri, Radford, Han,
  Tworek, Yuan, Tezak, Kim, Hallacy, \latin{et~al.}
  others]{neelakantan2022text}
Neelakantan,~A. \latin{et~al.}  Text and Code Embeddings by Contrastive
  Pre-Training. \emph{Preprint at https://arxiv.org/abs/2201.10005}
  \textbf{2022}, \relax
\mciteBstWouldAddEndPunctfalse
\mciteSetBstMidEndSepPunct{\mcitedefaultmidpunct}
{}{\mcitedefaultseppunct}\relax
\EndOfBibitem
\bibitem[Edwards \latin{et~al.}(2022)Edwards, Lai, Ros, Honke, Cho, and
  Ji]{edwards2022molt5}
Edwards,~C.; Lai,~T.; Ros,~K.; Honke,~G.; Cho,~K.; Ji,~H. Translation between
  Molecules and Natural Language. Proceedings of the 2022 Conference on
  Empirical Methods in Natural Language Processing. 2022; pp 375--413, DOI:
  \doi{10.18653/v1/2022.emnlp-main.26}\relax
\mciteBstWouldAddEndPuncttrue
\mciteSetBstMidEndSepPunct{\mcitedefaultmidpunct}
{\mcitedefaultendpunct}{\mcitedefaultseppunct}\relax
\EndOfBibitem
\bibitem[Bemis and Murcko(1996)Bemis, and Murcko]{bemis1996properties}
Bemis,~G.~W.; Murcko,~M.~A. The Properties of Known Drugs. 1. Molecular
  Frameworks. \emph{Journal of Medicinal Chemistry} \textbf{1996}, \emph{39},
  2887--2893, DOI: \doi{10.1021/jm9602928}\relax
\mciteBstWouldAddEndPuncttrue
\mciteSetBstMidEndSepPunct{\mcitedefaultmidpunct}
{\mcitedefaultendpunct}{\mcitedefaultseppunct}\relax
\EndOfBibitem
\bibitem[van~der Maaten and Hinton(2008)van~der Maaten, and
  Hinton]{van2008visualizing}
van~der Maaten,~L.; Hinton,~G. Visualizing Data using t-SNE. \emph{Journal of
  Machine Learning Research} \textbf{2008}, \emph{9}, 2579--2605\relax
\mciteBstWouldAddEndPuncttrue
\mciteSetBstMidEndSepPunct{\mcitedefaultmidpunct}
{\mcitedefaultendpunct}{\mcitedefaultseppunct}\relax
\EndOfBibitem
\end{mcitethebibliography}

\clearpage
\makeatletter
\setlength{\@fptop}{0pt plus 1fil}
\makeatother
\renewcommand{\theHfigure}{S\arabic{figure}}
\renewcommand{\theHtable}{S\arabic{table}}
\section*{Supplementary Information}

This document provides supplementary methods and analyses supporting the main manuscript, including structure--property alignment across TDC ADMET datasets, sensitivity analyses for the \methodname\ regularization settings, representation-space analyses, molecular generation and description details, structural-neighbour retrieval displays, and description-level retrieval case-study displays.

\setcounter{figure}{0}
\setcounter{table}{0}
\captionsetup[figure]{labelformat=empty,labelsep=none}
\captionsetup[table]{labelformat=empty,labelsep=none}

\begin{figure*}[!ht]
\centering
\setlength{\tabcolsep}{2pt}

\textbf{Regression Tasks}\\[0.5em]
\begin{tabular}{ccccc}
\includegraphics[width=0.19\textwidth]{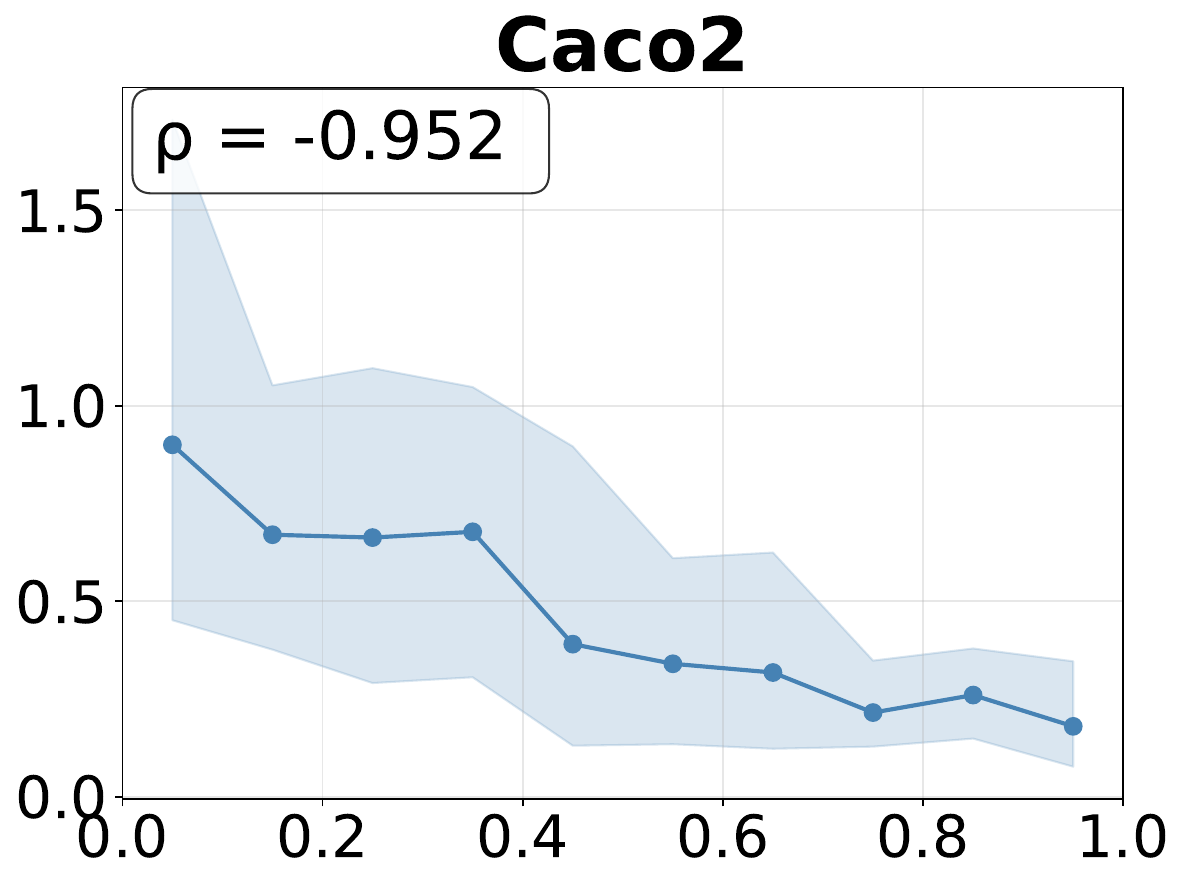} &
\includegraphics[width=0.19\textwidth]{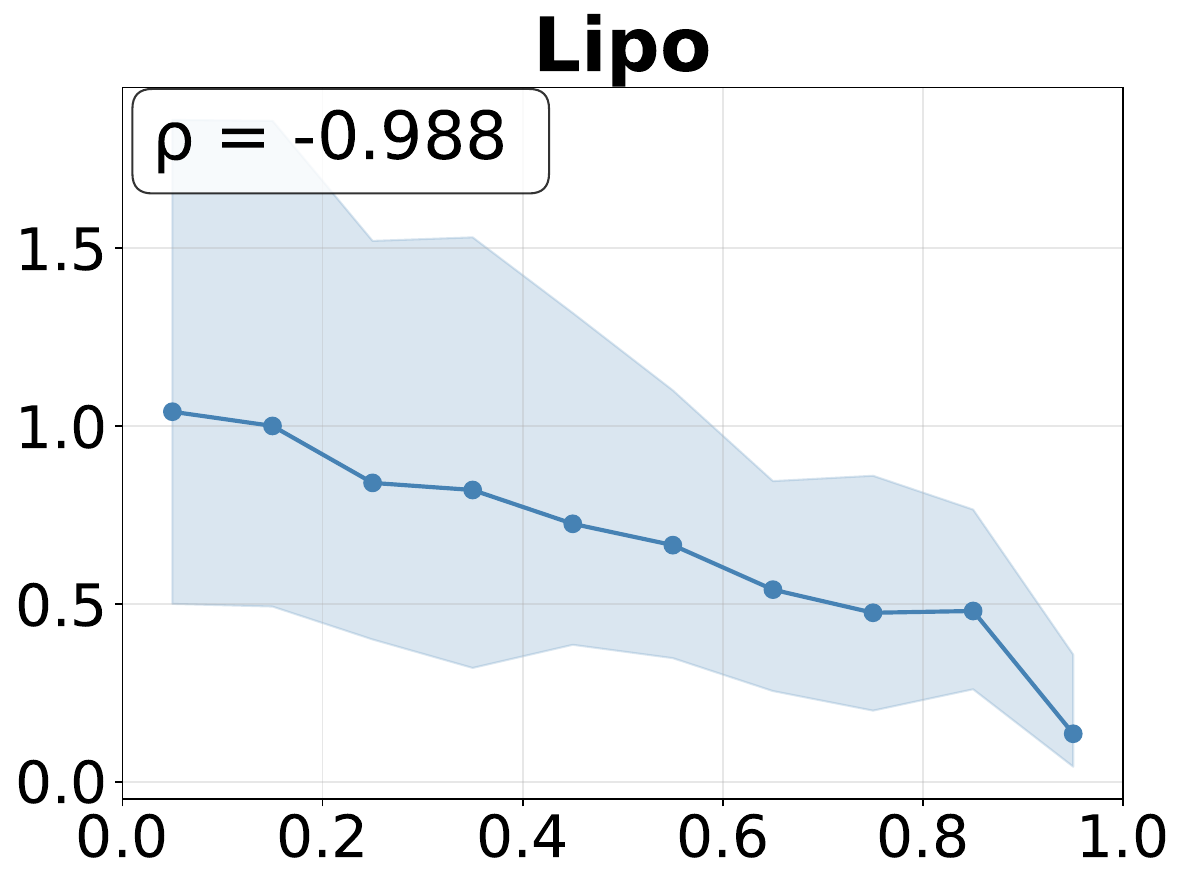} &
\includegraphics[width=0.19\textwidth]{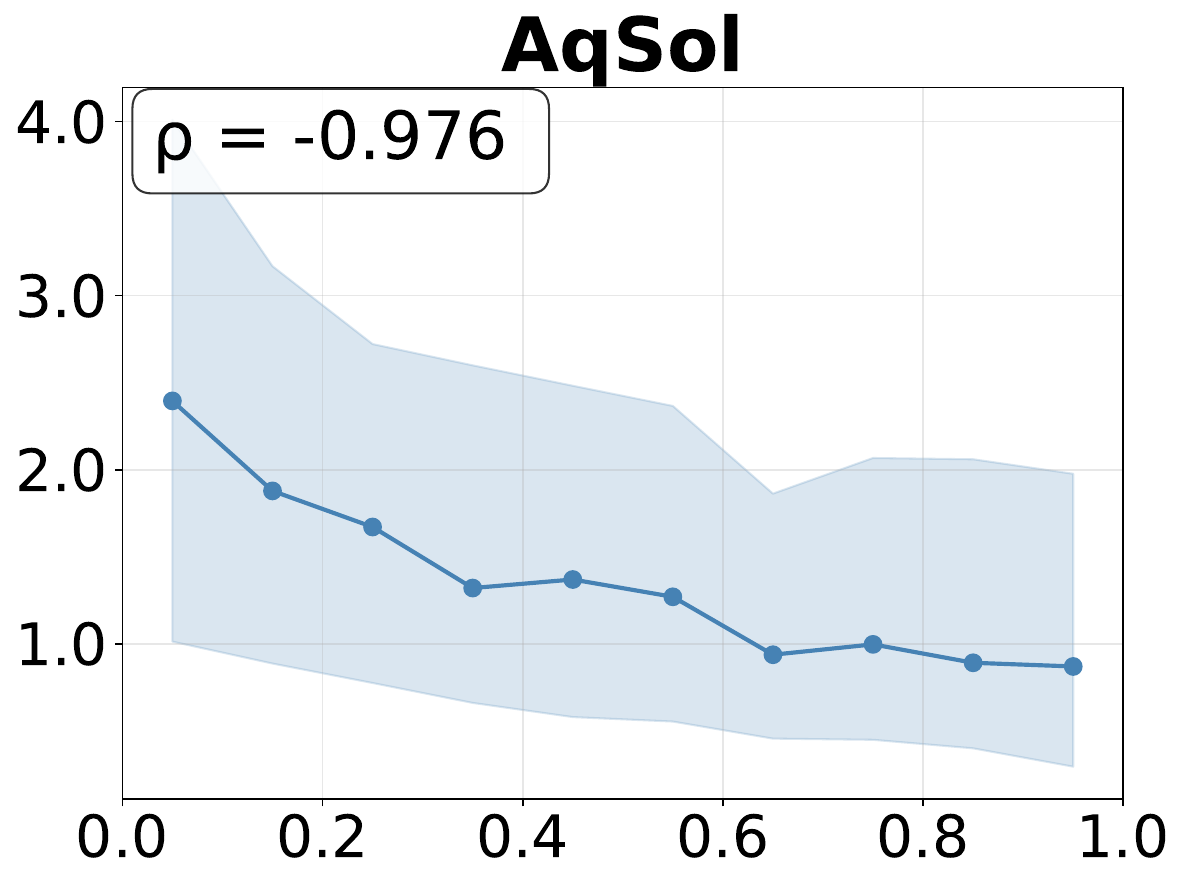} &
\includegraphics[width=0.19\textwidth]{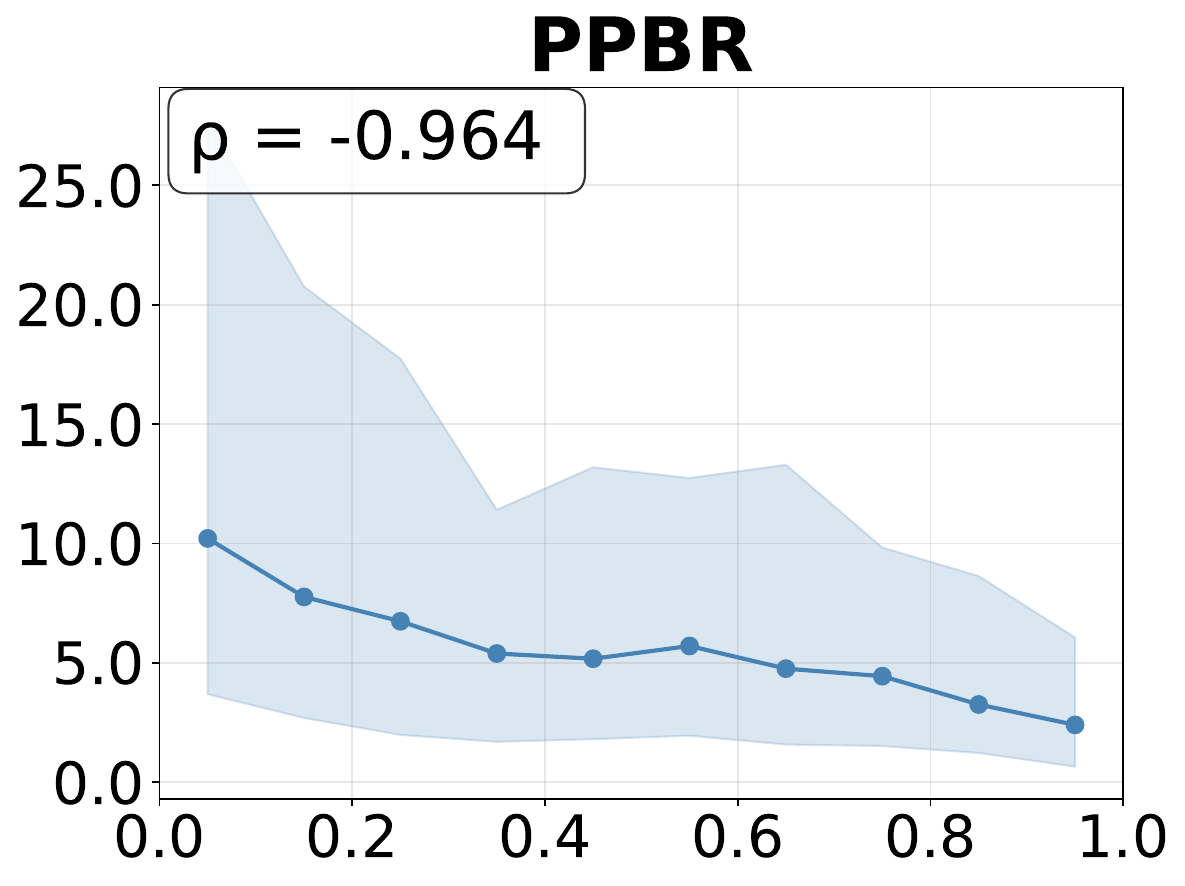} &
\includegraphics[width=0.19\textwidth]{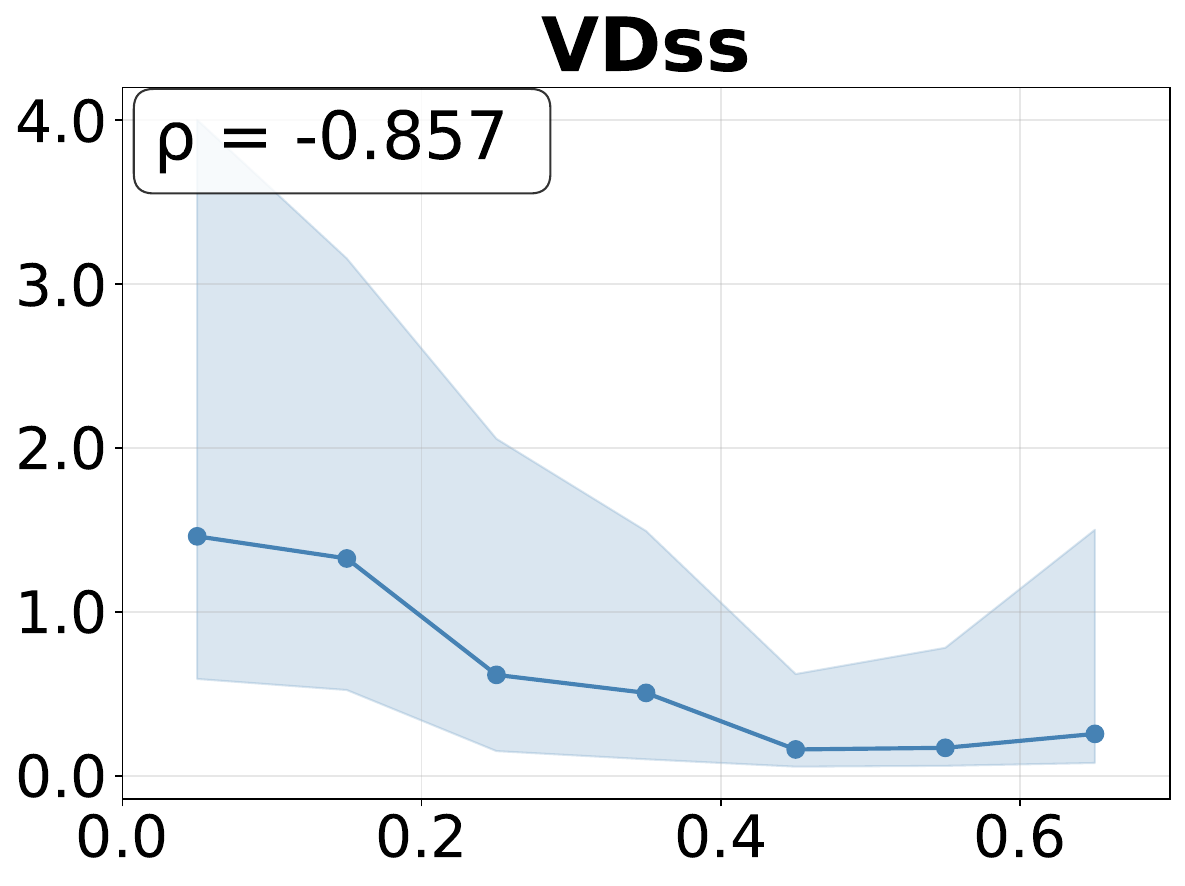} \\[0.5em]
\includegraphics[width=0.19\textwidth]{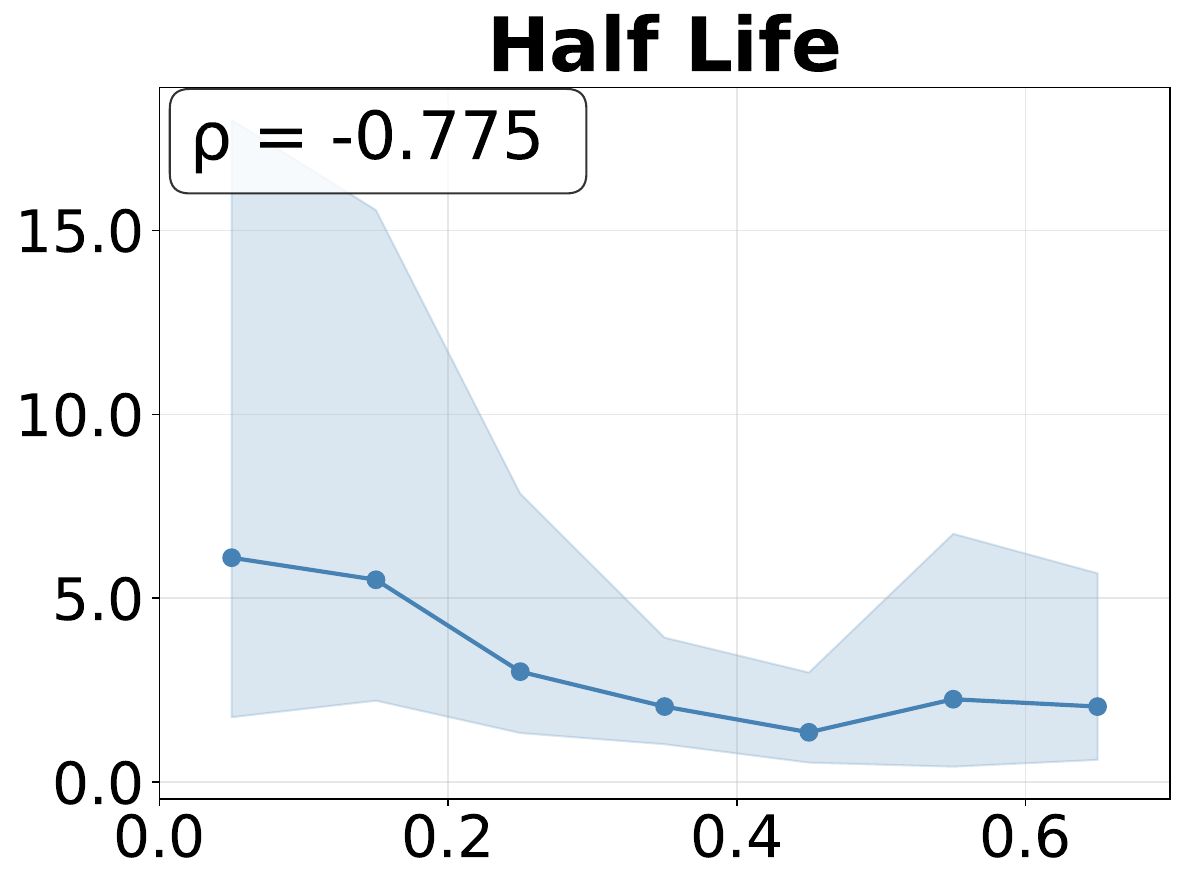} &
\includegraphics[width=0.19\textwidth]{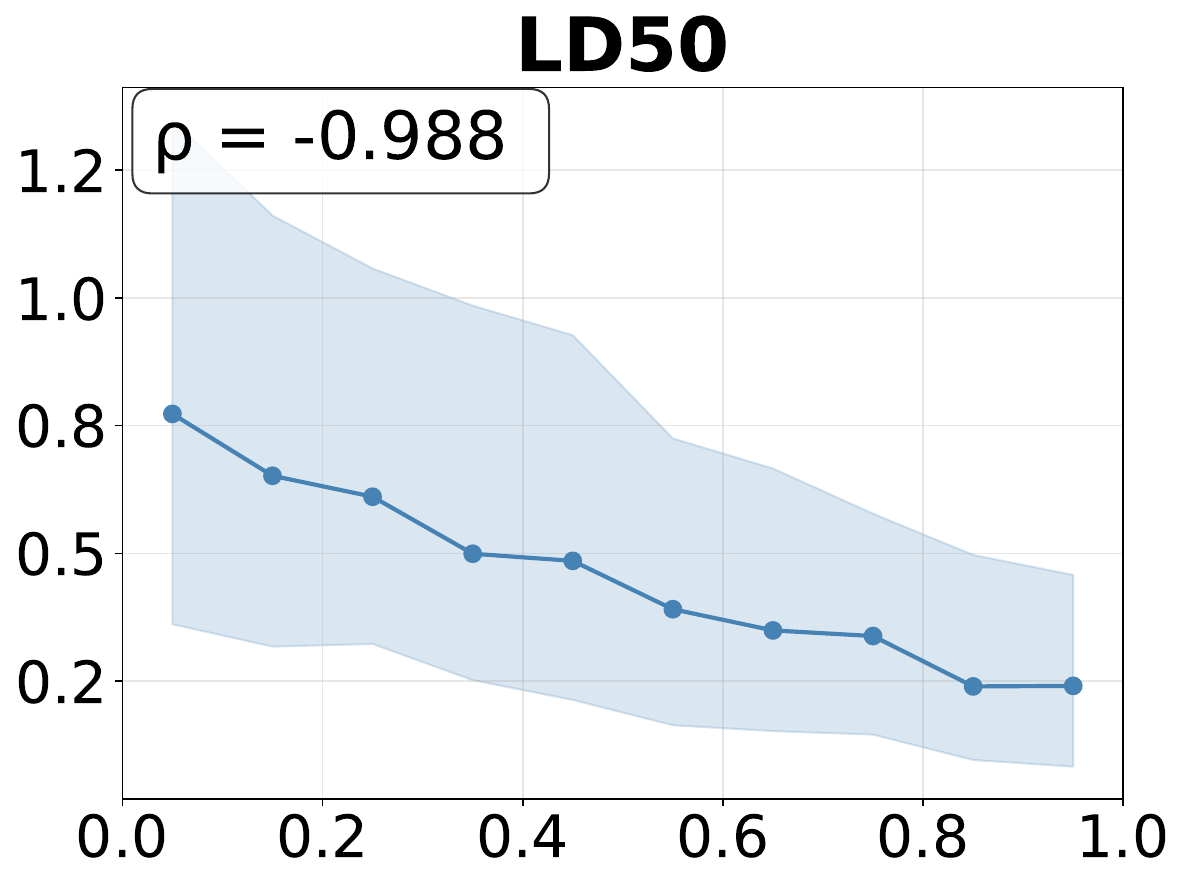} & & & \\
\end{tabular}

\vspace{1em}

\textbf{Binary Classification Tasks}\\[0.5em]
\begin{tabular}{ccccc}
\includegraphics[width=0.19\textwidth]{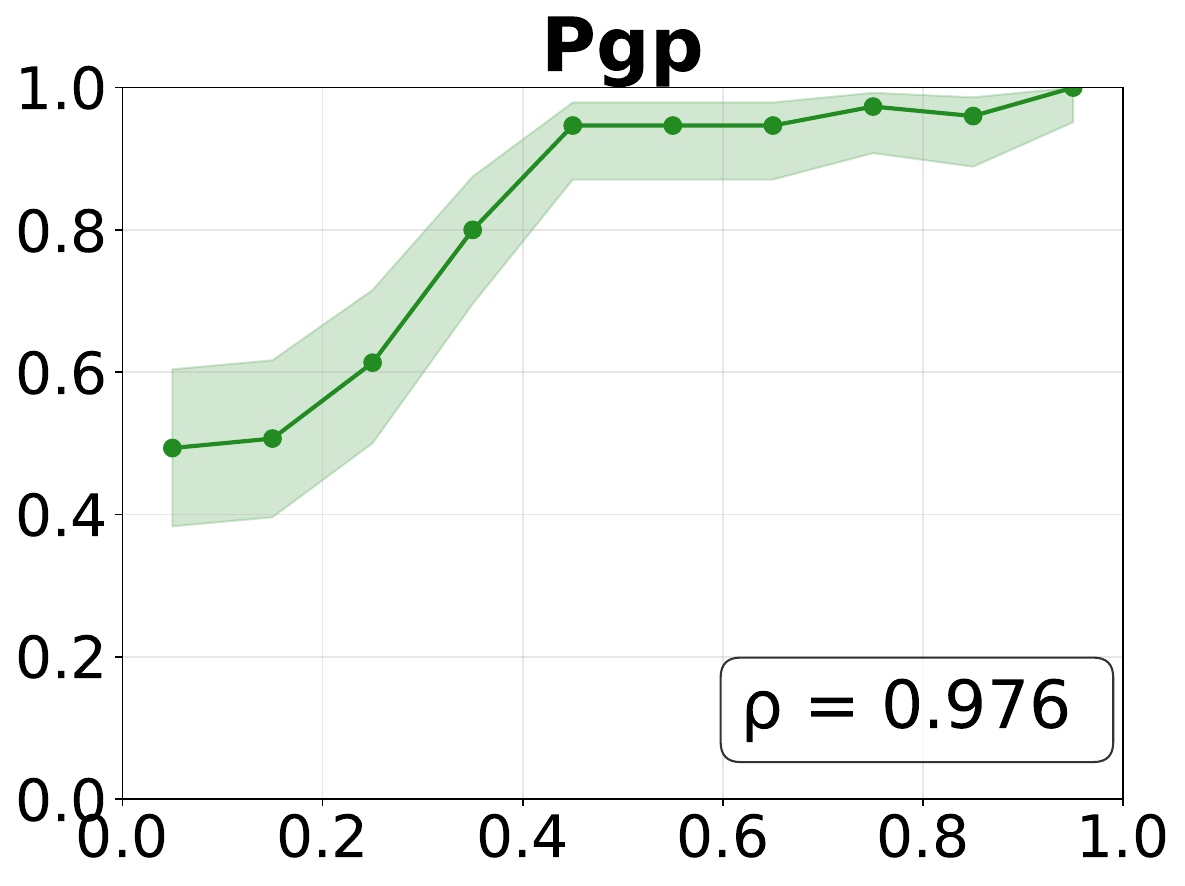} &
\includegraphics[width=0.19\textwidth]{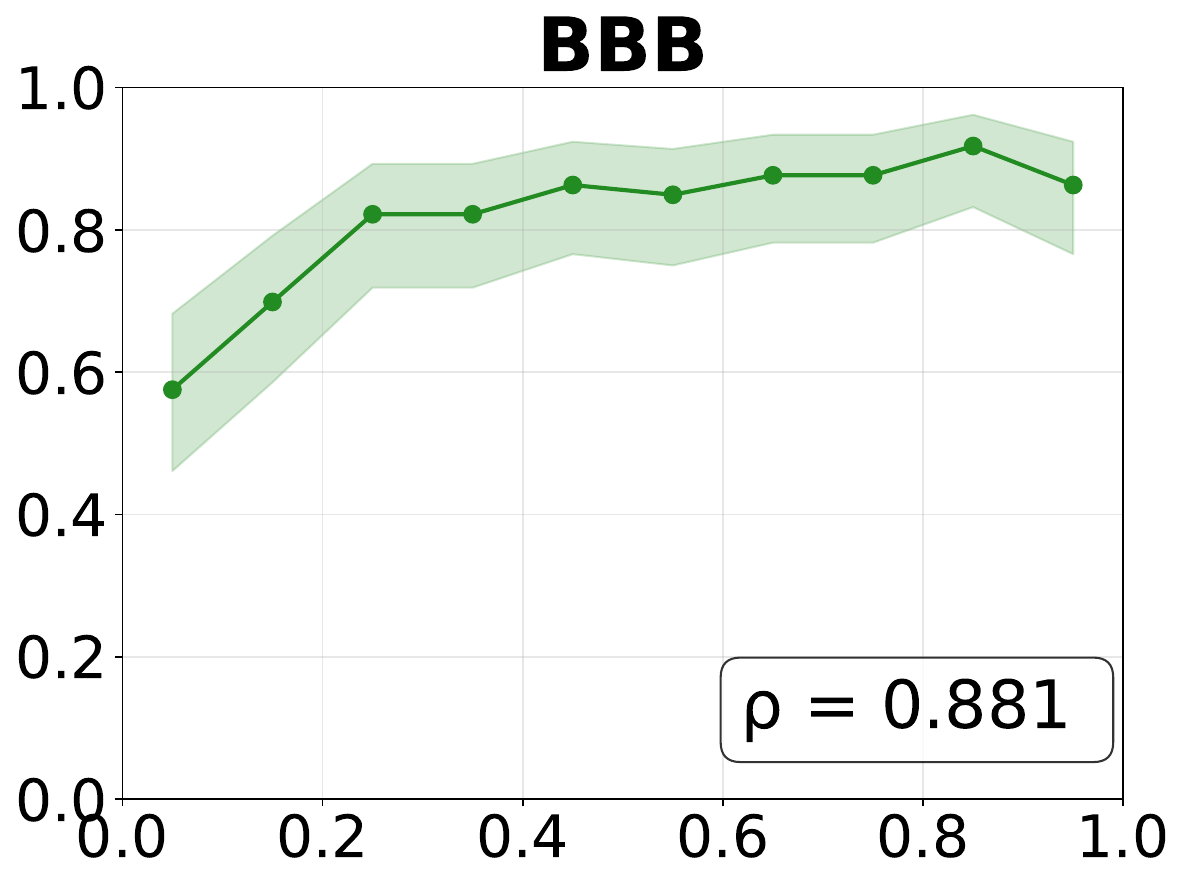} &
\includegraphics[width=0.19\textwidth]{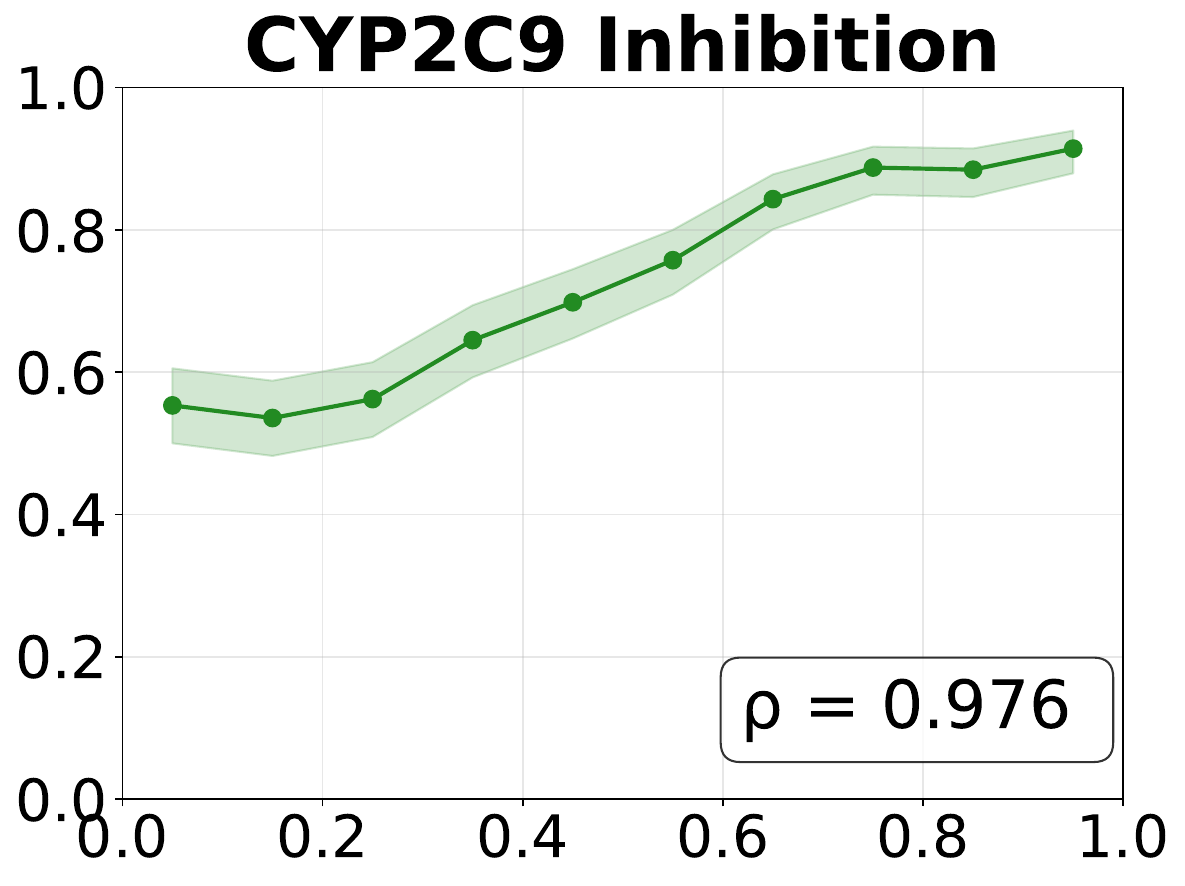} &
\includegraphics[width=0.19\textwidth]{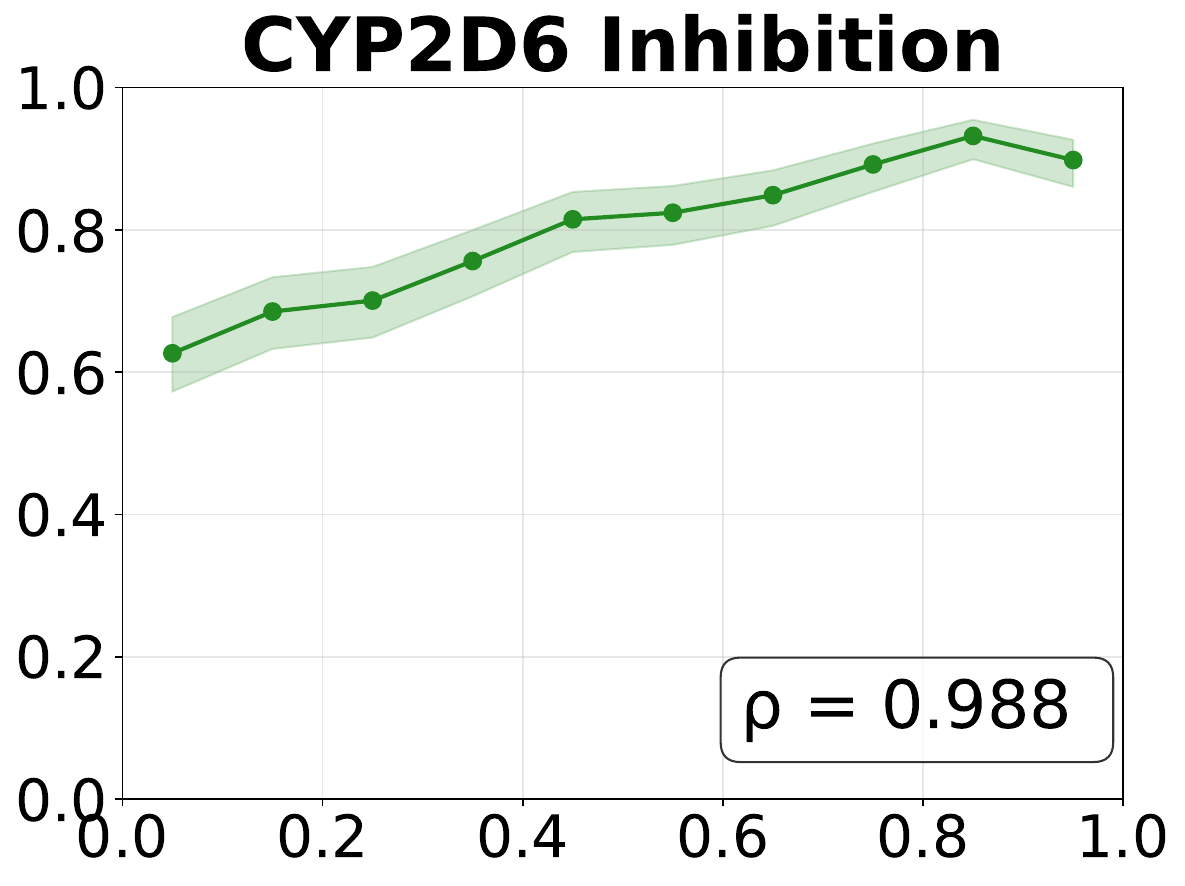} &
\includegraphics[width=0.19\textwidth]{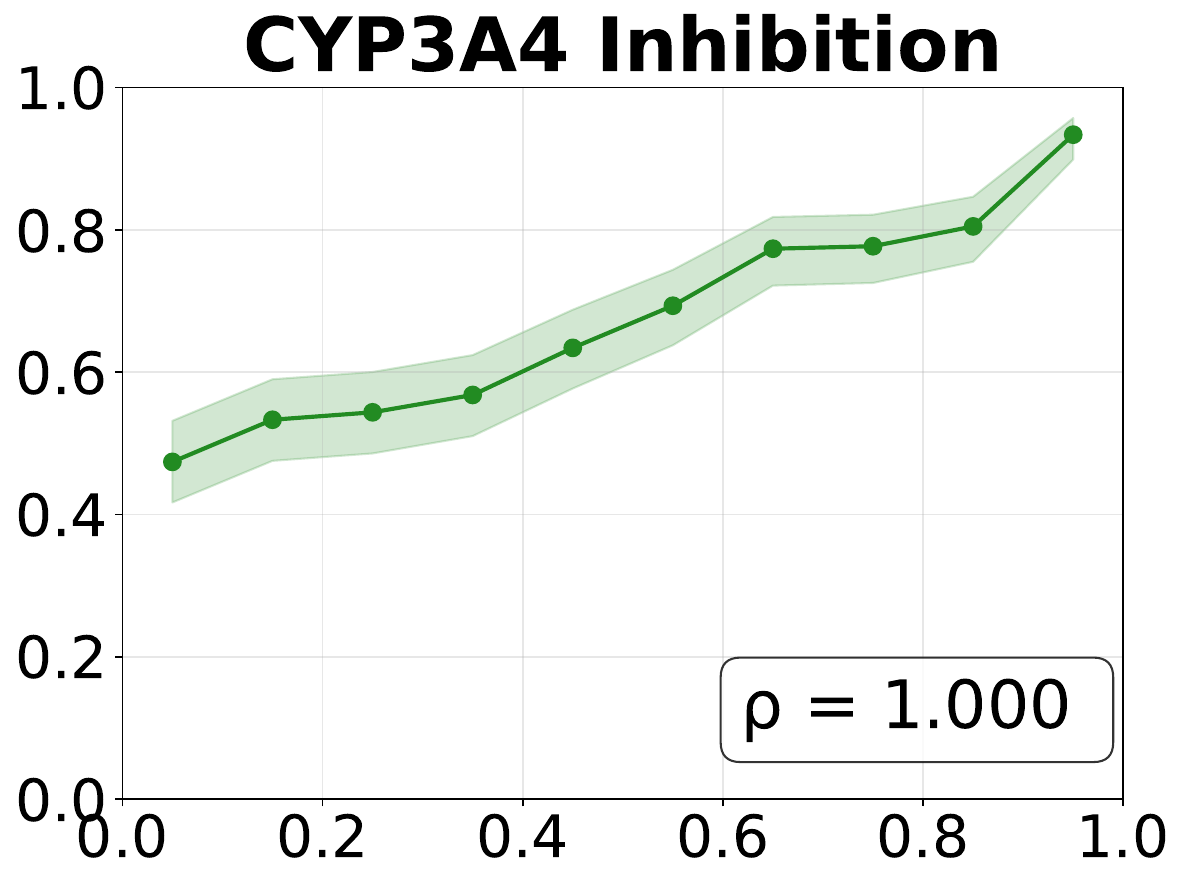} \\[0.5em]
\includegraphics[width=0.19\textwidth]{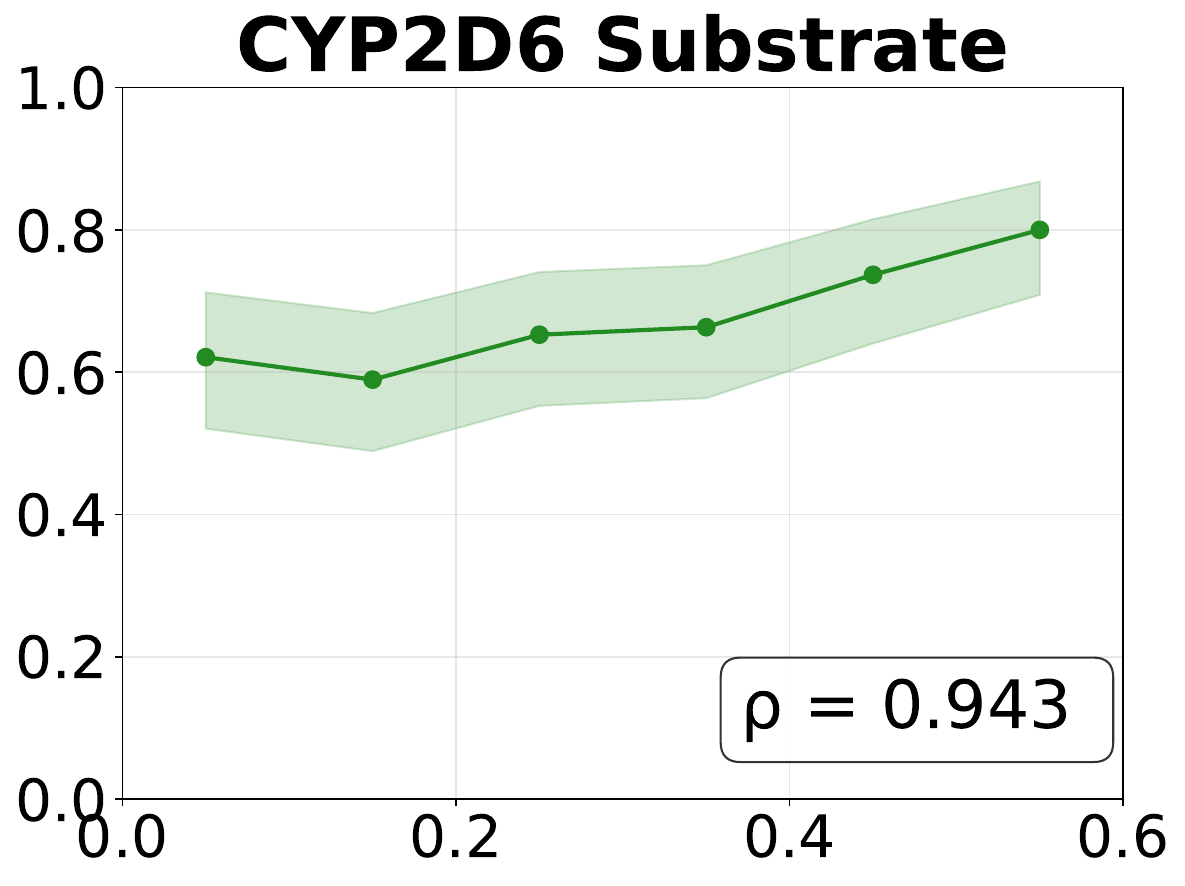} &
\includegraphics[width=0.19\textwidth]{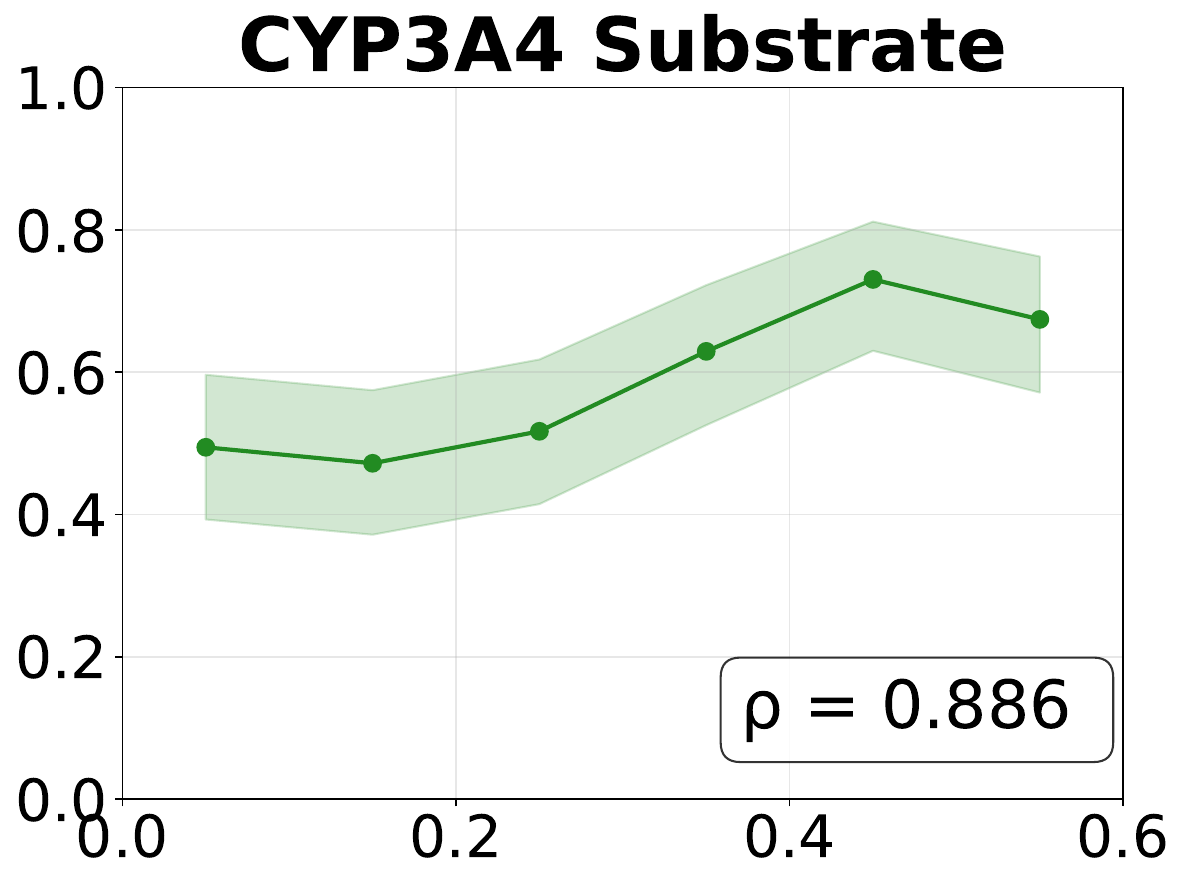} &
\includegraphics[width=0.19\textwidth]{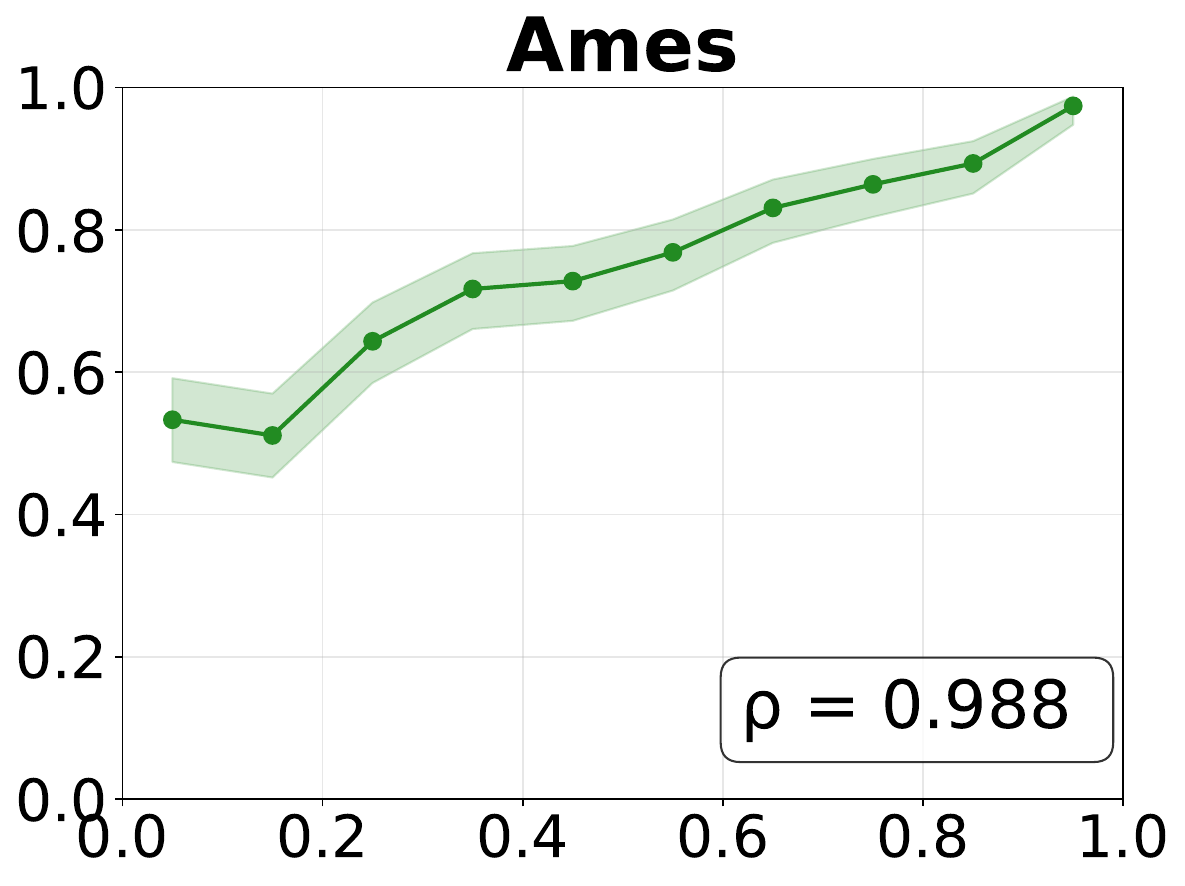} &
\includegraphics[width=0.19\textwidth]{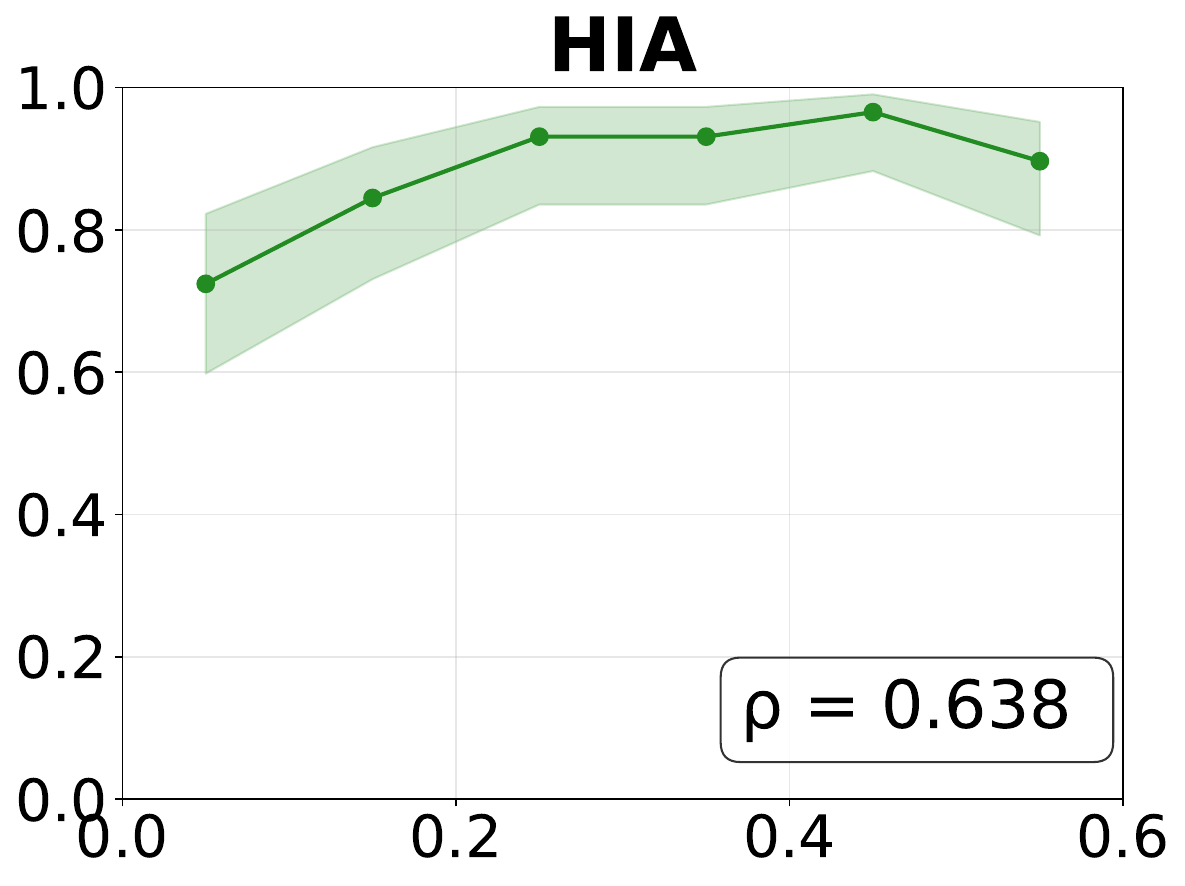} &
\includegraphics[width=0.19\textwidth]{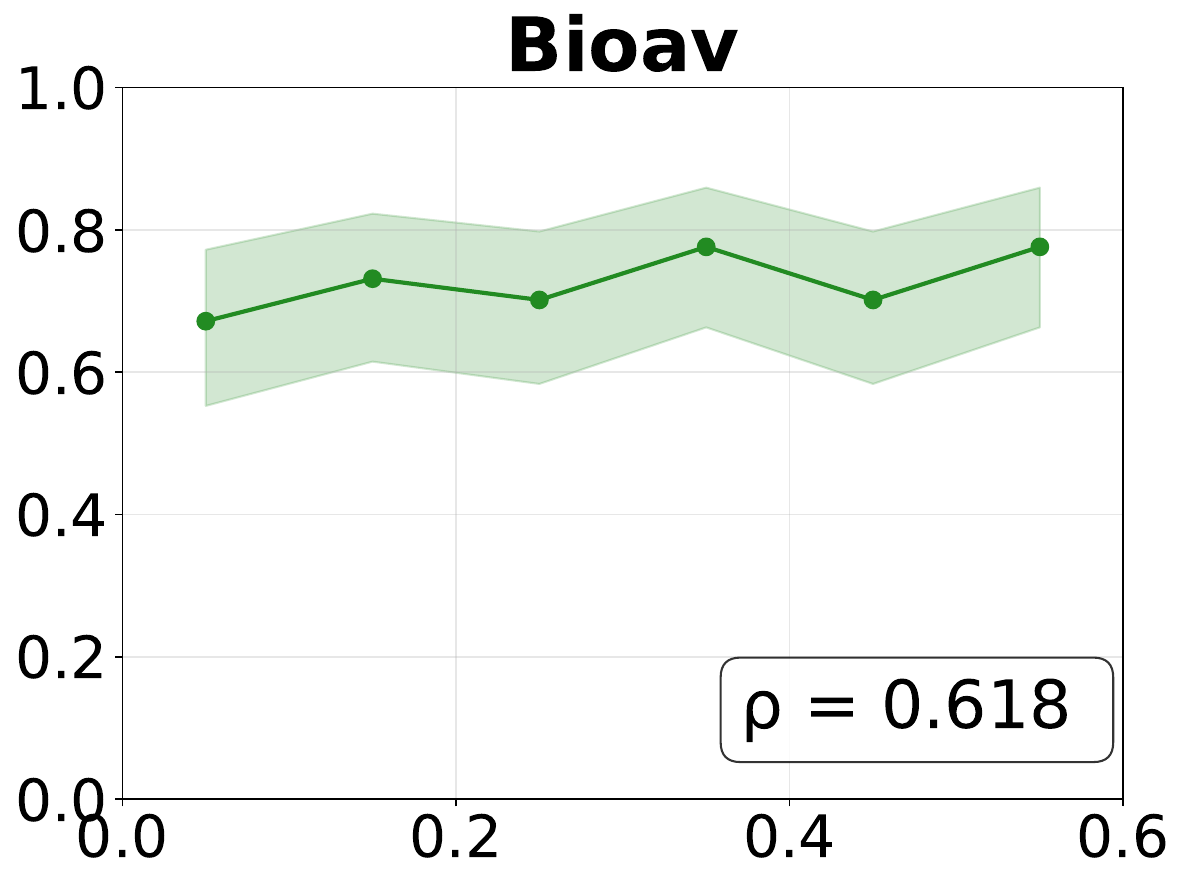} \\[0.5em]
\includegraphics[width=0.19\textwidth]{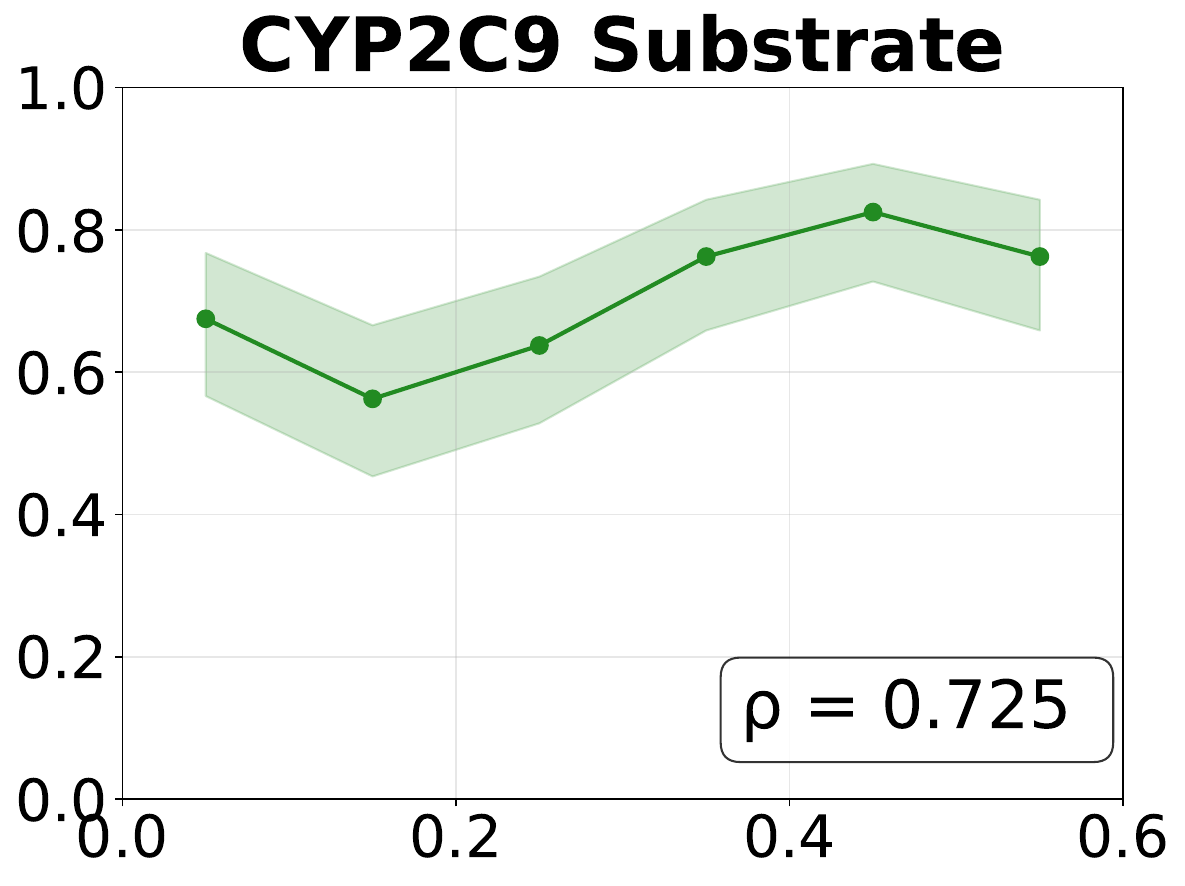} &
\includegraphics[width=0.19\textwidth]{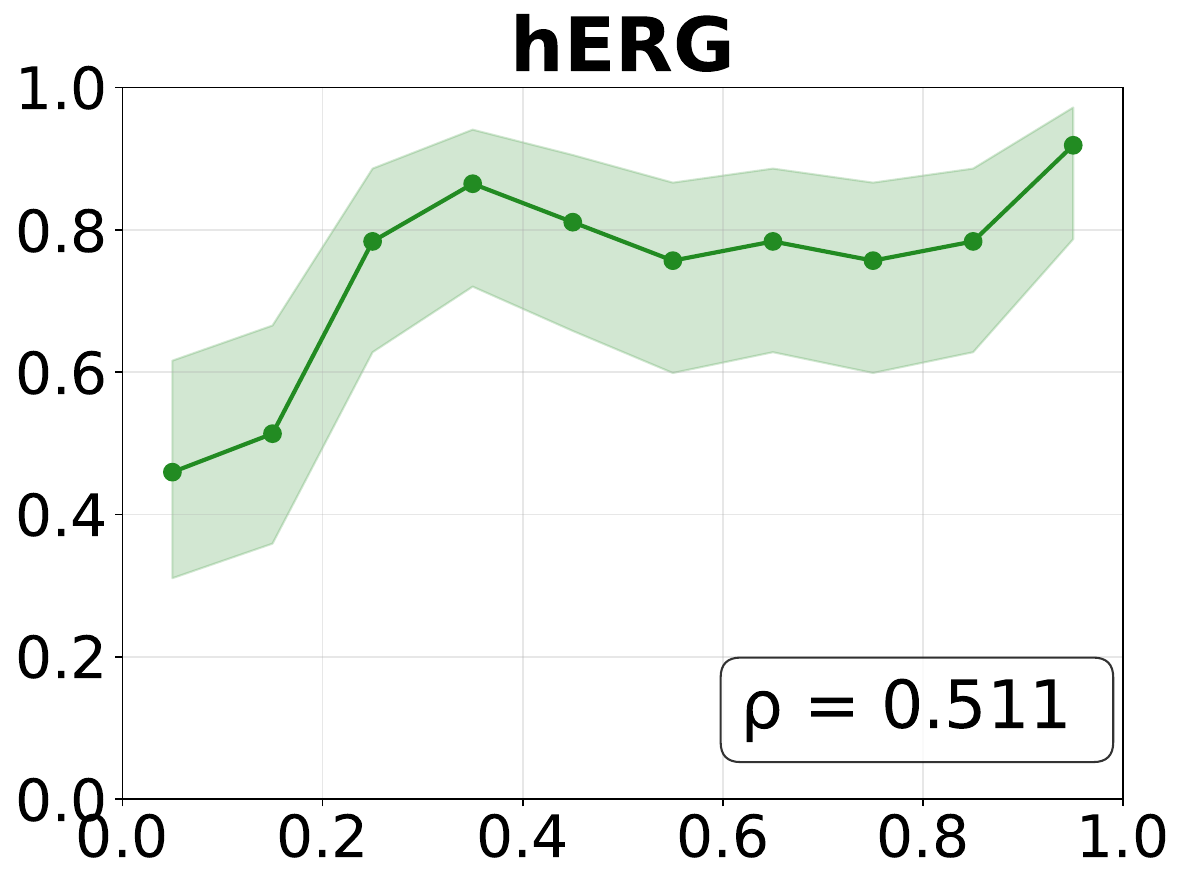} &
\includegraphics[width=0.19\textwidth]{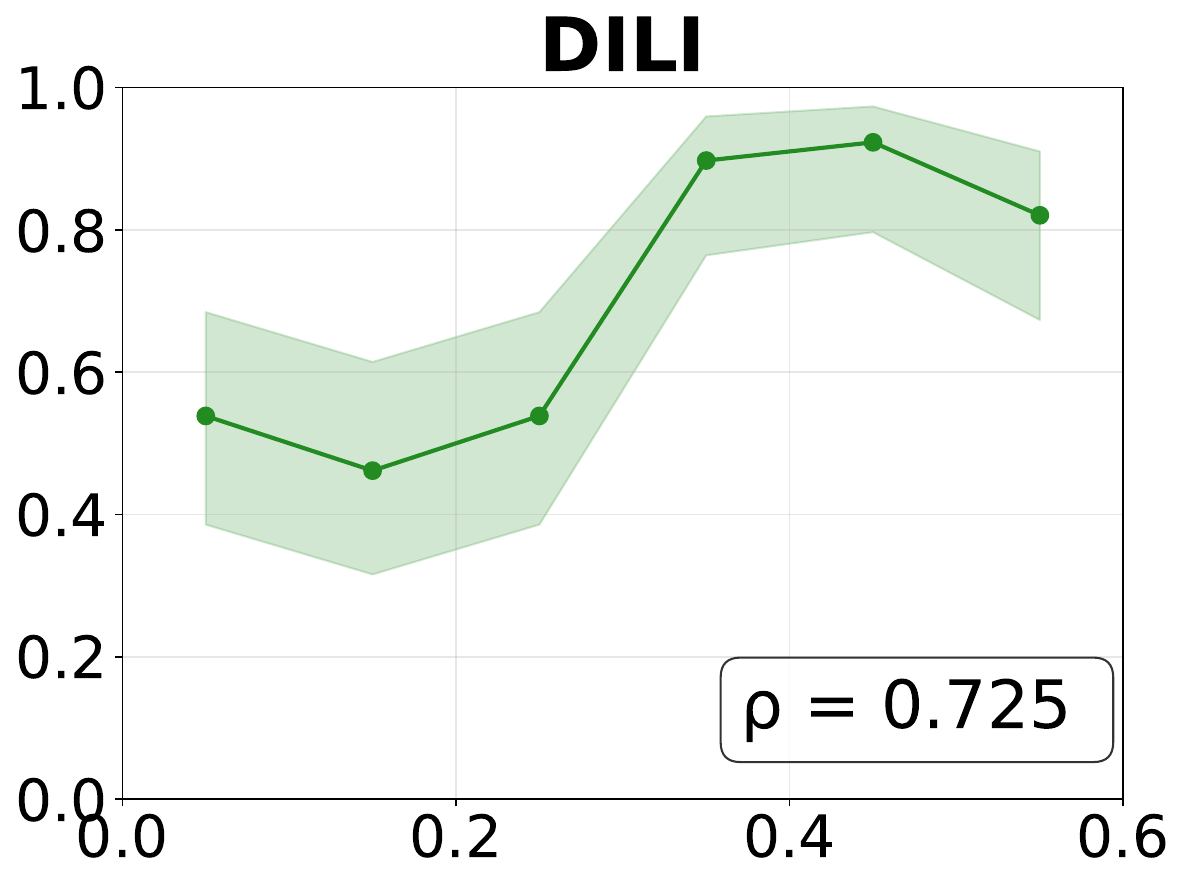} & & \\
\end{tabular}

\caption{\textbf{Supplementary Fig.~\ref{fig:supp_tanimoto_all} $|$ Structure-property alignment analysis for all 20 TDC ADMET datasets.} Relationship between Tanimoto similarity (x-axis) and property alignment for each task. Regression tasks (blue) show median absolute property difference with interquartile range shading (lower is better); classification tasks (green) show label agreement rate with 95\% Wilson confidence interval shading (higher is better). Spearman $\rho$ is annotated in each panel.}
\label{fig:supp_tanimoto_all}
\end{figure*}

\subsection*{Supplementary Note 1. Sensitivity analyses}

\paragraph{Regularization weight.}
We evaluated the regularization weight $\lambda$ for all three backbones to determine whether the structural prior has a stable operating range. MolCA showed an inverted-U response: retrieval improved as $\lambda$ increased to 100, but decreased at $\lambda=1000$, indicating that an overly strong structural constraint can interfere with the primary molecule--text objective. Atomas-B also benefited most around $\lambda=100$, whereas Atomas-L remained robust at larger weights. These trends support the use of $\lambda=100$ as a shared default for the main experiments, while showing that model capacity affects how much structural prior can be absorbed.

\paragraph{Fingerprint type.}
The fingerprint ablation tests whether the structural target must be compatible with the molecular encoder. Morgan fingerprints performed best for MolCA, consistent with the graph encoder's emphasis on local atom neighborhoods. MACCS keys performed better for both Atomas variants, consistent with a SMILES encoder that is more directly exposed to functional-group patterns. RDKit topological fingerprints were weaker across models and degraded MolCA, suggesting that path-based similarity provides a less compatible relational target for the molecule--language objectives evaluated here.

\paragraph{CKA sample size.}
The CKA sample size $n$ controls how many molecules define the relational alignment target. In MolCA, $n$ is tied to the training batch size, whereas in Atomas it is decoupled through the fingerprint queue. Moderate sample sizes were most effective: MolCA retained a stable relative gain across batch sizes, and both Atomas variants performed best around $n=128$. Very large CKA samples reduced the gain, especially for Atomas-B, suggesting that overly detailed pairwise targets can become difficult for smaller encoders to absorb.

\begin{table}[ht]
\centering
\renewcommand{\arraystretch}{0.92}
\caption{\textbf{Supplementary Table~1 $|$ Ablation study results on MolCA.} Validation set performance on ChEBI-20. For (a) and (b), the batch size is fixed at 64. For (c), since the CKA sample size $n$ equals the batch size in MolCA, a separate baseline is trained for each $n$ to isolate the effect of \methodname. $\Delta$: absolute difference from Base. \%: relative improvement.}
\label{tab:supp_molca_ablation}
\begin{tabular}{lcccccccc}
\toprule
& \multicolumn{4}{c}{\textbf{Text $\rightarrow$ Molecule}} & \multicolumn{4}{c}{\textbf{Molecule $\rightarrow$ Text}} \\
\cmidrule(lr){2-5} \cmidrule(lr){6-9}
\textbf{Setting} & R@1 & R@5 & R@10 & MRR & R@1 & R@5 & R@10 & MRR \\
\midrule
\multicolumn{9}{l}{\textit{(a) Regularization weight $\lambda$}} \\
\midrule
Base (no SAGA) & 31.88 & 47.17 & 52.98 & 39.18 & 28.87 & 42.59 & 47.56 & 35.67 \\
\midrule
$\lambda=0.1$ & 32.75 & 49.05 & 54.62 & 40.53 & 29.75 & 43.56 & 48.65 & 36.53 \\
\rowcolor{gray!10} \quad $\Delta$ & +0.87 & +1.88 & +1.64 & +1.35 & +0.88 & +0.97 & +1.09 & +0.86 \\
\rowcolor{gray!10} \quad (\%) & (+2.7\%) & (+4.0\%) & (+3.1\%) & (+3.4\%) & (+3.0\%) & (+2.3\%) & (+2.3\%) & (+2.4\%) \\
\midrule
$\lambda=1$ & 33.48 & 49.39 & 54.77 & 41.07 & 30.87 & 44.41 & 49.38 & 37.55 \\
\rowcolor{gray!10} \quad $\Delta$ & +1.60 & +2.22 & +1.79 & +1.89 & +2.00 & +1.82 & +1.82 & +1.88 \\
\rowcolor{gray!10} \quad (\%) & (+5.0\%) & (+4.7\%) & (+3.4\%) & (+4.8\%) & (+6.9\%) & (+4.3\%) & (+3.8\%) & (+5.3\%) \\
\midrule
$\lambda=10$ & 33.83 & 52.27 & 58.19 & 42.52 & 30.72 & 46.93 & 51.77 & 38.31 \\
\rowcolor{gray!10} \quad $\Delta$ & +1.95 & +5.10 & +5.21 & +3.34 & +1.85 & +4.34 & +4.21 & +2.64 \\
\rowcolor{gray!10} \quad (\%) & (+6.1\%) & (+10.8\%) & (+9.8\%) & (+8.5\%) & (+6.4\%) & (+10.2\%) & (+8.9\%) & (+7.4\%) \\
\midrule
\textbf{$\lambda=100$} & \textbf{35.66} & \textbf{57.10} & \textbf{65.07} & \textbf{45.87} & \textbf{31.20} & \textbf{50.59} & \textbf{57.26} & \textbf{40.33} \\
\rowcolor{gray!10} \quad $\Delta$ & \textbf{+3.78} & \textbf{+9.93} & \textbf{+12.09} & \textbf{+6.69} & \textbf{+2.33} & \textbf{+8.00} & \textbf{+9.70} & \textbf{+4.66} \\
\rowcolor{gray!10} \quad (\%) & \textbf{(+11.9\%)} & \textbf{(+21.1\%)} & \textbf{(+22.8\%)} & \textbf{(+17.1\%)} & \textbf{(+8.1\%)} & \textbf{(+18.8\%)} & \textbf{(+20.4\%)} & \textbf{(+13.1\%)} \\
\midrule
$\lambda=1000$ & 30.65 & 48.65 & 56.13 & 39.37 & 28.51 & 44.20 & 49.50 & 35.93 \\
\rowcolor{gray!10} \quad $\Delta$ & $-$1.23 & +1.48 & +3.15 & +0.19 & $-$0.36 & +1.61 & +1.94 & +0.26 \\
\rowcolor{gray!10} \quad (\%) & ($-$3.9\%) & (+3.1\%) & (+5.9\%) & (+0.5\%) & ($-$1.2\%) & (+3.8\%) & (+4.1\%) & (+0.7\%) \\
\midrule
\multicolumn{9}{l}{\textit{(b) Fingerprint type ($\lambda=100$)}} \\
\midrule
\textbf{Morgan} & \textbf{35.66} & \textbf{57.10} & \textbf{65.07} & \textbf{45.87} & 31.20 & \textbf{50.59} & \textbf{57.26} & 40.33 \\
\rowcolor{gray!10} \quad $\Delta$ & \textbf{+3.78} & \textbf{+9.93} & \textbf{+12.09} & \textbf{+6.69} & +2.33 & \textbf{+8.00} & \textbf{+9.70} & +4.66 \\
\rowcolor{gray!10} \quad (\%) & \textbf{(+11.9\%)} & \textbf{(+21.1\%)} & \textbf{(+22.8\%)} & \textbf{(+17.1\%)} & (+8.1\%) & \textbf{(+18.8\%)} & \textbf{(+20.4\%)} & (+13.1\%) \\
\midrule
MACCS & 34.84 & 53.65 & 61.07 & 44.00 & \textbf{32.57} & 49.17 & 55.17 & \textbf{40.38} \\
\rowcolor{gray!10} \quad $\Delta$ & +2.96 & +6.48 & +8.09 & +4.82 & \textbf{+3.70} & +6.58 & +7.61 & \textbf{+4.71} \\
\rowcolor{gray!10} \quad (\%) & (+9.3\%) & (+13.7\%) & (+15.3\%) & (+12.3\%) & \textbf{(+12.8\%)} & (+15.4\%) & (+16.0\%) & \textbf{(+13.2\%)} \\
\midrule
RDKit & 28.23 & 46.44 & 52.68 & 36.53 & 24.24 & 41.71 & 47.65 & 32.47 \\
\rowcolor{gray!10} \quad $\Delta$ & $-$3.65 & $-$0.73 & $-$0.30 & $-$2.65 & $-$4.63 & $-$0.88 & +0.09 & $-$3.20 \\
\rowcolor{gray!10} \quad (\%) & ($-$11.4\%) & ($-$1.5\%) & ($-$0.6\%) & ($-$6.8\%) & ($-$16.0\%) & ($-$2.1\%) & (+0.2\%) & ($-$9.0\%) \\
\midrule
\multicolumn{9}{l}{\textit{(c) CKA sample size $n$ ($\lambda=100$, Morgan)}} \\
\midrule
Base ($n$=16) & 33.72 & 52.56 & 58.92 & 42.42 & \textbf{31.29} & 46.65 & 51.83 & 38.66 \\
+ \methodname & \textbf{37.38} & \textbf{61.62} & \textbf{69.80} & \textbf{48.47} & 30.60 & \textbf{49.02} & \textbf{56.29} & \textbf{39.53} \\
\rowcolor{gray!10} \quad $\Delta$ & +3.66 & +9.06 & +10.88 & +6.05 & $-$0.69 & +2.37 & +4.46 & +0.87 \\
\rowcolor{gray!10} \quad (\%) & (+10.9\%) & (+17.2\%) & (+18.5\%) & (+14.3\%) & ($-$2.2\%) & (+5.1\%) & (+8.6\%) & (+2.3\%) \\
\midrule
Base ($n$=32) & 33.37 & 51.83 & 57.56 & 41.84 & 30.69 & 46.23 & 51.44 & 38.14 \\
+ \methodname & \textbf{37.09} & \textbf{58.86} & \textbf{66.68} & \textbf{47.21} & \textbf{32.08} & \textbf{52.68} & \textbf{59.98} & \textbf{41.73} \\
\rowcolor{gray!10} \quad $\Delta$ & +3.72 & +7.03 & +9.12 & +5.37 & +1.39 & +6.45 & +8.54 & +3.59 \\
\rowcolor{gray!10} \quad (\%) & (+11.1\%) & (+13.6\%) & (+15.8\%) & (+12.8\%) & (+4.5\%) & (+14.0\%) & (+16.6\%) & (+9.4\%) \\
\midrule
Base ($n$=64) & 31.88 & 47.17 & 52.98 & 39.18 & 28.87 & 42.59 & 47.56 & 35.67 \\
+ \methodname & \textbf{35.66} & \textbf{57.10} & \textbf{65.07} & \textbf{45.87} & \textbf{31.20} & \textbf{50.59} & \textbf{57.26} & \textbf{40.33} \\
\rowcolor{gray!10} \quad $\Delta$ & +3.78 & +9.93 & +12.09 & +6.69 & +2.33 & +8.00 & +9.70 & +4.66 \\
\rowcolor{gray!10} \quad (\%) & (+11.9\%) & (+21.1\%) & (+22.8\%) & (+17.1\%) & (+8.1\%) & (+18.8\%) & (+20.4\%) & (+13.1\%) \\
\midrule
Base ($n$=128) & 30.87 & 46.83 & 52.20 & 38.38 & 27.81 & 41.56 & 46.74 & 34.47 \\
+ \methodname & \textbf{34.47} & \textbf{57.19} & \textbf{64.28} & \textbf{44.83} & \textbf{30.29} & \textbf{48.95} & \textbf{56.35} & \textbf{39.29} \\
\rowcolor{gray!10} \quad $\Delta$ & +3.60 & +10.36 & +12.08 & +6.45 & +2.48 & +7.39 & +9.61 & +4.82 \\
\rowcolor{gray!10} \quad (\%) & (+11.7\%) & (+22.1\%) & (+23.1\%) & (+16.8\%) & (+8.9\%) & (+17.8\%) & (+20.6\%) & (+14.0\%) \\
\bottomrule
\end{tabular}
\end{table}

\begin{table}[ht]
\centering
\renewcommand{\arraystretch}{0.92}
\caption{\textbf{Supplementary Table~2 $|$ Ablation study results on Atomas-L.} Validation set performance on ChEBI-20. The training batch size is fixed at 16; the CKA sample size $n$ is decoupled via the fingerprint queue, with the fixed settings for each ablated factor given in the sub-section headers. $\Delta$: absolute difference from Base. \%: relative improvement.}
\label{tab:supp_atomas_large_ablation}
\begin{tabular}{lcccccccc}
\toprule
& \multicolumn{4}{c}{\textbf{Text $\rightarrow$ Molecule}} & \multicolumn{4}{c}{\textbf{Molecule $\rightarrow$ Text}} \\
\cmidrule(lr){2-5} \cmidrule(lr){6-9}
\textbf{Setting} & R@1 & R@5 & R@10 & MRR & R@1 & R@5 & R@10 & MRR \\
\midrule
\multicolumn{9}{l}{\textit{(a) Regularization weight $\lambda$ ($n=128$, Morgan)}} \\
\midrule
Base (no SAGA) & 43.15 & 69.21 & 77.96 & 55.23 & 41.56 & 65.01 & 73.70 & 52.49 \\
\midrule
$\lambda=0.1$ & 44.66 & 71.83 & 80.37 & 57.06 & 40.89 & 66.34 & 75.09 & 52.66 \\
\rowcolor{gray!10} \quad $\Delta$ & +1.51 & +2.62 & +2.41 & +1.83 & $-$0.67 & +1.33 & +1.39 & +0.17 \\
\rowcolor{gray!10} \quad (\%) & (+3.5\%) & (+3.8\%) & (+3.1\%) & (+3.3\%) & ($-$1.6\%) & (+2.0\%) & (+1.9\%) & (+0.3\%) \\
\midrule
$\lambda=1$ & 44.18 & 72.23 & 79.98 & 56.80 & 40.15 & 66.05 & 73.98 & 51.95 \\
\rowcolor{gray!10} \quad $\Delta$ & +1.03 & +3.02 & +2.02 & +1.57 & $-$1.41 & +1.04 & +0.28 & $-$0.54 \\
\rowcolor{gray!10} \quad (\%) & (+2.4\%) & (+4.4\%) & (+2.6\%) & (+2.8\%) & ($-$3.4\%) & (+1.6\%) & (+0.4\%) & ($-$1.0\%) \\
\midrule
$\lambda=10$ & 47.41 & 74.82 & 82.96 & 59.64 & 44.52 & 70.43 & 79.09 & 56.28 \\
\rowcolor{gray!10} \quad $\Delta$ & +4.26 & +5.61 & +5.00 & +4.41 & +2.96 & +5.42 & +5.39 & +3.79 \\
\rowcolor{gray!10} \quad (\%) & (+9.9\%) & (+8.1\%) & (+6.4\%) & (+8.0\%) & (+7.1\%) & (+8.3\%) & (+7.3\%) & (+7.2\%) \\
\midrule
$\lambda=100$ & 50.97 & 76.48 & 83.47 & 62.49 & 46.40 & 71.70 & 78.30 & 57.74 \\
\rowcolor{gray!10} \quad $\Delta$ & +7.82 & +7.27 & +5.51 & +7.26 & +4.84 & +6.69 & +4.60 & +5.25 \\
\rowcolor{gray!10} \quad (\%) & (+18.1\%) & (+10.5\%) & (+7.1\%) & (+13.1\%) & (+11.6\%) & (+10.3\%) & (+6.2\%) & (+10.0\%) \\
\midrule
\textbf{$\lambda=1000$} & \textbf{54.89} & \textbf{80.61} & \textbf{86.91} & \textbf{66.40} & \textbf{47.55} & \textbf{74.97} & \textbf{82.99} & \textbf{60.03} \\
\rowcolor{gray!10} \quad $\Delta$ & \textbf{+11.74} & \textbf{+11.40} & \textbf{+8.95} & \textbf{+11.17} & \textbf{+5.99} & \textbf{+9.96} & \textbf{+9.29} & \textbf{+7.54} \\
\rowcolor{gray!10} \quad (\%) & \textbf{(+27.2\%)} & \textbf{(+16.5\%)} & \textbf{(+11.5\%)} & \textbf{(+20.2\%)} & \textbf{(+14.4\%)} & \textbf{(+15.3\%)} & \textbf{(+12.6\%)} & \textbf{(+14.4\%)} \\
\midrule
\multicolumn{9}{l}{\textit{(b) Fingerprint type ($\lambda=100$, $n=128$)}} \\
\midrule
Morgan & 50.97 & 76.48 & 83.47 & 62.49 & \textbf{46.40} & 71.70 & 78.30 & 57.74 \\
\rowcolor{gray!10} \quad $\Delta$ & +9.59 & +9.62 & +8.47 & +9.54 & \textbf{+5.04} & +6.46 & +5.50 & +5.33 \\
\rowcolor{gray!10} \quad (\%) & (+22.5\%) & (+14.1\%) & (+11.1\%) & (+17.6\%) & \textbf{(+12.2\%)} & (+9.9\%) & (+7.5\%) & (+10.2\%) \\
\midrule
\textbf{MACCS} & \textbf{53.32} & \textbf{80.04} & \textbf{87.15} & \textbf{65.21} & 45.97 & \textbf{71.73} & \textbf{79.18} & \textbf{57.87} \\
\rowcolor{gray!10} \quad $\Delta$ & \textbf{+10.17} & \textbf{+10.83} & \textbf{+9.19} & \textbf{+9.98} & +4.41 & \textbf{+6.72} & \textbf{+5.48} & \textbf{+5.38} \\
\rowcolor{gray!10} \quad (\%) & \textbf{(+23.6\%)} & \textbf{(+15.6\%)} & \textbf{(+11.8\%)} & \textbf{(+18.1\%)} & (+10.6\%) & \textbf{(+10.3\%)} & \textbf{(+7.4\%)} & \textbf{(+10.3\%)} \\
\midrule
RDKit & 50.03 & 77.29 & 84.26 & 61.91 & 41.71 & 68.89 & 77.18 & 54.01 \\
\rowcolor{gray!10} \quad $\Delta$ & +6.88 & +8.08 & +6.30 & +6.68 & +0.15 & +3.88 & +3.48 & +1.52 \\
\rowcolor{gray!10} \quad (\%) & (+15.9\%) & (+11.7\%) & (+8.1\%) & (+12.1\%) & (+0.4\%) & (+6.0\%) & (+4.7\%) & (+2.9\%) \\
\midrule
\multicolumn{9}{l}{\textit{(c) CKA sample size $n$ ($\lambda=100$, Morgan)}} \\
\midrule
Base (no SAGA) & 43.15 & 69.21 & 77.96 & 55.23 & 41.56 & 65.01 & 73.70 & 52.49 \\
\midrule
\textbf{64} & \textbf{51.48} & \textbf{77.35} & \textbf{84.74} & \textbf{63.05} & 45.34 & 71.25 & 78.12 & 56.75 \\
\rowcolor{gray!10} \quad $\Delta$ & +7.66 & +8.71 & +8.47 & +8.01 & +3.19 & +3.38 & +3.29 & +3.12 \\
\rowcolor{gray!10} \quad (\%) & (+18.0\%) & (+12.8\%) & (+11.1\%) & (+14.8\%) & (+7.7\%) & (+5.2\%) & (+4.5\%) & (+6.0\%) \\
\midrule
128 & 50.97 & 76.48 & 83.47 & 62.49 & \textbf{46.40} & \textbf{71.70} & 78.30 & \textbf{57.74} \\
\rowcolor{gray!10} \quad $\Delta$ & \textbf{+9.59} & \textbf{+9.62} & +8.47 & \textbf{+9.54} & \textbf{+5.04} & \textbf{+6.46} & +5.50 & \textbf{+5.33} \\
\rowcolor{gray!10} \quad (\%) & \textbf{(+22.5\%)} & \textbf{(+14.1\%)} & (+11.1\%) & \textbf{(+17.6\%)} & \textbf{(+12.2\%)} & \textbf{(+9.9\%)} & (+7.5\%) & \textbf{(+10.2\%)} \\
\midrule
256 & 49.10 & 76.42 & 83.53 & 61.51 & 44.88 & 69.40 & 76.27 & 55.95 \\
\rowcolor{gray!10} \quad $\Delta$ & +6.55 & +8.14 & +7.35 & +7.27 & +3.70 & +3.86 & +3.08 & +3.61 \\
\rowcolor{gray!10} \quad (\%) & (+15.4\%) & (+11.9\%) & (+9.6\%) & (+13.4\%) & (+9.0\%) & (+5.9\%) & (+4.2\%) & (+6.9\%) \\
\midrule
512 & 48.73 & 76.51 & 84.08 & 61.24 & 45.04 & 71.52 & \textbf{79.03} & 57.05 \\
\rowcolor{gray!10} \quad $\Delta$ & +6.18 & +8.23 & +7.90 & +7.00 & +3.86 & +5.98 & \textbf{+5.84} & +4.71 \\
\rowcolor{gray!10} \quad (\%) & (+14.5\%) & (+12.1\%) & (+10.4\%) & (+12.9\%) & (+9.4\%) & (+9.1\%) & \textbf{(+8.0\%)} & (+9.0\%) \\
\midrule
1024 & 45.96 & 73.70 & 82.63 & 58.55 & 40.68 & 66.07 & 73.94 & 52.20 \\
\rowcolor{gray!10} \quad $\Delta$ & +4.77 & +6.21 & +5.85 & +5.08 & +0.59 & +0.86 & +1.42 & +0.72 \\
\rowcolor{gray!10} \quad (\%) & (+11.2\%) & (+9.1\%) & (+7.7\%) & (+9.4\%) & (+1.4\%) & (+1.3\%) & (+1.9\%) & (+1.4\%) \\
\bottomrule
\end{tabular}
\end{table}

\begin{table}[ht]
\centering
\renewcommand{\arraystretch}{0.92}
\caption{\textbf{Supplementary Table~3 $|$ Ablation study results on Atomas-B.} Validation set performance on ChEBI-20. The training batch size is fixed at 16; the CKA sample size $n$ is decoupled via the fingerprint queue, with the fixed settings for each ablated factor given in the sub-section headers. $\Delta$: absolute difference from Base. \%: relative improvement.}
\label{tab:supp_atomas_base_ablation}
\begin{tabular}{lcccccccc}
\toprule
& \multicolumn{4}{c}{\textbf{Text $\rightarrow$ Molecule}} & \multicolumn{4}{c}{\textbf{Molecule $\rightarrow$ Text}} \\
\cmidrule(lr){2-5} \cmidrule(lr){6-9}
\textbf{Setting} & R@1 & R@5 & R@10 & MRR & R@1 & R@5 & R@10 & MRR \\
\midrule
\multicolumn{9}{l}{\textit{(a) Regularization weight $\lambda$ ($n=128$, Morgan)}} \\
\midrule
Base (no SAGA) & 33.56 & 58.62 & 68.85 & 45.33 & 31.05 & 55.75 & 65.19 & 42.90 \\
\midrule
$\lambda=0.1$ & 31.18 & 57.39 & 68.40 & 43.47 & 29.57 & 54.15 & 63.68 & 41.02 \\
\rowcolor{gray!10} \quad $\Delta$ & $-$2.38 & $-$1.23 & $-$0.45 & $-$1.86 & $-$1.48 & $-$1.60 & $-$1.51 & $-$1.88 \\
\rowcolor{gray!10} \quad (\%) & ($-$7.1\%) & ($-$2.1\%) & ($-$0.7\%) & ($-$4.1\%) & ($-$4.8\%) & ($-$2.9\%) & ($-$2.3\%) & ($-$4.4\%) \\
\midrule
$\lambda=1$ & 31.18 & 57.63 & 68.24 & 43.55 & 29.72 & 55.84 & 65.10 & 41.85 \\
\rowcolor{gray!10} \quad $\Delta$ & $-$2.38 & $-$0.99 & $-$0.61 & $-$1.78 & $-$1.33 & +0.09 & $-$0.09 & $-$1.05 \\
\rowcolor{gray!10} \quad (\%) & ($-$7.1\%) & ($-$1.7\%) & ($-$0.9\%) & ($-$3.9\%) & ($-$4.3\%) & (+0.2\%) & ($-$0.1\%) & ($-$2.4\%) \\
\midrule
$\lambda=10$ & 34.26 & 61.91 & 71.89 & 46.91 & 33.29 & 59.87 & 68.98 & 45.67 \\
\rowcolor{gray!10} \quad $\Delta$ & +0.70 & +3.29 & +3.04 & +1.58 & +2.24 & +4.12 & +3.79 & +2.77 \\
\rowcolor{gray!10} \quad (\%) & (+2.1\%) & (+5.6\%) & (+4.4\%) & (+3.5\%) & (+7.2\%) & (+7.4\%) & (+5.8\%) & (+6.5\%) \\
\midrule
$\lambda=100$ & 40.44 & 67.70 & \textbf{76.39} & 52.80 & \textbf{36.65} & 62.11 & 71.19 & \textbf{48.50} \\
\rowcolor{gray!10} \quad $\Delta$ & +6.88 & +9.08 & \textbf{+7.54} & +7.47 & \textbf{+5.60} & +6.36 & +6.00 & \textbf{+5.60} \\
\rowcolor{gray!10} \quad (\%) & (+20.5\%) & (+15.5\%) & \textbf{(+11.0\%)} & (+16.5\%) & \textbf{(+18.0\%)} & (+11.4\%) & (+9.2\%) & \textbf{(+13.1\%)} \\
\midrule
\textbf{$\lambda=1000$} & \textbf{42.16} & \textbf{68.37} & 76.21 & \textbf{54.09} & 35.32 & \textbf{62.29} & \textbf{71.61} & 47.76 \\
\rowcolor{gray!10} \quad $\Delta$ & \textbf{+8.60} & \textbf{+9.75} & +7.36 & \textbf{+8.76} & +4.27 & \textbf{+6.54} & \textbf{+6.42} & +4.86 \\
\rowcolor{gray!10} \quad (\%) & \textbf{(+25.6\%)} & \textbf{(+16.6\%)} & (+10.7\%) & \textbf{(+19.3\%)} & (+13.8\%) & \textbf{(+11.7\%)} & \textbf{(+9.8\%)} & (+11.3\%) \\
\midrule
\multicolumn{9}{l}{\textit{(b) Fingerprint type ($\lambda=100$, $n=128$)}} \\
\midrule
Morgan & 40.44 & 67.70 & 76.39 & 52.80 & 36.65 & 62.11 & 71.19 & 48.50 \\
\rowcolor{gray!10} \quad $\Delta$ & +6.88 & +9.08 & +7.54 & +7.47 & +5.60 & +6.36 & +6.00 & +5.60 \\
\rowcolor{gray!10} \quad (\%) & (+20.5\%) & (+15.5\%) & (+11.0\%) & (+16.5\%) & (+18.0\%) & (+11.4\%) & (+9.2\%) & (+13.1\%) \\
\midrule
\textbf{MACCS} & \textbf{41.92} & \textbf{70.33} & \textbf{78.83} & \textbf{54.65} & \textbf{39.01} & \textbf{66.98} & \textbf{75.54} & \textbf{51.50} \\
\rowcolor{gray!10} \quad $\Delta$ & \textbf{+8.36} & \textbf{+11.71} & \textbf{+9.98} & \textbf{+9.32} & \textbf{+7.96} & \textbf{+11.23} & \textbf{+10.35} & \textbf{+8.60} \\
\rowcolor{gray!10} \quad (\%) & \textbf{(+24.9\%)} & \textbf{(+20.0\%)} & \textbf{(+14.5\%)} & \textbf{(+20.6\%)} & \textbf{(+25.6\%)} & \textbf{(+20.1\%)} & \textbf{(+15.9\%)} & \textbf{(+20.0\%)} \\
\midrule
RDKit & 35.01 & 62.12 & 71.11 & 47.41 & 31.05 & 56.57 & 65.47 & 42.80 \\
\rowcolor{gray!10} \quad $\Delta$ & +1.45 & +3.50 & +2.26 & +2.08 & 0.00 & +0.82 & +0.28 & $-$0.10 \\
\rowcolor{gray!10} \quad (\%) & (+4.3\%) & (+6.0\%) & (+3.3\%) & (+4.6\%) & (0.0\%) & (+1.5\%) & (+0.4\%) & ($-$0.2\%) \\
\midrule
\multicolumn{9}{l}{\textit{(c) CKA sample size $n$ ($\lambda=100$, Morgan)}} \\
\midrule
Base (no SAGA) & 33.56 & 58.62 & 68.85 & 45.33 & 31.05 & 55.75 & 65.19 & 42.90 \\
\midrule
64 & 39.99 & 67.07 & \textbf{76.42} & 52.19 & 35.93 & \textbf{63.29} & \textbf{72.49} & 48.25 \\
\rowcolor{gray!10} \quad $\Delta$ & +6.43 & +8.45 & \textbf{+7.57} & +6.86 & +4.88 & \textbf{+7.54} & \textbf{+7.30} & +5.35 \\
\rowcolor{gray!10} \quad (\%) & (+19.2\%) & (+14.4\%) & (+11.0\%) & (+15.1\%) & (+15.7\%) & \textbf{(+13.5\%)} & \textbf{(+11.2\%)} & (+12.5\%) \\
\midrule
\textbf{128} & \textbf{40.44} & \textbf{67.70} & 76.39 & \textbf{52.80} & \textbf{36.65} & 62.11 & 71.19 & \textbf{48.50} \\
\rowcolor{gray!10} \quad $\Delta$ & \textbf{+6.88} & \textbf{+9.08} & +7.54 & \textbf{+7.47} & \textbf{+5.60} & +6.36 & +6.00 & \textbf{+5.60} \\
\rowcolor{gray!10} \quad (\%) & \textbf{(+20.5\%)} & \textbf{(+15.5\%)} & (+11.0\%) & \textbf{(+16.5\%)} & \textbf{(+18.0\%)} & (+11.4\%) & (+9.2\%) & \textbf{(+13.1\%)} \\
\midrule
256 & 37.09 & 64.63 & 73.49 & 49.69 & 34.32 & 59.41 & 68.01 & 46.19 \\
\rowcolor{gray!10} \quad $\Delta$ & +3.53 & +6.01 & +4.64 & +4.36 & +3.27 & +3.66 & +2.82 & +3.29 \\
\rowcolor{gray!10} \quad (\%) & (+10.5\%) & (+10.3\%) & (+6.7\%) & (+9.6\%) & (+10.5\%) & (+6.6\%) & (+4.3\%) & (+7.7\%) \\
\midrule
512 & 37.03 & 65.29 & 74.85 & 49.91 & 32.69 & 58.81 & 68.83 & 44.76 \\
\rowcolor{gray!10} \quad $\Delta$ & +3.47 & +6.67 & +6.00 & +4.58 & +1.64 & +3.06 & +3.64 & +1.86 \\
\rowcolor{gray!10} \quad (\%) & (+10.3\%) & (+11.4\%) & (+8.7\%) & (+10.1\%) & (+5.3\%) & (+5.5\%) & (+5.6\%) & (+4.3\%) \\
\midrule
1024 & 31.24 & 57.66 & 67.28 & 43.34 & 28.66 & 51.66 & 61.77 & 39.79 \\
\rowcolor{gray!10} \quad $\Delta$ & $-$2.32 & $-$0.96 & $-$1.57 & $-$1.99 & $-$2.39 & $-$4.09 & $-$3.42 & $-$3.11 \\
\rowcolor{gray!10} \quad (\%) & ($-$6.9\%) & ($-$1.6\%) & ($-$2.3\%) & ($-$4.4\%) & ($-$7.7\%) & ($-$7.3\%) & ($-$5.2\%) & ($-$7.2\%) \\
\bottomrule
\end{tabular}
\end{table}

\begin{figure}[ht]
\centering
\begin{overpic}[width=\textwidth]{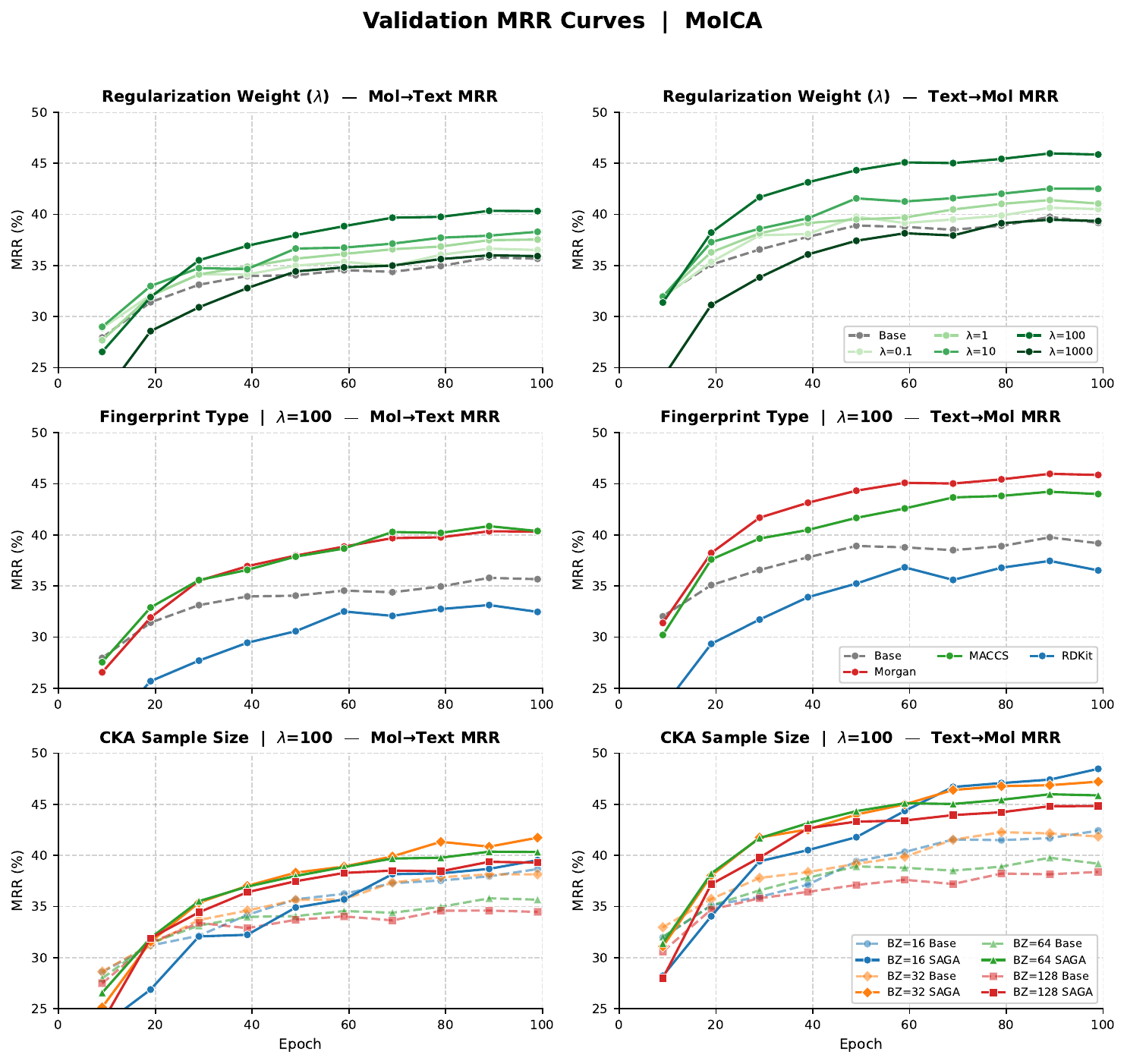}
  \put(1,86){\normalsize\textbf{(a)}}
  \put(1,58){\normalsize\textbf{(b)}}
  \put(1,29){\normalsize\textbf{(c)}}
\end{overpic}
\caption{\textbf{Supplementary Fig.~\ref{fig:supp_molca_ablation} $|$ Validation MRR curves for \methodname\ configurations on MolCA.} Mol-to-Text (left) and Text-to-Mol (right) retrieval on ChEBI-20 across training epochs. \textbf{(a)} Effect of SAGA regularization weight $\lambda$ with batch size 64. Darker green indicates higher $\lambda$; $\lambda=100$ achieves optimal performance. \textbf{(b)} Comparison of fingerprint types (Morgan, MACCS, RDKit) with $\lambda=100$. \textbf{(c)} Effect of batch size with $\lambda=100$; solid lines denote \methodname, dashed lines denote baseline.}
\label{fig:supp_molca_ablation}
\end{figure}

\begin{figure}[ht]
\centering
\begin{overpic}[width=\textwidth]{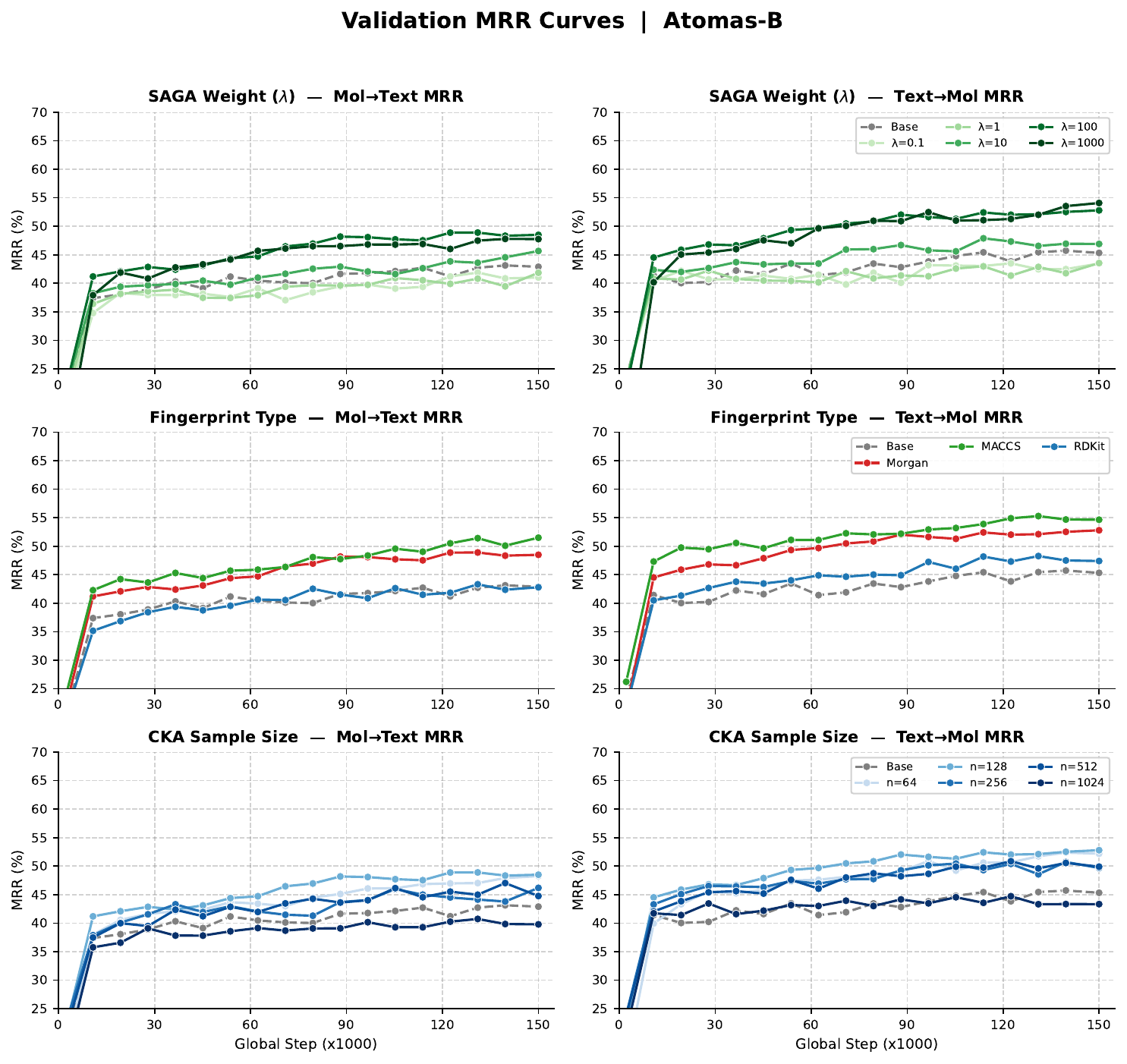}
  \put(1,86){\normalsize\textbf{(a)}}
  \put(1,58){\normalsize\textbf{(b)}}
  \put(1,29){\normalsize\textbf{(c)}}
\end{overpic}
\caption{\textbf{Supplementary Fig.~\ref{fig:supp_atomas_b_ablation} $|$ Validation MRR curves for \methodname\ configurations on Atomas-B (MolT5-base).} Mol-to-Text (left) and Text-to-Mol (right) retrieval on ChEBI-20 across training steps. \textbf{(a)} Effect of SAGA regularization weight $\lambda$ at CKA sample size $n=128$. \textbf{(b)} Comparison of fingerprint types (Morgan, MACCS, RDKit) at $\lambda=100$, $n=128$. \textbf{(c)} Effect of CKA sample size $n$ at $\lambda=100$; solid lines denote \methodname, dashed lines denote baseline.}
\label{fig:supp_atomas_b_ablation}
\end{figure}

\begin{figure}[ht]
\centering
\begin{overpic}[width=\textwidth]{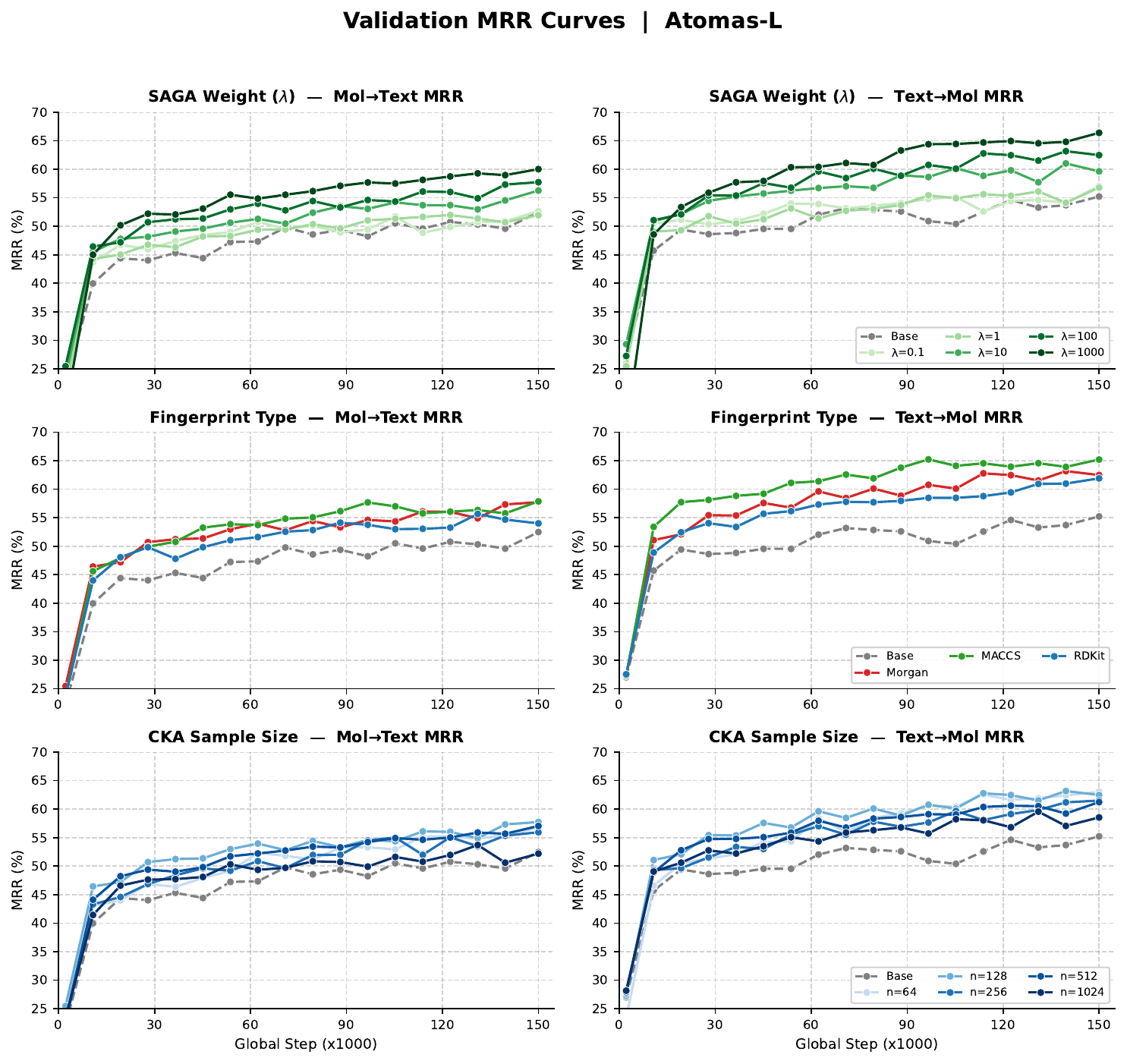}
  \put(1,86){\normalsize\textbf{(a)}}
  \put(1,58){\normalsize\textbf{(b)}}
  \put(1,29){\normalsize\textbf{(c)}}
\end{overpic}
\caption{\textbf{Supplementary Fig.~\ref{fig:supp_atomas_l_ablation} $|$ Validation MRR curves for \methodname\ configurations on Atomas-L (MolT5-large).} Mol-to-Text (left) and Text-to-Mol (right) retrieval on ChEBI-20 across training steps. \textbf{(a)} Effect of SAGA regularization weight $\lambda$ at CKA sample size $n=128$; $\lambda=1000$ achieves the best performance. \textbf{(b)} Comparison of fingerprint types (Morgan, MACCS, RDKit) at $\lambda=100$, $n=128$. \textbf{(c)} Effect of CKA sample size $n$ at $\lambda=100$; solid lines denote \methodname, dashed lines denote baseline.}
\label{fig:supp_atomas_l_ablation}
\end{figure}

\subsection*{Supplementary Note 2. Representation-space analyses}

\paragraph{Molecular alignment and text transfer.}
The representation analyses quantify how \methodname\ changes the geometry of molecule and text embeddings after training. Across all backbones, molecule--fingerprint CKA increased, showing that the learned molecular representation became more aligned with fingerprint-derived structural relationships. This reorganization also propagated to the text side through contrastive training: text--fingerprint CKA increased even though the CKA loss was applied only to molecular representations. Molecule--text CKA remained stable or improved for MolCA and Atomas-L, and changed only marginally for Atomas-B, indicating that the added structural organization does not disrupt cross-modal alignment.

\paragraph{Effects of regularization weight and fingerprint type.}
Representation-space trends mirrored the retrieval sensitivity analyses. Increasing $\lambda$ generally increased molecule--fingerprint alignment and cluster separation, but for MolCA the highest weight produced a representation that was more structurally organized while retrieval declined, consistent with over-constraining the embedding geometry. Fingerprint effects also followed the encoder split: Morgan produced the cleanest MolCA clustering, whereas MACCS produced stronger Atomas alignment and clustering metrics.

\paragraph{Generalization beyond the training signal.}
To test whether the learned organization extends beyond the exact fingerprint-similarity target, we also evaluated clustering by Murcko scaffold identity. Scaffold labels are not used in the \methodname\ objective, yet scaffold-based Calinski--Harabasz and Silhouette scores improved across all three backbones. Because scaffold frequencies in ChEBI-20 are long-tailed, Supplementary Table~4 reports both top-10 scaffold results and a filtered setting that removes very small clusters.

\begin{table}[ht]
\centering
\caption{\textbf{Supplementary Table~4 $|$ Scaffold-based clustering metrics on molecular and text embeddings.} Silhouette and Calinski--Harabasz computed with Murcko scaffold identity as cluster labels. Because scaffold frequencies are long-tailed, we report both \emph{Top 10 scaffolds} (matching the scaffold-coloured t-SNE panel in the main text) and \emph{$\geq 20$ samples} (which drops small-cluster noise). Scaffold identity is not used during \methodname\ training. Bold marks the column maximum within each model section.}
\label{tab:supp_scaffold_clustering}
\begin{tabular}{lcccccccc}
\toprule
& \multicolumn{4}{c}{\textbf{Top 10 scaffolds}} & \multicolumn{4}{c}{\textbf{Scaffolds with $\geq 20$ samples}} \\
\cmidrule(lr){2-5} \cmidrule(lr){6-9}
& \multicolumn{2}{c}{Mol Clustering} & \multicolumn{2}{c}{Text Clustering} & \multicolumn{2}{c}{Mol Clustering} & \multicolumn{2}{c}{Text Clustering} \\
\cmidrule(lr){2-3} \cmidrule(lr){4-5} \cmidrule(lr){6-7} \cmidrule(lr){8-9}
\textbf{Setting} & Sil & CH & Sil & CH & Sil & CH & Sil & CH \\
\midrule
\rowcolor{gray!10} \multicolumn{9}{l}{\textit{MolCA (Graph Embedding)}} \\
Base & $-$0.058 & 4.73 & $-$0.010 & 2.90 & $-$0.008 & 7.47 & $-$0.006 & 3.25 \\
$\lambda=1$ & $-$0.043 & 6.59 & $-$0.010 & 2.98 & $-$0.012 & 11.97 & +0.006 & 3.29 \\
$\lambda=100$ (Ours) & $-$0.067 & \textbf{13.71} & +0.008 & 4.49 & \textbf{+0.018} & \textbf{33.09} & +0.023 & 5.90 \\
$\lambda=1000$ & \textbf{$-$0.040} & 12.36 & \textbf{+0.017} & \textbf{5.22} & +0.011 & 28.13 & \textbf{+0.028} & \textbf{7.26} \\
\cdashline{1-9}
MACCS & $-$0.071 & 6.17 & +0.007 & 4.23 & $-$0.062 & 11.28 & +0.021 & 5.61 \\
RDKit & $-$0.041 & 9.60 & +0.002 & 4.27 & +0.007 & 20.13 & +0.021 & 5.44 \\
\midrule
\rowcolor{gray!10} \multicolumn{9}{l}{\textit{Atomas-B (SMILES Embedding)}} \\
Base & $-$0.012 & 5.04 & +0.003 & \textbf{4.45} & +0.007 & 7.66 & +0.016 & 6.20 \\
$\lambda=1$ & $-$0.002 & 5.95 & \textbf{+0.005} & 4.11 & +0.017 & 10.11 & +0.016 & 5.82 \\
$\lambda=100$ (Ours) & +0.019 & 7.58 & +0.001 & 4.08 & +0.042 & 13.12 & +0.016 & 5.82 \\
$\lambda=1000$ & \textbf{+0.047} & 8.77 & \textbf{+0.005} & 3.93 & \textbf{+0.063} & 15.78 & +0.015 & 5.90 \\
\cdashline{1-9}
MACCS & +0.030 & 9.17 & $-$0.001 & 4.38 & +0.059 & \textbf{17.39} & \textbf{+0.018} & \textbf{6.38} \\
RDKit & +0.036 & \textbf{9.19} & +0.001 & 4.18 & +0.054 & 16.62 & +0.016 & 6.02 \\
\midrule
\rowcolor{gray!10} \multicolumn{9}{l}{\textit{Atomas-L (SMILES Embedding)}} \\
Base & $-$0.012 & 4.90 & +0.000 & 4.12 & +0.006 & 7.83 & +0.013 & 5.75 \\
$\lambda=1$ & $-$0.011 & 5.37 & $-$0.000 & 4.69 & +0.013 & 8.79 & +0.019 & 6.94 \\
$\lambda=100$ (Ours) & +0.021 & 7.65 & +0.005 & 4.31 & +0.038 & 13.46 & +0.017 & 6.29 \\
$\lambda=1000$ & \textbf{+0.042} & 8.26 & \textbf{+0.009} & 4.29 & \textbf{+0.059} & 14.75 & +0.019 & 6.32 \\
\cdashline{1-9}
MACCS & +0.020 & 8.59 & $-$0.002 & 4.55 & +0.050 & 15.93 & +0.018 & 6.62 \\
RDKit & +0.034 & 9.30 & +0.007 & \textbf{5.09} & +0.056 & \textbf{17.14} & \textbf{+0.024} & \textbf{7.60} \\
\bottomrule
\end{tabular}
\end{table}

\begin{figure}[ht]
\centering
\includegraphics[width=\textwidth]{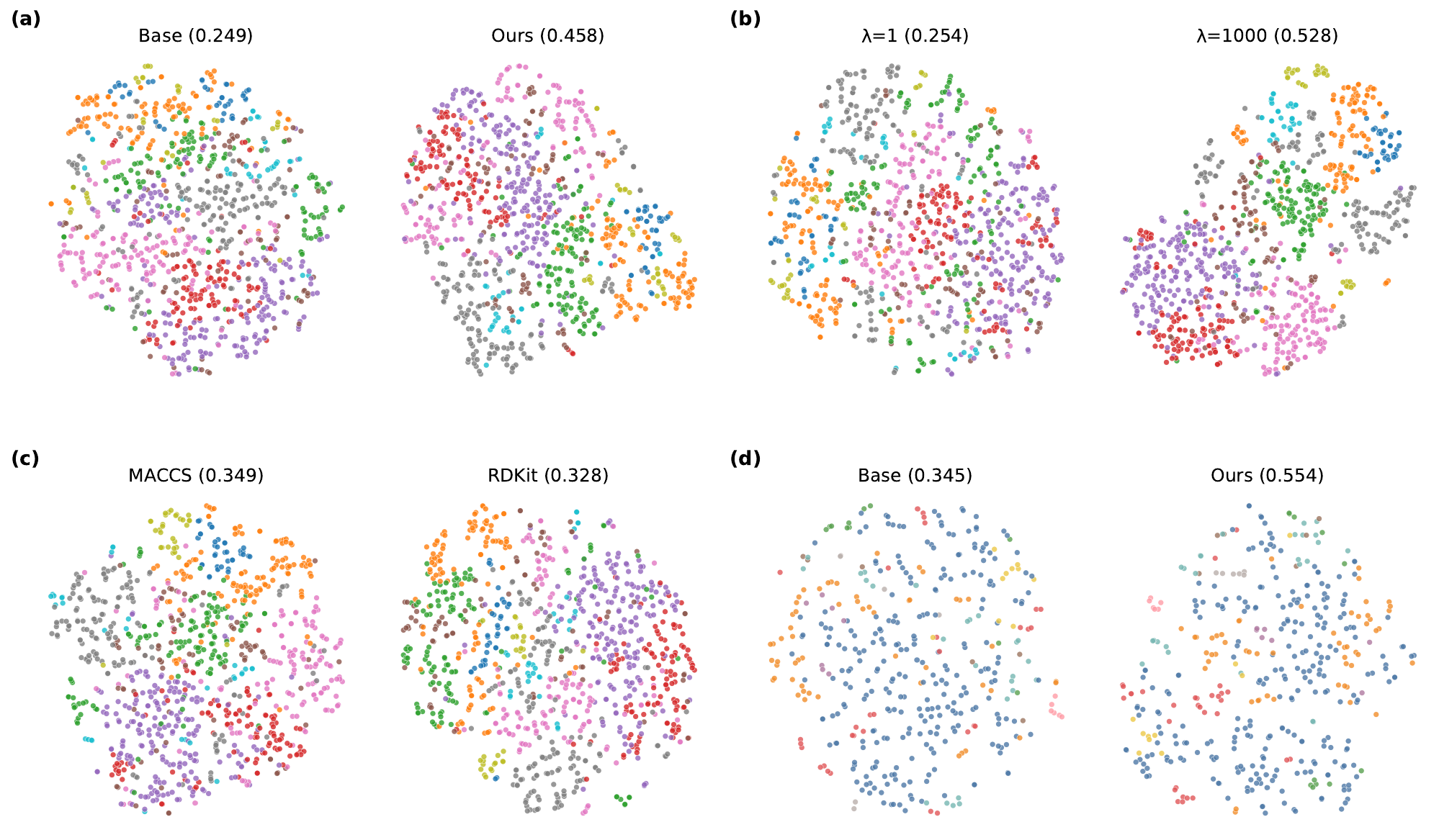}
\caption{\textbf{Supplementary Fig.~\ref{fig:supp_molca_text} $|$ t-SNE of MolCA text embeddings on the ChEBI-20 validation set.} \textbf{(a)} Base vs \methodname\ ($\lambda=100$, Morgan), colored by Tanimoto-based clusters. \textbf{(b)} Effect of regularization weight ($\lambda=1$ vs $\lambda=1000$). \textbf{(c)} Fingerprint type comparison (MACCS vs RDKit). \textbf{(d)} Base vs \methodname\ colored by Murcko scaffold identity. Parenthetical values denote text--fingerprint CKA.}
\label{fig:supp_molca_text}
\end{figure}

\begin{figure}[ht]
\centering
\includegraphics[width=\textwidth]{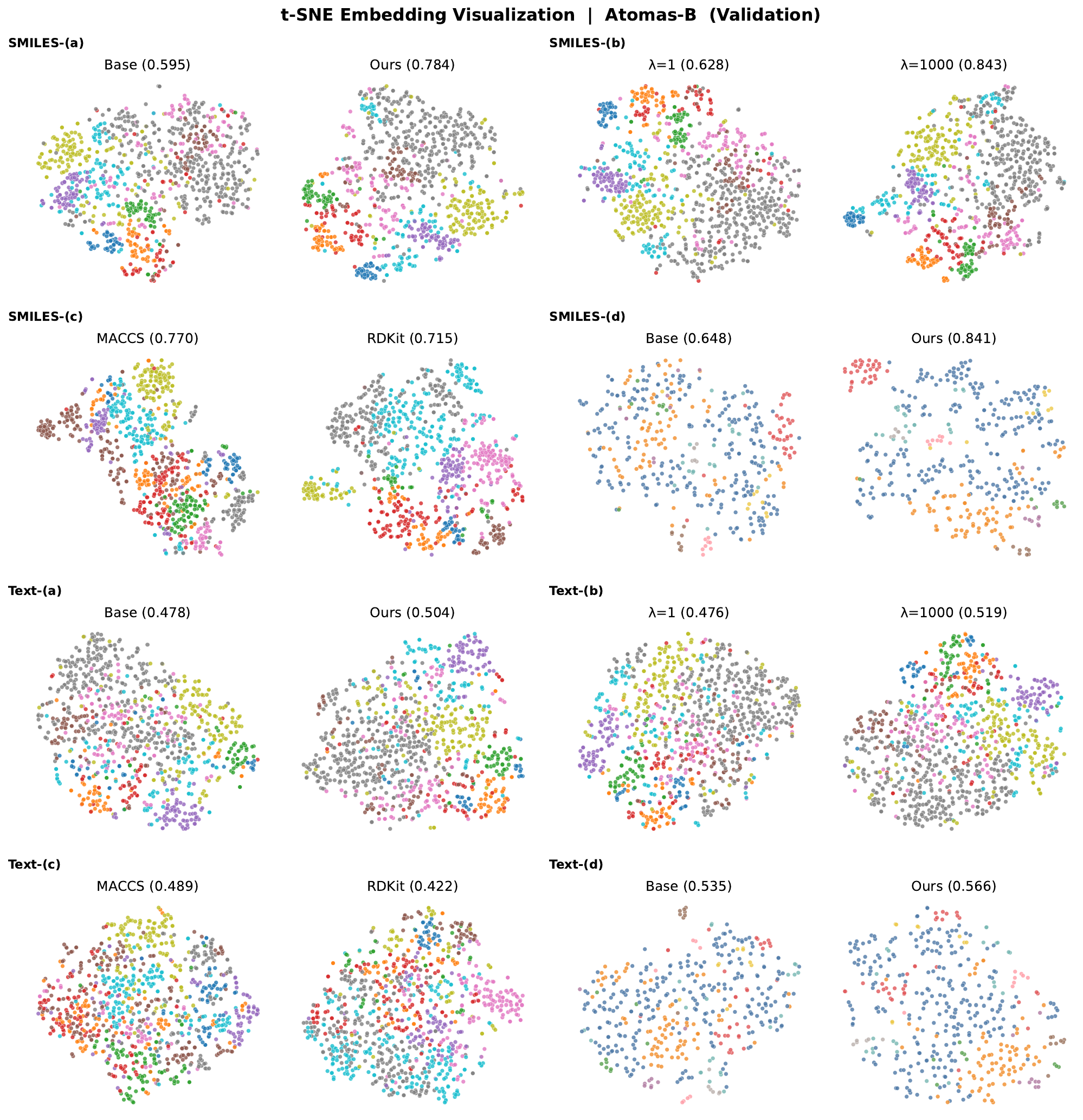}
\caption{\textbf{Supplementary Fig.~\ref{fig:supp_atomas_b_tsne} $|$ t-SNE of Atomas-B embeddings on the ChEBI-20 validation set.} Top two rows: SMILES embeddings; bottom two rows: text embeddings. Within each modality, \textbf{(a)} Base vs \methodname\ ($\lambda=100$, Morgan), colored by Tanimoto-based clusters. \textbf{(b)} Effect of regularization weight ($\lambda=1$ vs $\lambda=1000$). \textbf{(c)} Fingerprint type comparison (MACCS vs RDKit). \textbf{(d)} Base vs \methodname\ colored by Murcko scaffold identity. Parenthetical values denote embedding--fingerprint CKA.}
\label{fig:supp_atomas_b_tsne}
\end{figure}

\begin{figure}[ht]
\centering
\includegraphics[width=\textwidth]{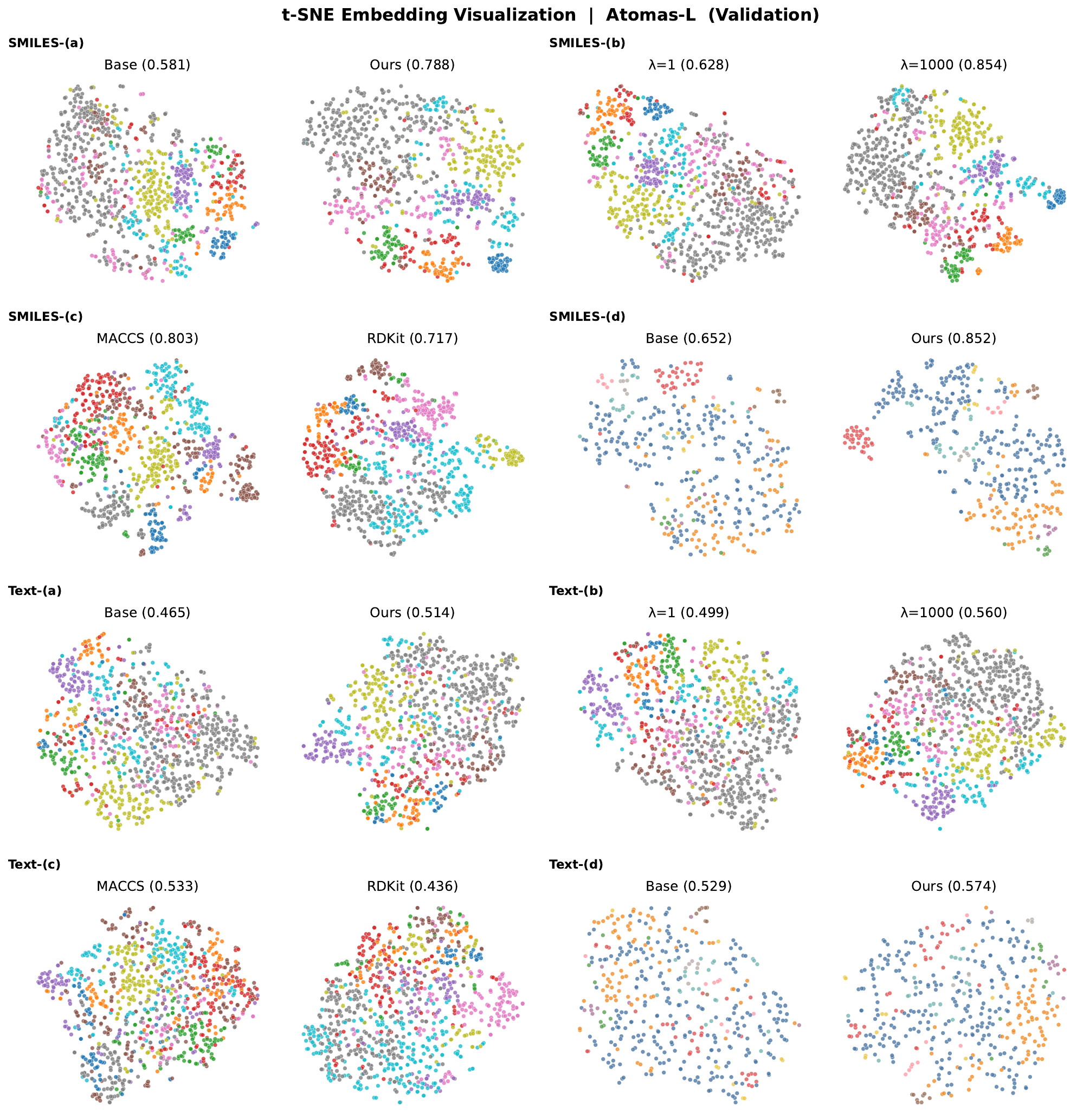}
\caption{\textbf{Supplementary Fig.~\ref{fig:supp_atomas_l_tsne} $|$ t-SNE of Atomas-L embeddings on the ChEBI-20 validation set.} Top two rows: SMILES embeddings; bottom two rows: text embeddings. Within each modality, \textbf{(a)} Base vs \methodname\ ($\lambda=100$, Morgan), colored by Tanimoto-based clusters. \textbf{(b)} Effect of regularization weight ($\lambda=1$ vs $\lambda=1000$). \textbf{(c)} Fingerprint type comparison (MACCS vs RDKit). \textbf{(d)} Base vs \methodname\ colored by Murcko scaffold identity. Parenthetical values denote embedding--fingerprint CKA.}
\label{fig:supp_atomas_l_tsne}
\end{figure}

\subsection*{Supplementary Note 3. Molecular generation and description experiments}

\paragraph{Scope and reporting.}
Generation experiments were conducted on ChEBI-20 using the native fine-tuning and decoding protocols of each backbone. The ablation compares each baseline with the corresponding \methodname-pretrained initialization. MolCA supports molecule-to-text description generation, whereas Atomas supports both molecule-to-text description generation and text-to-molecule SMILES generation. All generation values reported in the main text are single-run results rather than averages over repeated seeds.

\paragraph{MolCA description generation.}
For MolCA, baseline and \methodname\ stage-1 checkpoints were used to initialize the original stage-2 ChEBI-20 description-generation recipe. The Galactica-1.3B language model was fine-tuned with LoRA adapters using micro-batches of 16 and gradient accumulation of 4, corresponding to an effective batch size of 64, for 100 fine-tuning epochs. Evaluation used the ChEBI-20 test split with 3,300 molecule-description pairs. Captions were decoded with beam search using five beams and no sampling. The MolCA evaluator computes BLEU-2, BLEU-4, ROUGE-1, ROUGE-2 and ROUGE-L values on a 0--100 scale; these are the metrics shown in the main generation table.

\paragraph{Atomas description generation.}
Atomas description generation used the native \texttt{gentext} task, in which the SMILES encoder output conditions text generation. Atomas Base and Large were fine-tuned separately from the corresponding task-specific baseline or \methodname-pretrained checkpoint. Fine-tuning itself did not include the fingerprint-based CKA loss; the \methodname\ condition differs from the baseline by initialization from a fingerprint-CKA-pretrained checkpoint. For these description-generation runs, the \methodname\ pre-training checkpoint used Morgan fingerprints, a fingerprint queue size of 128 and $\lambda=1000$. Metrics were computed on the ChEBI-20 validation split using the original Atomas text-translation evaluator with SciBERT tokenization for BLEU-2/4 and ROUGE-1/2/L F-measures. The Atomas evaluator returns values on a 0--1 scale, which were multiplied by 100 for the main table.

\paragraph{Atomas text-to-molecule generation.}
Atomas text-to-molecule generation used the native \texttt{genmol} task and was fine-tuned separately from the \texttt{gentext} task. Base and Large variants were evaluated on the ChEBI-20 validation split. The \methodname-pretrained initialization for \texttt{genmol} used Morgan fingerprints, a fingerprint queue size of 128 and $\lambda=100$, which is distinct from the \texttt{gentext} checkpoint described above. SMILES-generation metrics follow the Atomas/MolT5 evaluator: BLEU is computed over character-tokenized SMILES strings; exact match is computed after converting generated and reference SMILES to InChI strings; Levenshtein is the mean SMILES edit distance; validity is the fraction of generated SMILES strings parsed by RDKit; and MACCS, RDKit and Morgan FTS values are mean Tanimoto similarities between generated and reference molecules using the corresponding fingerprint family. In the main table, BLEU, exact match, FTS and validity are multiplied by 100, whereas Levenshtein distance is reported on its raw scale.

\begin{figure}[ht]
\centering
\includegraphics[width=\textwidth]{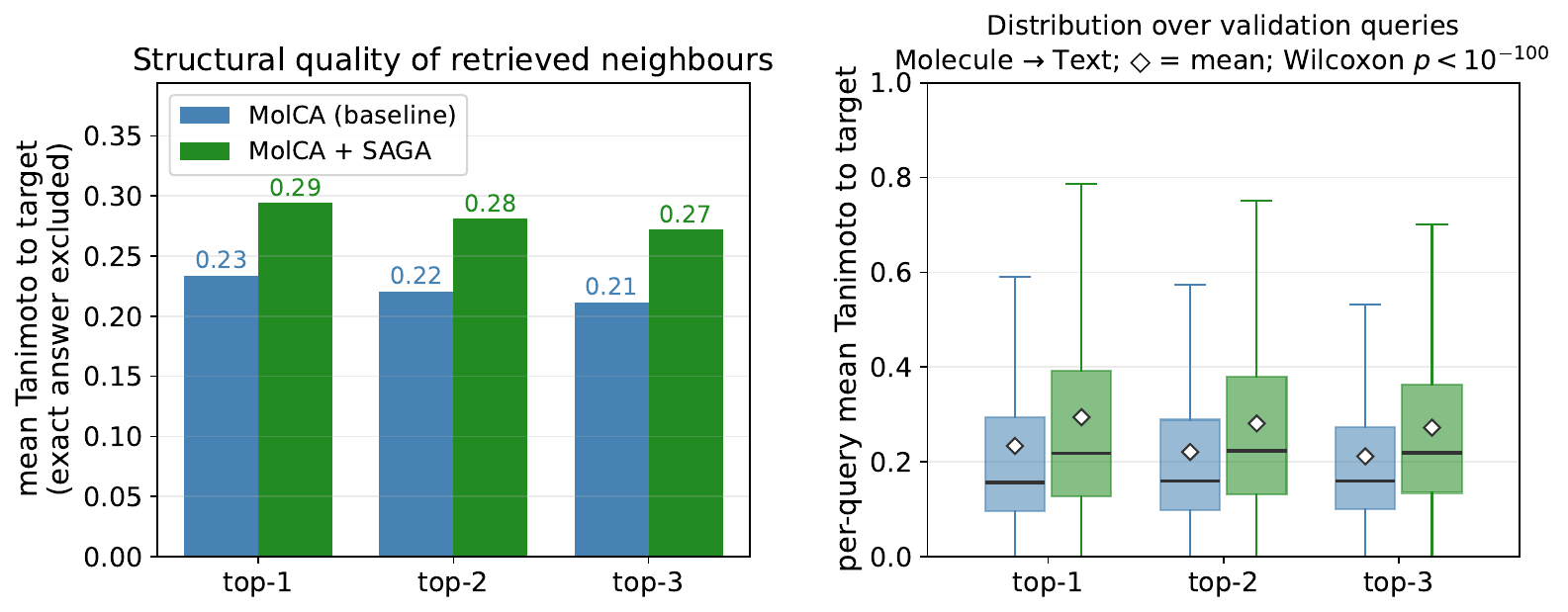}
\caption{\textbf{Supplementary Fig.~\ref{fig:supp_casestudy_molca_g2t} $|$ Structural quality of MolCA Molecule$\rightarrow$Text retrieval neighbours.}
Exclude-self structural neighbourhoods on the ChEBI-20 validation split ($n=3{,}301$). For each molecule query, the exact paired text was excluded before ranking, and the Tanimoto similarity between the query molecule and the molecules paired with the next retrieved texts was measured using Morgan fingerprints (radius 2, 2{,}048 bits). The mean top-3 Tanimoto similarity increased from 0.211 to 0.272 with \methodname, and \methodname\ retrieved structurally closer neighbours for 73.8\% of queries (one-sided Wilcoxon signed-rank test, $p=1.63\times10^{-229}$).}
\label{fig:supp_casestudy_molca_g2t}
\end{figure}

\begin{figure}[ht]
\centering
\includegraphics[width=\textwidth]{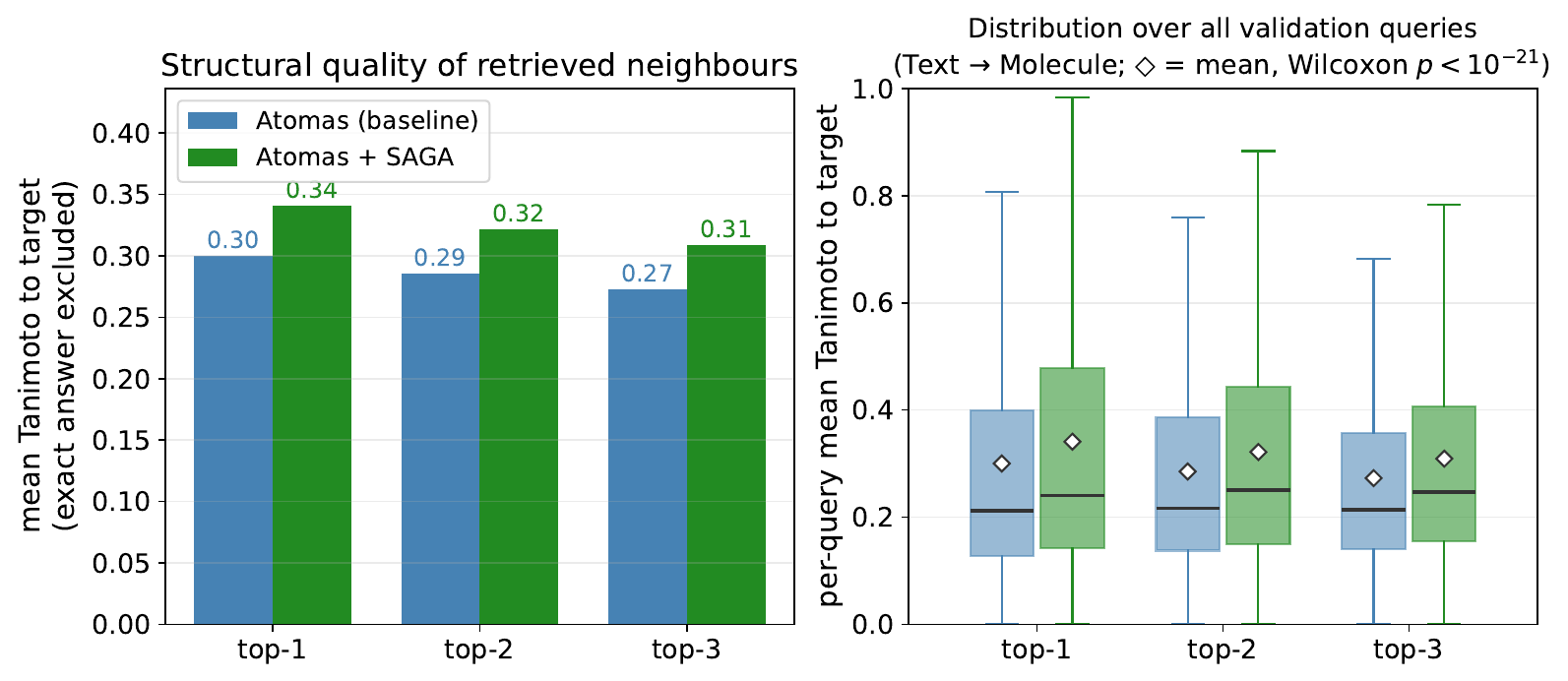}
\caption{\textbf{Supplementary Fig.~\ref{fig:supp_casestudy_atomas_b_t2g} $|$ Structural quality of Atomas-B Text$\rightarrow$Molecule retrieval neighbours.}
Exclude-self structural neighbourhoods on the ChEBI-20 validation split ($n=3{,}299$). The mean top-3 Tanimoto similarity increased from 0.273 to 0.309 with \methodname, and \methodname\ retrieved structurally closer neighbours for 57.8\% of queries (one-sided Wilcoxon signed-rank test, $p=5.77\times10^{-37}$).}
\label{fig:supp_casestudy_atomas_b_t2g}
\end{figure}

\begin{figure}[ht]
\centering
\includegraphics[width=\textwidth]{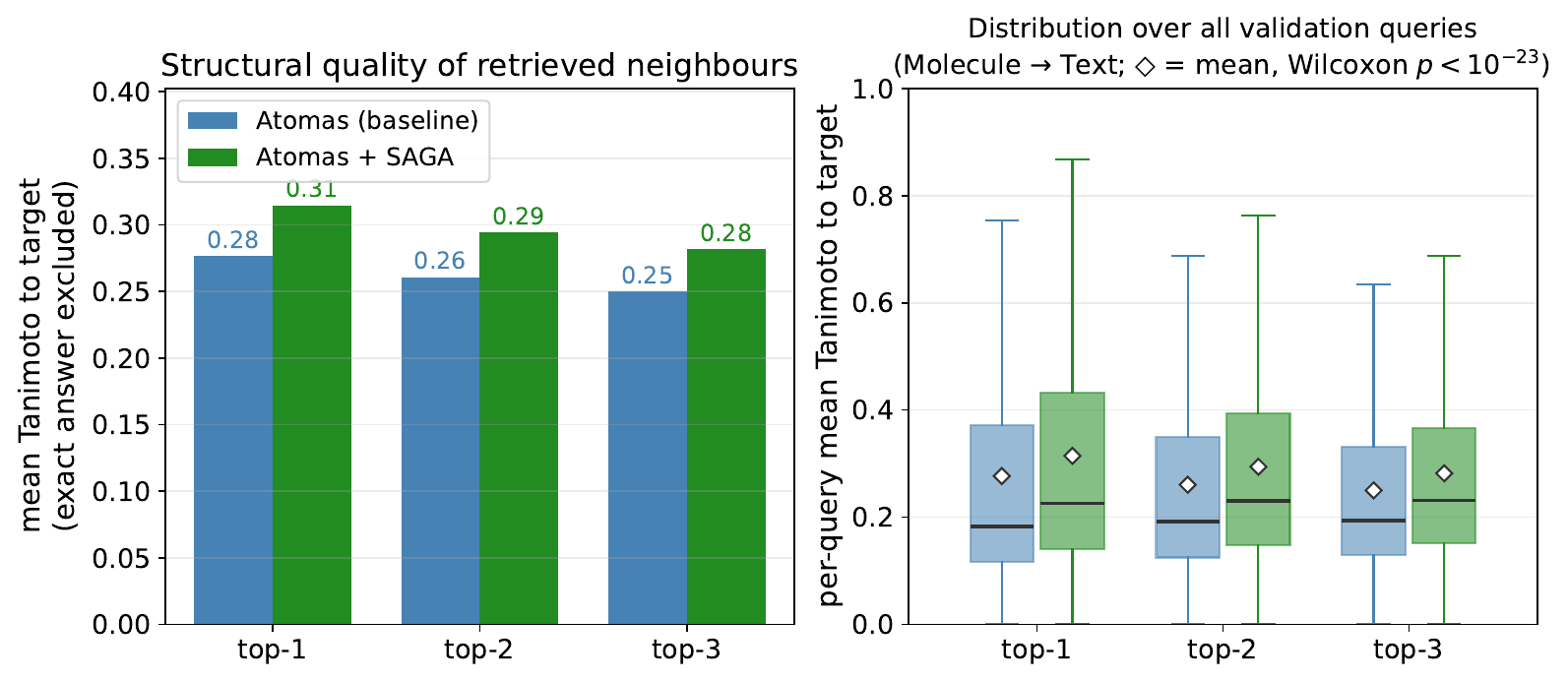}
\caption{\textbf{Supplementary Fig.~\ref{fig:supp_casestudy_atomas_b_g2t} $|$ Structural quality of Atomas-B Molecule$\rightarrow$Text retrieval neighbours.}
Exclude-self structural neighbourhoods on the ChEBI-20 validation split ($n=3{,}299$). The mean top-3 Tanimoto similarity increased from 0.250 to 0.282 with \methodname, and \methodname\ retrieved structurally closer neighbours for 60.0\% of queries (one-sided Wilcoxon signed-rank test, $p=1.91\times10^{-46}$).}
\label{fig:supp_casestudy_atomas_b_g2t}
\end{figure}

\begin{figure}[ht]
\centering
\includegraphics[width=\textwidth]{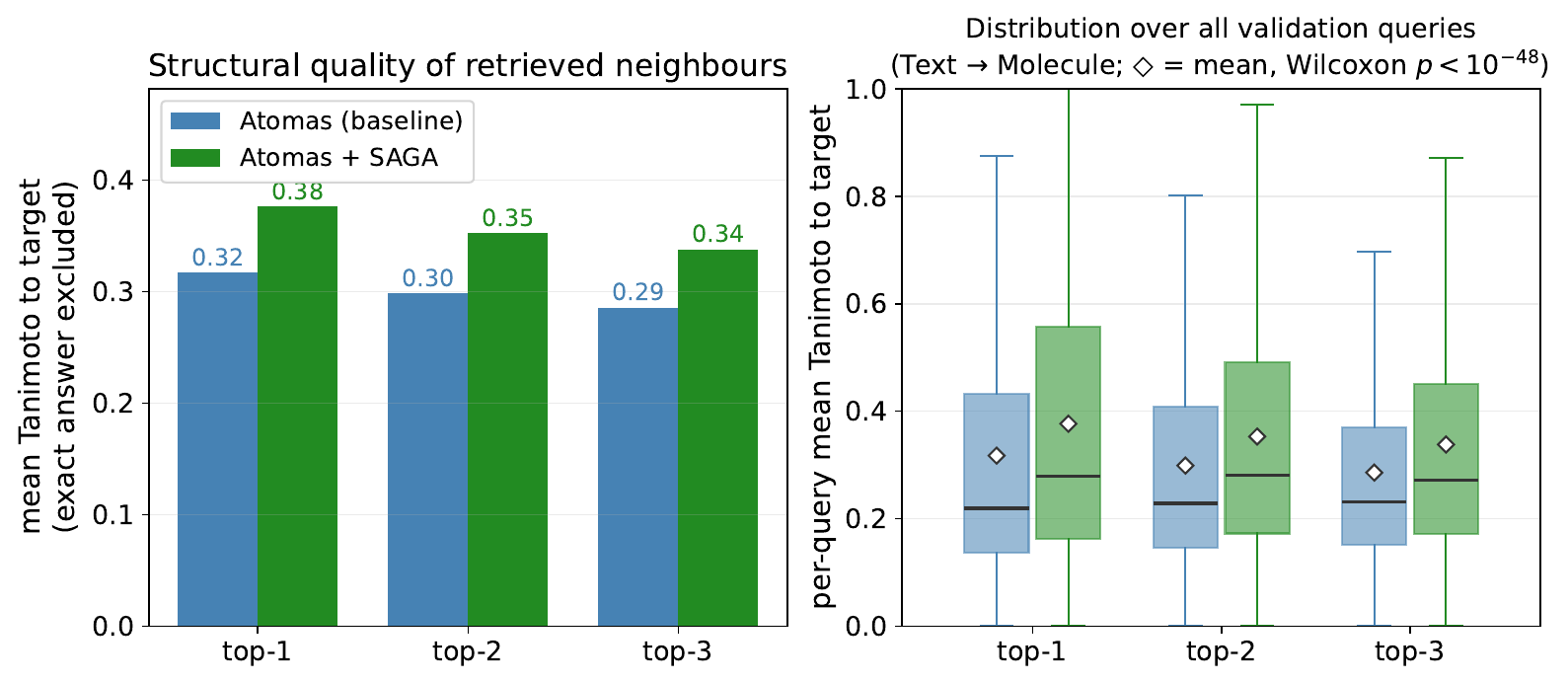}
\caption{\textbf{Supplementary Fig.~\ref{fig:supp_casestudy_atomas_l_t2g} $|$ Structural quality of Atomas-L Text$\rightarrow$Molecule retrieval neighbours.}
Exclude-self structural neighbourhoods on the ChEBI-20 validation split ($n=3{,}299$). The mean top-3 Tanimoto similarity increased from 0.286 to 0.338 with \methodname, and \methodname\ retrieved structurally closer neighbours for 63.9\% of queries (one-sided Wilcoxon signed-rank test, $p=9.99\times10^{-98}$).}
\label{fig:supp_casestudy_atomas_l_t2g}
\end{figure}

\begin{figure}[ht]
\centering
\includegraphics[width=\textwidth]{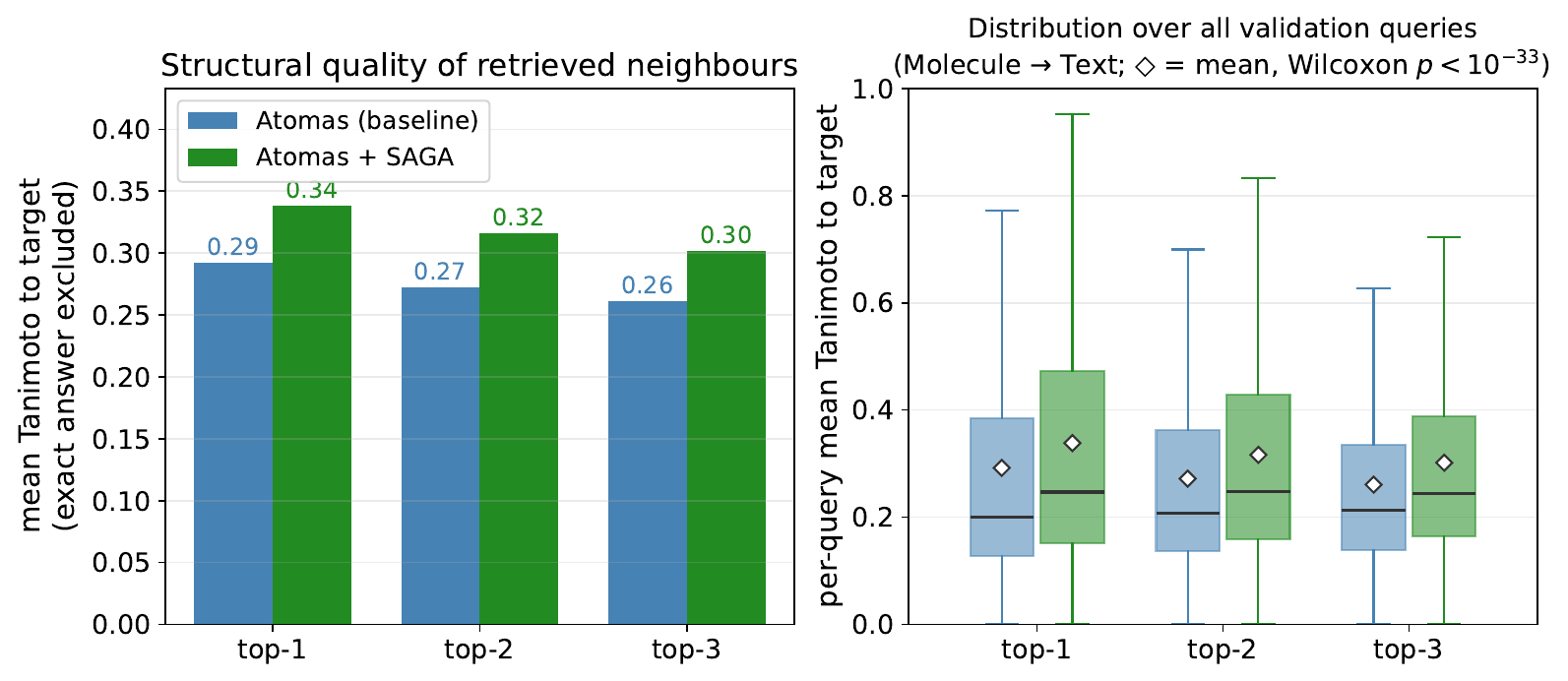}
\caption{\textbf{Supplementary Fig.~\ref{fig:supp_casestudy_atomas_l_g2t} $|$ Structural quality of Atomas-L Molecule$\rightarrow$Text retrieval neighbours.}
Exclude-self structural neighbourhoods on the ChEBI-20 validation split ($n=3{,}299$). The mean top-3 Tanimoto similarity increased from 0.261 to 0.302 with \methodname, and \methodname\ retrieved structurally closer neighbours for 62.5\% of queries (one-sided Wilcoxon signed-rank test, $p=2.79\times10^{-69}$).}
\label{fig:supp_casestudy_atomas_l_g2t}
\end{figure}

\definecolor{SAGABlue}{HTML}{4682B4}
\definecolor{SAGAGreen}{HTML}{228B22}
\definecolor{SAGARed}{HTML}{CC0000}
\definecolor{SAGAGray}{HTML}{9AA0A6}
\definecolor{SAGADark}{HTML}{222222}
\makeatletter
\setlength{\@fptop}{0pt}
\makeatother
\begin{figure}[p]
\centering

\begingroup
\setlength{\fboxsep}{2pt}
\setlength{\parindent}{0pt}
\footnotesize
{\fontsize{8.8}{9.6}\selectfont\bfseries Query 1 \quad Text$\rightarrow$Molecule \quad target CID 5282155}\par\vspace{1.2mm}
\begin{minipage}[t]{0.14\textwidth}
\vspace{0pt}
\fcolorbox{SAGADark}{white}{\includegraphics[width=\dimexpr\linewidth-2\fboxsep-2\fboxrule\relax]{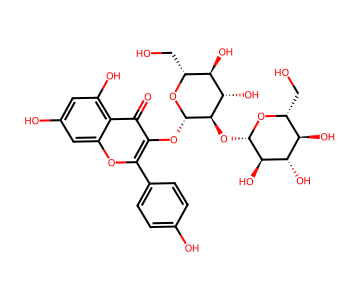}}
\end{minipage}\hfill
\begin{minipage}[t]{0.84\textwidth}
\vspace{0pt}
{\fontsize{8.0}{8.8}\selectfont\raggedright\sloppy The molecule is a sophoroside that is kaempferol attached to a beta-\allowbreak{}D-\allowbreak{}sophorosyl residue at position 3 via a glycosidic linkage. It has a role as a plant metabolite. It is a trihydroxyflavone and a sophoroside.\par}
\end{minipage}\par\vspace{2.4mm}
\noindent\begin{minipage}[t]{0.485\textwidth}{\fontsize{8.4}{9.2}\selectfont\textcolor{SAGABlue}{\textbf{MolCA (baseline): top-3 retrieved}}}\end{minipage}\hfill\begin{minipage}[t]{0.485\textwidth}{\fontsize{8.4}{9.2}\selectfont\textcolor{SAGAGreen}{\textbf{MolCA + \methodname: top-3 retrieved}}}\end{minipage}
\noindent\begin{minipage}[t]{0.485\textwidth}
\vspace{0pt}
\noindent\begin{minipage}[t]{0.285\linewidth}
\vspace{0pt}
\fcolorbox{SAGAGray}{white}{\includegraphics[width=\dimexpr\linewidth-2\fboxsep-2\fboxrule\relax]{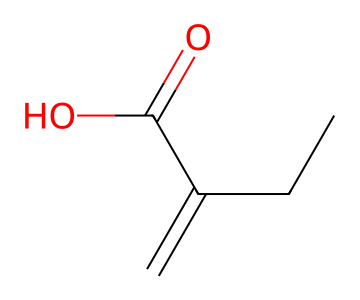}}
\end{minipage}\hfill
\begin{minipage}[t]{0.685\linewidth}
\vspace{0pt}
{\fontsize{8.0}{8.8}\selectfont\bfseries\textcolor{SAGABlue}{\#1 score=0.340 \quad T=0.04}}\\[-1pt]
{\fontsize{7.55}{8.35}\selectfont\raggedright\sloppy The molecule is an alpha,\allowbreak{}beta-\allowbreak{}unsaturated monocarboxylic acid that is acrylic acid in which the hydrogen at position 2 is substituted by an ethyl group. It has a role as a mammalian metabolite. It derives from an acrylic acid. It is a conjugate acid of a 2-\allowbreak{}ethylacrylate.\par}
\end{minipage}
\end{minipage}\hfill\begin{minipage}[t]{0.485\textwidth}
\vspace{0pt}
\noindent\begin{minipage}[t]{0.285\linewidth}
\vspace{0pt}
\fcolorbox{SAGAGreen}{white}{\includegraphics[width=\dimexpr\linewidth-2\fboxsep-2\fboxrule\relax]{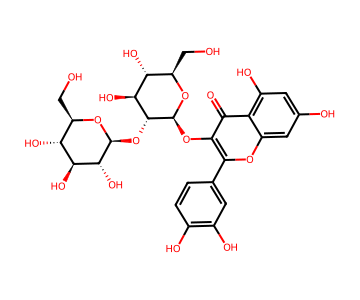}}
\end{minipage}\hfill
\begin{minipage}[t]{0.685\linewidth}
\vspace{0pt}
{\fontsize{8.0}{8.8}\selectfont\bfseries\textcolor{SAGAGreen}{\#1 score=0.507 \quad T=0.87}}\\[-1pt]
{\fontsize{7.55}{8.35}\selectfont\raggedright\sloppy The molecule is a quercetin \textcolor{SAGARed}{O-\allowbreak{}glucoside} that is quercetin attached to a \textcolor{SAGARed}{beta-\allowbreak{}D-\allowbreak{}sophorosyl} residue at position 3 via a \textcolor{SAGARed}{glycosidic} \textcolor{SAGARed}{linkage.} It has a role as an antioxidant and a plant metabolite. It is a \textcolor{SAGARed}{tetrahydroxyflavone} and a \textcolor{SAGARed}{sophoroside.} It is a conjugate acid of a quercetin \textcolor{SAGARed}{3-\allowbreak{}O-\allowbreak{}beta-\allowbreak{}D-\allowbreak{}glucosyl-\allowbreak{}(1-\allowbreak{}\textgreater{}2)\allowbreak{}-\allowbreak{}beta-\allowbreak{}D-\allowbreak{}glucoside(1-\allowbreak{})\allowbreak{}.}\par}
\end{minipage}
\end{minipage}\par\vspace{1.2mm}
\noindent\begin{minipage}[t]{0.485\textwidth}
\vspace{0pt}
\noindent\begin{minipage}[t]{0.285\linewidth}
\vspace{0pt}
\fcolorbox{SAGAGray}{white}{\includegraphics[width=\dimexpr\linewidth-2\fboxsep-2\fboxrule\relax]{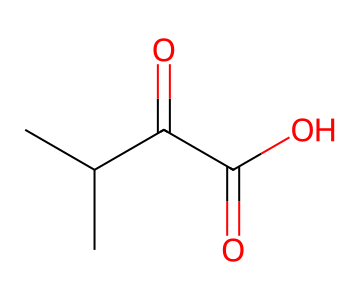}}
\end{minipage}\hfill
\begin{minipage}[t]{0.685\linewidth}
\vspace{0pt}
{\fontsize{8.0}{8.8}\selectfont\bfseries\textcolor{SAGABlue}{\#2 score=0.328 \quad T=0.04}}\\[-1pt]
{\fontsize{7.55}{8.35}\selectfont\raggedright\sloppy The molecule is a 2-\allowbreak{}oxo monocarboxylic acid that is the 2-\allowbreak{}oxo derivative of isovaleric acid. It has a role as a human metabolite and a Saccharomyces cerevisiae metabolite. It derives from a butyric acid. It is a conjugate acid of a 3-\allowbreak{}methyl-\allowbreak{}2-\allowbreak{}oxobutanoate. It is a tautomer of a 2-\allowbreak{}hydroxy-\allowbreak{}3-\allowbreak{}methyl-\allowbreak{}2-\allowbreak{}butenoic acid.\par}
\end{minipage}
\end{minipage}\hfill\begin{minipage}[t]{0.485\textwidth}
\vspace{0pt}
\noindent\begin{minipage}[t]{0.285\linewidth}
\vspace{0pt}
\fcolorbox{SAGAGreen}{white}{\includegraphics[width=\dimexpr\linewidth-2\fboxsep-2\fboxrule\relax]{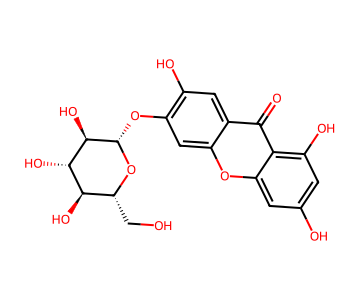}}
\end{minipage}\hfill
\begin{minipage}[t]{0.685\linewidth}
\vspace{0pt}
{\fontsize{8.0}{8.8}\selectfont\bfseries\textcolor{SAGAGreen}{\#2 score=0.499 \quad T=0.54}}\\[-1pt]
{\fontsize{7.55}{8.35}\selectfont\raggedright\sloppy The molecule is a xanthone \textcolor{SAGARed}{glycoside} that is norathyriol attached to a beta-\allowbreak{}D-\allowbreak{}glucopyranosyl residue at position 6 via a \textcolor{SAGARed}{glycosidic} \textcolor{SAGARed}{linkage.} It has a role as a plant metabolite. It is a \textcolor{SAGARed}{polyphenol,\allowbreak{}} a \textcolor{SAGARed}{beta-\allowbreak{}D-\allowbreak{}glucoside,\allowbreak{}} a monosaccharide derivative and a \textcolor{SAGARed}{xanthone} \textcolor{SAGARed}{glycoside.} It derives from a norathyriol.\par}
\end{minipage}
\end{minipage}\par\vspace{1.2mm}
\noindent\begin{minipage}[t]{0.485\textwidth}
\vspace{0pt}
\noindent\begin{minipage}[t]{0.285\linewidth}
\vspace{0pt}
\fcolorbox{SAGAGray}{white}{\includegraphics[width=\dimexpr\linewidth-2\fboxsep-2\fboxrule\relax]{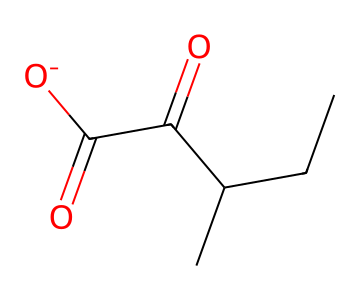}}
\end{minipage}\hfill
\begin{minipage}[t]{0.685\linewidth}
\vspace{0pt}
{\fontsize{8.0}{8.8}\selectfont\bfseries\textcolor{SAGABlue}{\#3 score=0.320 \quad T=0.06}}\\[-1pt]
{\fontsize{7.55}{8.35}\selectfont\raggedright\sloppy The molecule is a 2-\allowbreak{}oxo monocarboxylic acid anion that is the conjugate base of 3-\allowbreak{}methyl-\allowbreak{}2-\allowbreak{}oxovaleric acid,\allowbreak{} arising from deprotonation of the carboxy group;\allowbreak{} major species at pH 7.3. It has a role as a human metabolite. It derives from a valerate. It is a conjugate base of a 3-\allowbreak{}methyl-\allowbreak{}2-\allowbreak{}oxovaleric acid.\par}
\end{minipage}
\end{minipage}\hfill\begin{minipage}[t]{0.485\textwidth}
\vspace{0pt}
\noindent\begin{minipage}[t]{0.285\linewidth}
\vspace{0pt}
\fcolorbox{SAGAGreen}{white}{\includegraphics[width=\dimexpr\linewidth-2\fboxsep-2\fboxrule\relax]{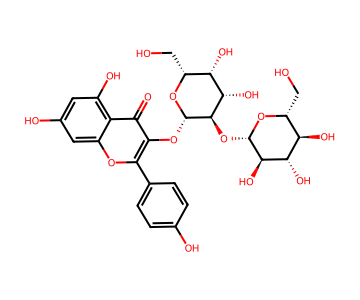}}
\end{minipage}\hfill
\begin{minipage}[t]{0.685\linewidth}
\vspace{0pt}
{\fontsize{8.0}{8.8}\selectfont\bfseries\textcolor{SAGAGreen}{\#3 score=0.469 \quad T=1.00}}\\[-1pt]
{\fontsize{7.55}{8.35}\selectfont\raggedright\sloppy The molecule is a glycosyloxyflavone that is \textcolor{SAGARed}{kaempferol} substituted by a 2-\allowbreak{}O-\allowbreak{}beta-\allowbreak{}D-\allowbreak{}glucopyranosyl-\allowbreak{}beta-\allowbreak{}D-\allowbreak{}galactopyranosyl residue at position 3 via a \textcolor{SAGARed}{glycosidic} \textcolor{SAGARed}{linkage.} It has a role as a plant metabolite. It is a glycosyloxyflavone,\allowbreak{} a \textcolor{SAGARed}{trihydroxyflavone} and a disaccharide derivative. It derives from a \textcolor{SAGARed}{kaempferol.}\par}
\end{minipage}
\end{minipage}\par\vspace{1.2mm}
\par\vspace{1.0mm}
{\fontsize{7.6}{8.3}\selectfont\textcolor{SAGARed}{red} = structural-class term shared with the query \quad
score = text-\allowbreak{}-\allowbreak{}graph match \quad
T = Morgan-fingerprint Tanimoto similarity to target}
\endgroup

\caption{\textbf{Supplementary Fig.~\ref{fig:supp_casestudy_text_overlap_molca} $|$ Description-level structural-class overlap for MolCA retrieval examples.} For representative Text$\rightarrow$Molecule queries discussed in the main text, each panel shows the target molecule, query description, and top-3 non-exact molecules retrieved by the MolCA baseline and MolCA$+$\methodname. Retrieved candidates are annotated with the native text--graph match score and their Morgan-fingerprint Tanimoto similarity to the target molecule. Red text denotes structural-class terms shared with the query description. The examples show that structurally closer \methodname\ neighbours also recover description terms corresponding to shared chemotypes or substructures.}
\label{fig:supp_casestudy_text_overlap_molca}
\end{figure}

\begin{figure}[p]
\ContinuedFloat
\centering

\begingroup
\setlength{\fboxsep}{2pt}
\setlength{\parindent}{0pt}
\footnotesize
{\fontsize{8.8}{9.6}\selectfont\bfseries Query 2 \quad Text$\rightarrow$Molecule \quad target CID 86289899}\par\vspace{1.2mm}
\begin{minipage}[t]{0.14\textwidth}
\vspace{0pt}
\fcolorbox{SAGADark}{white}{\includegraphics[width=\dimexpr\linewidth-2\fboxsep-2\fboxrule\relax]{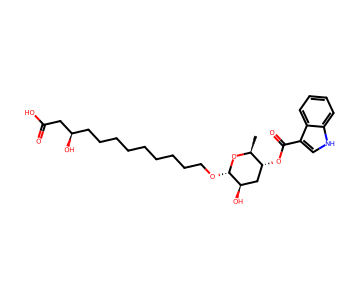}}
\end{minipage}\hfill
\begin{minipage}[t]{0.84\textwidth}
\vspace{0pt}
{\fontsize{8.0}{8.8}\selectfont\raggedright\sloppy The molecule is an omega-\allowbreak{}hydroxy fatty acid ascaroside that is bhos\#20 in which the hydroxy group at position 4 of the ascarylopyranose moiety has been has been converted to the corresponding 1H-\allowbreak{}indole-\allowbreak{}3-\allowbreak{}carboxylate ester. It is a metabolite of the nematode Caenorhabditis elegans. It has a role as a Caenorhabditis elegans metabolite. It is a 3-\allowbreak{}hydroxy carboxylic acid,\allowbreak{} a 4-\allowbreak{}O-\allowbreak{}(1H-\allowbreak{}indol-\allowbreak{}3-\allowbreak{}ylcarbonyl)\allowbreak{}ascaroside,\allowbreak{} an omega-\allowbreak{}hydroxy fatty acid ascaroside and a monocarboxylic acid. It derives from a bhos\#20 and a (3R)\allowbreak{}-\allowbreak{}3,\allowbreak{}12-\allowbreak{}dihydroxylauric acid.\par}
\end{minipage}\par\vspace{2.4mm}
\noindent\begin{minipage}[t]{0.485\textwidth}{\fontsize{8.4}{9.2}\selectfont\textcolor{SAGABlue}{\textbf{MolCA (baseline): top-3 retrieved}}}\end{minipage}\hfill\begin{minipage}[t]{0.485\textwidth}{\fontsize{8.4}{9.2}\selectfont\textcolor{SAGAGreen}{\textbf{MolCA + \methodname: top-3 retrieved}}}\end{minipage}
\noindent\begin{minipage}[t]{0.485\textwidth}
\vspace{0pt}
\noindent\begin{minipage}[t]{0.285\linewidth}
\vspace{0pt}
\fcolorbox{SAGAGray}{white}{\includegraphics[width=\dimexpr\linewidth-2\fboxsep-2\fboxrule\relax]{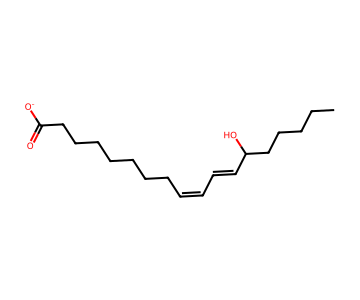}}
\end{minipage}\hfill
\begin{minipage}[t]{0.685\linewidth}
\vspace{0pt}
{\fontsize{8.0}{8.8}\selectfont\bfseries\textcolor{SAGABlue}{\#1 score=0.489 \quad T=0.14}}\\[-1pt]
{\fontsize{7.55}{8.35}\selectfont\raggedright\sloppy The molecule is a \textcolor{SAGARed}{HODE(1-\allowbreak{})\allowbreak{}} that is the conjugate base of \textcolor{SAGARed}{13-\allowbreak{}HODE,\allowbreak{}} obtained by deprotonation of the carboxy group;\allowbreak{} major species at pH 7.3. It derives from a \textcolor{SAGARed}{linoleate.} It is a conjugate base of a \textcolor{SAGARed}{13-\allowbreak{}HODE.}\par}
\end{minipage}
\end{minipage}\hfill\begin{minipage}[t]{0.485\textwidth}
\vspace{0pt}
\noindent\begin{minipage}[t]{0.285\linewidth}
\vspace{0pt}
\fcolorbox{SAGAGreen}{white}{\includegraphics[width=\dimexpr\linewidth-2\fboxsep-2\fboxrule\relax]{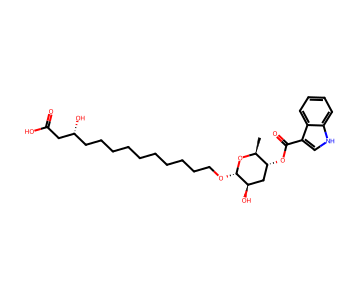}}
\end{minipage}\hfill
\begin{minipage}[t]{0.685\linewidth}
\vspace{0pt}
{\fontsize{8.0}{8.8}\selectfont\bfseries\textcolor{SAGAGreen}{\#1 score=0.536 \quad T=1.00}}\\[-1pt]
{\fontsize{7.55}{8.35}\selectfont\raggedright\sloppy The molecule is an \textcolor{SAGARed}{omega-\allowbreak{}hydroxy} \textcolor{SAGARed}{fatty} \textcolor{SAGARed}{acid} \textcolor{SAGARed}{ascaroside} that is bhos\#22 in which the hydroxy group at position 4 of the \textcolor{SAGARed}{ascarylopyranose} moiety has been has been converted to the corresponding \textcolor{SAGARed}{1H-\allowbreak{}indole-\allowbreak{}3-\allowbreak{}carboxylate} \textcolor{SAGARed}{ester.} It is a metabolite of the nematode Caenorhabditis elegans. It has a role as a Caenorhabditis elegans metabolite. It is a \textcolor{SAGARed}{3-\allowbreak{}hydroxy} \textcolor{SAGARed}{carboxylic} \textcolor{SAGARed}{acid,\allowbreak{}} a \textcolor{SAGARed}{4-\allowbreak{}O-\allowbreak{}(1H-\allowbreak{}indol-\allowbreak{}3-\allowbreak{}ylcarbonyl)\allowbreak{}ascaroside,\allowbreak{}} an \textcolor{SAGARed}{omega-\allowbreak{}hydroxy} \textcolor{SAGARed}{fatty} \textcolor{SAGARed}{acid} \textcolor{SAGARed}{ascaroside} and a \textcolor{SAGARed}{monocarboxylic} \textcolor{SAGARed}{acid.} It derives from a bhos\#22 and a (3R)\allowbreak{}-\allowbreak{}3,\allowbreak{}13-\allowbreak{}dihydroxytridecanoic acid.\par}
\end{minipage}
\end{minipage}\par\vspace{1.2mm}
\noindent\begin{minipage}[t]{0.485\textwidth}
\vspace{0pt}
\noindent\begin{minipage}[t]{0.285\linewidth}
\vspace{0pt}
\fcolorbox{SAGAGray}{white}{\includegraphics[width=\dimexpr\linewidth-2\fboxsep-2\fboxrule\relax]{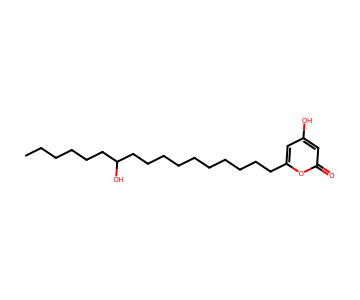}}
\end{minipage}\hfill
\begin{minipage}[t]{0.685\linewidth}
\vspace{0pt}
{\fontsize{8.0}{8.8}\selectfont\bfseries\textcolor{SAGABlue}{\#2 score=0.440 \quad T=0.18}}\\[-1pt]
{\fontsize{7.55}{8.35}\selectfont\raggedright\sloppy The molecule is a 6-\allowbreak{}alkyl-\allowbreak{}4-\allowbreak{}hydroxy-\allowbreak{}2H-\allowbreak{}pyran-\allowbreak{}2-\allowbreak{}one that is 4-\allowbreak{}hydroxy-\allowbreak{}2H-\allowbreak{}pyran-\allowbreak{}2-\allowbreak{}one in which the hydrogen at position 6 is replaced by an 11-\allowbreak{}hydroxyheptadecyl group. It is a 6-\allowbreak{}alkyl-\allowbreak{}4-\allowbreak{}hydroxy-\allowbreak{}2H-\allowbreak{}pyran-\allowbreak{}2-\allowbreak{}one and a secondary alcohol.\par}
\end{minipage}
\end{minipage}\hfill\begin{minipage}[t]{0.485\textwidth}
\vspace{0pt}
\noindent\begin{minipage}[t]{0.285\linewidth}
\vspace{0pt}
\fcolorbox{SAGAGreen}{white}{\includegraphics[width=\dimexpr\linewidth-2\fboxsep-2\fboxrule\relax]{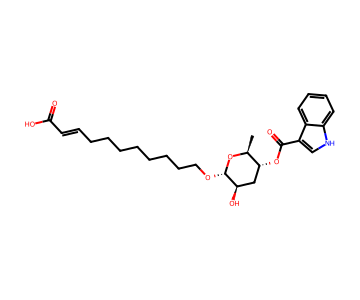}}
\end{minipage}\hfill
\begin{minipage}[t]{0.685\linewidth}
\vspace{0pt}
{\fontsize{8.0}{8.8}\selectfont\bfseries\textcolor{SAGAGreen}{\#2 score=0.534 \quad T=0.72}}\\[-1pt]
{\fontsize{7.55}{8.35}\selectfont\raggedright\sloppy The molecule is a \textcolor{SAGARed}{4-\allowbreak{}O-\allowbreak{}(1H-\allowbreak{}indol-\allowbreak{}3-\allowbreak{}ylcarbonyl)\allowbreak{}ascaroside} derived from (2E)\allowbreak{}-\allowbreak{}11-\allowbreak{}hydroxyundec-\allowbreak{}2-\allowbreak{}enoic acid. It is a metabolite of the nematode Caenorhabditis elegans. It has a role as a Caenorhabditis elegans metabolite. It is a \textcolor{SAGARed}{4-\allowbreak{}O-\allowbreak{}(1H-\allowbreak{}indol-\allowbreak{}3-\allowbreak{}ylcarbonyl)\allowbreak{}ascaroside,\allowbreak{}} an alpha,\allowbreak{}beta-\allowbreak{}unsaturated \textcolor{SAGARed}{monocarboxylic} \textcolor{SAGARed}{acid} and an \textcolor{SAGARed}{omega-\allowbreak{}hydroxy} \textcolor{SAGARed}{fatty} \textcolor{SAGARed}{acid} \textcolor{SAGARed}{ascaroside.} It derives from a (2E)\allowbreak{}-\allowbreak{}11-\allowbreak{}hydroxyundec-\allowbreak{}2-\allowbreak{}enoic acid and an oscr\#17.\par}
\end{minipage}
\end{minipage}\par\vspace{1.2mm}
\noindent\begin{minipage}[t]{0.485\textwidth}
\vspace{0pt}
\noindent\begin{minipage}[t]{0.285\linewidth}
\vspace{0pt}
\fcolorbox{SAGAGray}{white}{\includegraphics[width=\dimexpr\linewidth-2\fboxsep-2\fboxrule\relax]{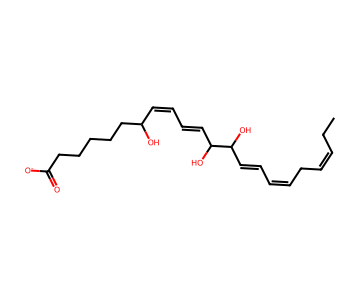}}
\end{minipage}\hfill
\begin{minipage}[t]{0.685\linewidth}
\vspace{0pt}
{\fontsize{8.0}{8.8}\selectfont\bfseries\textcolor{SAGABlue}{\#3 score=0.437 \quad T=0.15}}\\[-1pt]
{\fontsize{7.55}{8.35}\selectfont\raggedright\sloppy The molecule is a docosanoid anion that is the conjugate base of resolvin T2,\allowbreak{} obtained by deprotonation of the carboxy group;\allowbreak{} major species at pH 7.3. It is a \textcolor{SAGARed}{hydroxy} \textcolor{SAGARed}{fatty} \textcolor{SAGARed}{acid} anion and a docosanoid anion. It is a conjugate base of a resolvin T2.\par}
\end{minipage}
\end{minipage}\hfill\begin{minipage}[t]{0.485\textwidth}
\vspace{0pt}
\noindent\begin{minipage}[t]{0.285\linewidth}
\vspace{0pt}
\fcolorbox{SAGAGreen}{white}{\includegraphics[width=\dimexpr\linewidth-2\fboxsep-2\fboxrule\relax]{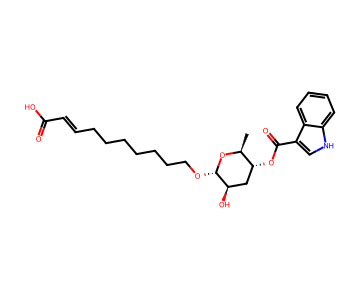}}
\end{minipage}\hfill
\begin{minipage}[t]{0.685\linewidth}
\vspace{0pt}
{\fontsize{8.0}{8.8}\selectfont\bfseries\textcolor{SAGAGreen}{\#3 score=0.531 \quad T=0.72}}\\[-1pt]
{\fontsize{7.55}{8.35}\selectfont\raggedright\sloppy The molecule is a \textcolor{SAGARed}{4-\allowbreak{}O-\allowbreak{}(1H-\allowbreak{}indol-\allowbreak{}3-\allowbreak{}ylcarbonyl)\allowbreak{}ascaroside} derived from (2E)\allowbreak{}-\allowbreak{}10-\allowbreak{}hydroxydec-\allowbreak{}2-\allowbreak{}enoic acid. It is a metabolite of the nematode Caenorhabditis elegans. It has a role as a Caenorhabditis elegans metabolite. It is a \textcolor{SAGARed}{4-\allowbreak{}O-\allowbreak{}(1H-\allowbreak{}indol-\allowbreak{}3-\allowbreak{}ylcarbonyl)\allowbreak{}ascaroside,\allowbreak{}} an alpha,\allowbreak{}beta-\allowbreak{}unsaturated \textcolor{SAGARed}{monocarboxylic} \textcolor{SAGARed}{acid} and an \textcolor{SAGARed}{omega-\allowbreak{}hydroxy} \textcolor{SAGARed}{fatty} \textcolor{SAGARed}{acid} \textcolor{SAGARed}{ascaroside.} It derives from an (E)\allowbreak{}-\allowbreak{}10-\allowbreak{}hydroxydec-\allowbreak{}2-\allowbreak{}enoic acid and an oscr\#15.\par}
\end{minipage}
\end{minipage}\par\vspace{1.2mm}
\par\vspace{1.0mm}
{\fontsize{7.6}{8.3}\selectfont\textcolor{SAGARed}{red} = structural-class term shared with the query \quad
score = text-\allowbreak{}-\allowbreak{}graph match \quad
T = Morgan-fingerprint Tanimoto similarity to target}
\endgroup

\caption[]{\textbf{Supplementary Fig.~\ref{fig:supp_casestudy_text_overlap_molca} $|$ Description-level structural-class overlap for MolCA retrieval examples, continued.} Same layout and annotation scheme as the first panel; red text marks structural-class terms shared with the query description.}
\end{figure}

\begin{figure}[p]
\ContinuedFloat
\centering

\begingroup
\setlength{\fboxsep}{2pt}
\setlength{\parindent}{0pt}
\footnotesize
{\fontsize{8.8}{9.6}\selectfont\bfseries Query 3 \quad Text$\rightarrow$Molecule \quad target CID 449179}\par\vspace{1.2mm}
\begin{minipage}[t]{0.14\textwidth}
\vspace{0pt}
\fcolorbox{SAGADark}{white}{\includegraphics[width=\dimexpr\linewidth-2\fboxsep-2\fboxrule\relax]{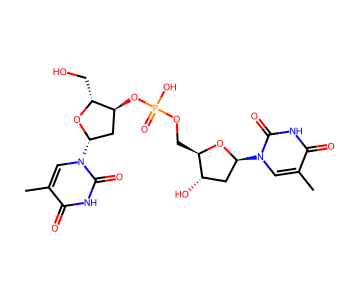}}
\end{minipage}\hfill
\begin{minipage}[t]{0.84\textwidth}
\vspace{0pt}
{\fontsize{8.0}{8.8}\selectfont\raggedright\sloppy The molecule is an oligonucleotide comprised of two thymidine residues connected via a 3'-\allowbreak{}\textgreater{}5' phosphodiester linkage. It contains a dTMP 3'-\allowbreak{}end residue.\par}
\end{minipage}\par\vspace{2.4mm}
\noindent\begin{minipage}[t]{0.485\textwidth}{\fontsize{8.4}{9.2}\selectfont\textcolor{SAGABlue}{\textbf{MolCA (baseline): top-3 retrieved}}}\end{minipage}\hfill\begin{minipage}[t]{0.485\textwidth}{\fontsize{8.4}{9.2}\selectfont\textcolor{SAGAGreen}{\textbf{MolCA + \methodname: top-3 retrieved}}}\end{minipage}
\noindent\begin{minipage}[t]{0.485\textwidth}
\vspace{0pt}
\noindent\begin{minipage}[t]{0.285\linewidth}
\vspace{0pt}
\fcolorbox{SAGAGray}{white}{\includegraphics[width=\dimexpr\linewidth-2\fboxsep-2\fboxrule\relax]{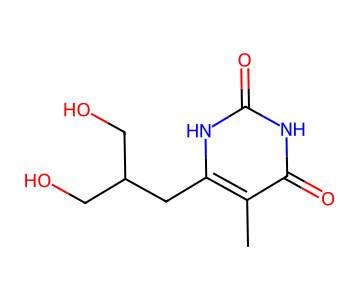}}
\end{minipage}\hfill
\begin{minipage}[t]{0.685\linewidth}
\vspace{0pt}
{\fontsize{8.0}{8.8}\selectfont\bfseries\textcolor{SAGABlue}{\#1 score=0.403 \quad T=0.15}}\\[-1pt]
{\fontsize{7.55}{8.35}\selectfont\raggedright\sloppy The molecule is a \textcolor{SAGARed}{pyrimidone} that is \textcolor{SAGARed}{thymine} in which the hydrogen at position 6 is substituted by a 1,\allowbreak{}3-\allowbreak{}dihydroxyisobutyl group. It derives from a \textcolor{SAGARed}{thymine.}\par}
\end{minipage}
\end{minipage}\hfill\begin{minipage}[t]{0.485\textwidth}
\vspace{0pt}
\noindent\begin{minipage}[t]{0.285\linewidth}
\vspace{0pt}
\fcolorbox{SAGAGreen}{white}{\includegraphics[width=\dimexpr\linewidth-2\fboxsep-2\fboxrule\relax]{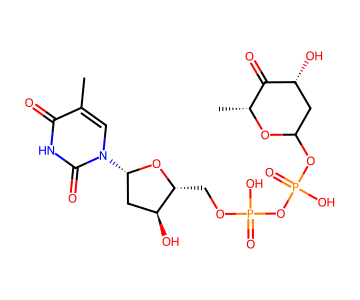}}
\end{minipage}\hfill
\begin{minipage}[t]{0.685\linewidth}
\vspace{0pt}
{\fontsize{8.0}{8.8}\selectfont\bfseries\textcolor{SAGAGreen}{\#1 score=0.628 \quad T=0.65}}\\[-1pt]
{\fontsize{7.55}{8.35}\selectfont\raggedright\sloppy The molecule is a \textcolor{SAGARed}{dTDP-\allowbreak{}sugar} having 4-\allowbreak{}dehydro-\allowbreak{}2,\allowbreak{}6-\allowbreak{}dideoxy-\allowbreak{}D-\allowbreak{}glucose as the sugar component. It is a \textcolor{SAGARed}{dTDP-\allowbreak{}sugar} and a secondary alpha-\allowbreak{}hydroxy ketone. It derives from a \textcolor{SAGARed}{dTDP-\allowbreak{}D-\allowbreak{}glucose.}\par}
\end{minipage}
\end{minipage}\par\vspace{1.2mm}
\noindent\begin{minipage}[t]{0.485\textwidth}
\vspace{0pt}
\noindent\begin{minipage}[t]{0.285\linewidth}
\vspace{0pt}
\fcolorbox{SAGAGray}{white}{\includegraphics[width=\dimexpr\linewidth-2\fboxsep-2\fboxrule\relax]{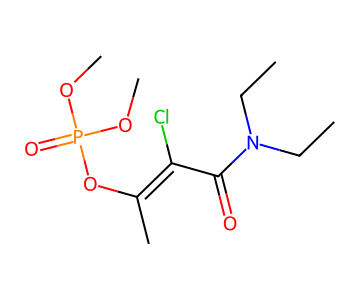}}
\end{minipage}\hfill
\begin{minipage}[t]{0.685\linewidth}
\vspace{0pt}
{\fontsize{8.0}{8.8}\selectfont\bfseries\textcolor{SAGABlue}{\#2 score=0.344 \quad T=0.12}}\\[-1pt]
{\fontsize{7.55}{8.35}\selectfont\raggedright\sloppy The molecule is a trialkyl \textcolor{SAGARed}{phosphate,\allowbreak{}} an \textcolor{SAGARed}{organophosphate} insecticide,\allowbreak{} an organochlorine insecticide and an \textcolor{SAGARed}{organophosphate} nematicide. It has a role as an EC 3.1.1.7 (acetylcholinesterase)\allowbreak{} inhibitor,\allowbreak{} an EC 3.1.1.8 (cholinesterase)\allowbreak{} inhibitor,\allowbreak{} an acaricide and an agrochemical. It derives from a N,\allowbreak{}N-\allowbreak{}diethyl-\allowbreak{}3-\allowbreak{}hydroxybut-\allowbreak{}2-\allowbreak{}enamide.\par}
\end{minipage}
\end{minipage}\hfill\begin{minipage}[t]{0.485\textwidth}
\vspace{0pt}
\noindent\begin{minipage}[t]{0.285\linewidth}
\vspace{0pt}
\fcolorbox{SAGAGreen}{white}{\includegraphics[width=\dimexpr\linewidth-2\fboxsep-2\fboxrule\relax]{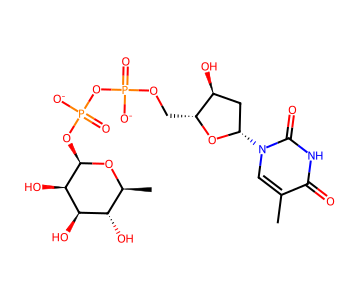}}
\end{minipage}\hfill
\begin{minipage}[t]{0.685\linewidth}
\vspace{0pt}
{\fontsize{8.0}{8.8}\selectfont\bfseries\textcolor{SAGAGreen}{\#2 score=0.570 \quad T=0.58}}\\[-1pt]
{\fontsize{7.55}{8.35}\selectfont\raggedright\sloppy The molecule is dianion of \textcolor{SAGARed}{dTDP-\allowbreak{}6-\allowbreak{}deoxy-\allowbreak{}beta-\allowbreak{}L-\allowbreak{}mannose} arising from deprotonation of both free OH groups of the diphosphate. It is a conjugate base of a \textcolor{SAGARed}{dTDP-\allowbreak{}beta-\allowbreak{}L-\allowbreak{}rhamnose.}\par}
\end{minipage}
\end{minipage}\par\vspace{1.2mm}
\noindent\begin{minipage}[t]{0.485\textwidth}
\vspace{0pt}
\noindent\begin{minipage}[t]{0.285\linewidth}
\vspace{0pt}
\fcolorbox{SAGAGray}{white}{\includegraphics[width=\dimexpr\linewidth-2\fboxsep-2\fboxrule\relax]{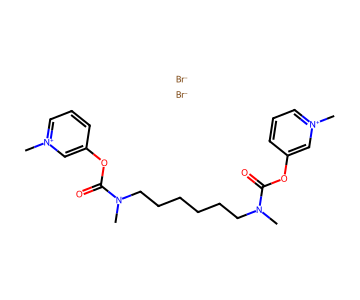}}
\end{minipage}\hfill
\begin{minipage}[t]{0.685\linewidth}
\vspace{0pt}
{\fontsize{8.0}{8.8}\selectfont\bfseries\textcolor{SAGABlue}{\#3 score=0.334 \quad T=0.08}}\\[-1pt]
{\fontsize{7.55}{8.35}\selectfont\raggedright\sloppy The molecule is an organic bromide salt of distigmine. It is an anticholinesterase drug used for the treatment of myasthenia gravis and postoperative urinary retention. It has a role as an EC 3.1.1.8 (cholinesterase)\allowbreak{} inhibitor. It is a pyridinium salt and an organic bromide salt. It contains a distigmine.\par}
\end{minipage}
\end{minipage}\hfill\begin{minipage}[t]{0.485\textwidth}
\vspace{0pt}
\noindent\begin{minipage}[t]{0.285\linewidth}
\vspace{0pt}
\fcolorbox{SAGAGreen}{white}{\includegraphics[width=\dimexpr\linewidth-2\fboxsep-2\fboxrule\relax]{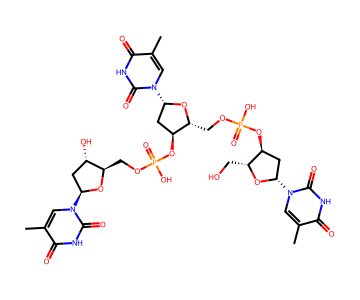}}
\end{minipage}\hfill
\begin{minipage}[t]{0.685\linewidth}
\vspace{0pt}
{\fontsize{8.0}{8.8}\selectfont\bfseries\textcolor{SAGAGreen}{\#3 score=0.562 \quad T=0.98}}\\[-1pt]
{\fontsize{7.55}{8.35}\selectfont\raggedright\sloppy The molecule is an \textcolor{SAGARed}{oligonucleotide} composed of three \textcolor{SAGARed}{deoxythymidylic} \textcolor{SAGARed}{acid} residues connected by \textcolor{SAGARed}{3'-\allowbreak{}\textgreater{}5'} \textcolor{SAGARed}{phosphodiester} \textcolor{SAGARed}{linkages.} It contains a \textcolor{SAGARed}{thymidine} \textcolor{SAGARed}{5'-\allowbreak{}monophosphate} residue.\par}
\end{minipage}
\end{minipage}\par\vspace{1.2mm}
\par\vspace{1.0mm}
{\fontsize{7.6}{8.3}\selectfont\textcolor{SAGARed}{red} = structural-class term shared with the query \quad
score = text-\allowbreak{}-\allowbreak{}graph match \quad
T = Morgan-fingerprint Tanimoto similarity to target}
\endgroup

\caption[]{\textbf{Supplementary Fig.~\ref{fig:supp_casestudy_text_overlap_molca} $|$ Description-level structural-class overlap for MolCA retrieval examples, continued.} Same layout and annotation scheme as the first panel; red text marks structural-class terms shared with the query description.}
\end{figure}

\begin{figure}[p]
\centering

\begingroup
\setlength{\fboxsep}{2pt}
\setlength{\parindent}{0pt}
\footnotesize
{\fontsize{8.8}{9.6}\selectfont\bfseries Query 1 \quad Text$\rightarrow$Molecule \quad target CID 129011046}\par\vspace{1.2mm}
\begin{minipage}[t]{0.14\textwidth}
\vspace{0pt}
\fcolorbox{SAGADark}{white}{\includegraphics[width=\dimexpr\linewidth-2\fboxsep-2\fboxrule\relax]{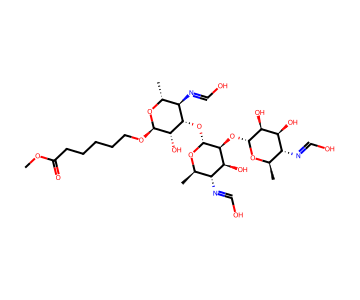}}
\end{minipage}\hfill
\begin{minipage}[t]{0.84\textwidth}
\vspace{0pt}
{\fontsize{8.0}{8.8}\selectfont\raggedright\sloppy The molecule is a glycoside that consists of three N-\allowbreak{}formyl-\allowbreak{}alpha-\allowbreak{}D-\allowbreak{}perosamine residues linked sequentially (1-\allowbreak{}\textgreater{}2)\allowbreak{} and (1-\allowbreak{}\textgreater{}3)\allowbreak{} and linked at the reducing end glycosidically to a 5-\allowbreak{}(methoxycarbonyl)\allowbreak{}pentyl group. It is a glycoside,\allowbreak{} a methyl ester and a trisaccharide derivative. It derives from an alpha-\allowbreak{}D-\allowbreak{}Rhap4NFo-\allowbreak{}(1-\allowbreak{}\textgreater{}2)\allowbreak{}-\allowbreak{}alpha-\allowbreak{}D-\allowbreak{}Rhap4NFo-\allowbreak{}(1-\allowbreak{}\textgreater{}3)\allowbreak{}-\allowbreak{}alpha-\allowbreak{}D-\allowbreak{}Rhap4NFo.\par}
\end{minipage}\par\vspace{2.4mm}
\noindent\begin{minipage}[t]{0.485\textwidth}{\fontsize{8.4}{9.2}\selectfont\textcolor{SAGABlue}{\textbf{Atomas-\allowbreak{}B (baseline): top-3 retrieved}}}\end{minipage}\hfill\begin{minipage}[t]{0.485\textwidth}{\fontsize{8.4}{9.2}\selectfont\textcolor{SAGAGreen}{\textbf{Atomas-\allowbreak{}B + \methodname: top-3 retrieved}}}\end{minipage}
\noindent\begin{minipage}[t]{0.485\textwidth}
\vspace{0pt}
\noindent\begin{minipage}[t]{0.285\linewidth}
\vspace{0pt}
\fcolorbox{SAGAGray}{white}{\includegraphics[width=\dimexpr\linewidth-2\fboxsep-2\fboxrule\relax]{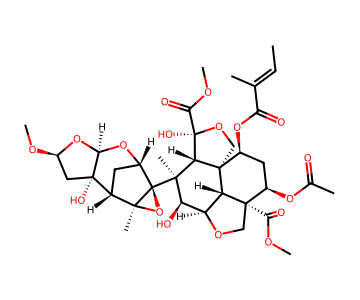}}
\end{minipage}\hfill
\begin{minipage}[t]{0.685\linewidth}
\vspace{0pt}
{\fontsize{8.0}{8.8}\selectfont\bfseries\textcolor{SAGABlue}{\#1 score=0.329 \quad T=0.11}}\\[-1pt]
{\fontsize{7.55}{8.35}\selectfont\raggedright\sloppy The molecule is a limonoid found in Azadirachta indica. It has a role as a plant metabolite. It is a limonoid,\allowbreak{} an acetate ester,\allowbreak{} an epoxide,\allowbreak{} a tertiary alcohol,\allowbreak{} a secondary alcohol,\allowbreak{} a \textcolor{SAGARed}{methyl} \textcolor{SAGARed}{ester} and an enoate ester. It derives from a tiglic acid.\par}
\end{minipage}
\end{minipage}\hfill\begin{minipage}[t]{0.485\textwidth}
\vspace{0pt}
\noindent\begin{minipage}[t]{0.285\linewidth}
\vspace{0pt}
\fcolorbox{SAGAGreen}{white}{\includegraphics[width=\dimexpr\linewidth-2\fboxsep-2\fboxrule\relax]{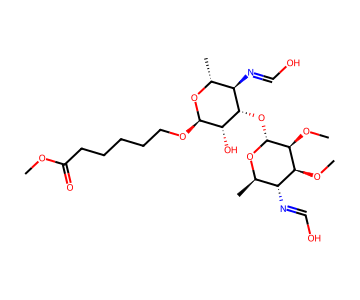}}
\end{minipage}\hfill
\begin{minipage}[t]{0.685\linewidth}
\vspace{0pt}
{\fontsize{8.0}{8.8}\selectfont\bfseries\textcolor{SAGAGreen}{\#1 score=0.337 \quad T=0.82}}\\[-1pt]
{\fontsize{7.55}{8.35}\selectfont\raggedright\sloppy The molecule is a \textcolor{SAGARed}{glycoside} that consists of an \textcolor{SAGARed}{N-\allowbreak{}formyl-\allowbreak{}2,\allowbreak{}3-\allowbreak{}di-\allowbreak{}O-\allowbreak{}methyl-\allowbreak{}alpha-\allowbreak{}D-\allowbreak{}perosamine} \textcolor{SAGARed}{residue} linked \textcolor{SAGARed}{(1-\allowbreak{}\textgreater{}3)\allowbreak{}} to an \textcolor{SAGARed}{N-\allowbreak{}formyl-\allowbreak{}alpha-\allowbreak{}D-\allowbreak{}perosamine} \textcolor{SAGARed}{residue} which is linked glycosidically to a \textcolor{SAGARed}{5-\allowbreak{}(methoxycarbonyl)\allowbreak{}pentyl} group. It is a disaccharide derivative,\allowbreak{} a \textcolor{SAGARed}{glycoside} and a \textcolor{SAGARed}{methyl} \textcolor{SAGARed}{ester.}\par}
\end{minipage}
\end{minipage}\par\vspace{1.2mm}
\noindent\begin{minipage}[t]{0.485\textwidth}
\vspace{0pt}
\noindent\begin{minipage}[t]{0.285\linewidth}
\vspace{0pt}
\fcolorbox{SAGAGray}{white}{\includegraphics[width=\dimexpr\linewidth-2\fboxsep-2\fboxrule\relax]{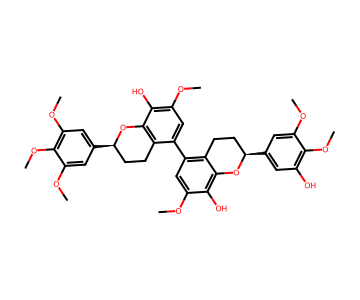}}
\end{minipage}\hfill
\begin{minipage}[t]{0.685\linewidth}
\vspace{0pt}
{\fontsize{8.0}{8.8}\selectfont\bfseries\textcolor{SAGABlue}{\#2 score=0.325 \quad T=0.07}}\\[-1pt]
{\fontsize{7.55}{8.35}\selectfont\raggedright\sloppy The molecule is a biflavonoid obtained by coupling of 8,\allowbreak{}3'-\allowbreak{}dihydroxy-\allowbreak{}7,\allowbreak{}4',\allowbreak{}5'-\allowbreak{}trimethoxyflavan and 8-\allowbreak{}hydroxy-\allowbreak{}7,\allowbreak{}3',\allowbreak{}4',\allowbreak{}5'-\allowbreak{}tetramethoxyflavan resulting in a bond between C-\allowbreak{}5 positions of the two chromene rings. Isolated from Muntingia calabura,\allowbreak{} it exhibits antineoplastic activity. It has a role as a metabolite and an antineoplastic agent. It is a biflavonoid,\allowbreak{} a hydroxyflavan,\allowbreak{} a polyphenol,\allowbreak{} a methoxyflavan and a ring assembly.\par}
\end{minipage}
\end{minipage}\hfill\begin{minipage}[t]{0.485\textwidth}
\vspace{0pt}
\noindent\begin{minipage}[t]{0.285\linewidth}
\vspace{0pt}
\fcolorbox{SAGAGreen}{white}{\includegraphics[width=\dimexpr\linewidth-2\fboxsep-2\fboxrule\relax]{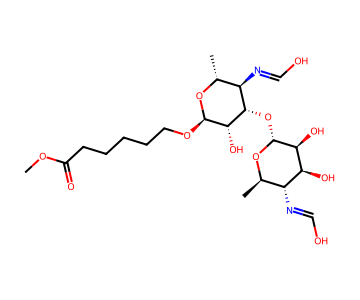}}
\end{minipage}\hfill
\begin{minipage}[t]{0.685\linewidth}
\vspace{0pt}
{\fontsize{8.0}{8.8}\selectfont\bfseries\textcolor{SAGAGreen}{\#2 score=0.318 \quad T=0.94}}\\[-1pt]
{\fontsize{7.55}{8.35}\selectfont\raggedright\sloppy The molecule is a \textcolor{SAGARed}{glycoside} that consists of two \textcolor{SAGARed}{N-\allowbreak{}formyl-\allowbreak{}alpha-\allowbreak{}D-\allowbreak{}perosamine} residues linked \textcolor{SAGARed}{(1-\allowbreak{}\textgreater{}3)\allowbreak{}} and linked at the reducing end glycosidically to a \textcolor{SAGARed}{5-\allowbreak{}(methoxycarbonyl)\allowbreak{}pentyl} group. It is a \textcolor{SAGARed}{glycoside,\allowbreak{}} a \textcolor{SAGARed}{methyl} \textcolor{SAGARed}{ester} and a disaccharide derivative. It derives from an alpha-\allowbreak{}D-\allowbreak{}Rhap4NFo-\allowbreak{}(1-\allowbreak{}\textgreater{}3)\allowbreak{}-\allowbreak{}alpha-\allowbreak{}D-\allowbreak{}Rhap4NFo.\par}
\end{minipage}
\end{minipage}\par\vspace{1.2mm}
\noindent\begin{minipage}[t]{0.485\textwidth}
\vspace{0pt}
\noindent\begin{minipage}[t]{0.285\linewidth}
\vspace{0pt}
\fcolorbox{SAGAGray}{white}{\includegraphics[width=\dimexpr\linewidth-2\fboxsep-2\fboxrule\relax]{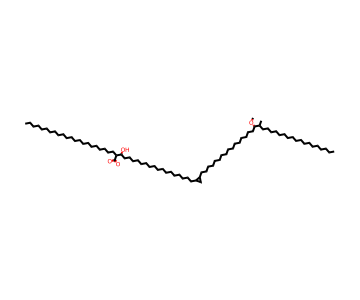}}
\end{minipage}\hfill
\begin{minipage}[t]{0.685\linewidth}
\vspace{0pt}
{\fontsize{8.0}{8.8}\selectfont\bfseries\textcolor{SAGABlue}{\#3 score=0.312 \quad T=0.13}}\\[-1pt]
{\fontsize{7.55}{8.35}\selectfont\raggedright\sloppy The molecule is an ultra-\allowbreak{}long-\allowbreak{}chain fatty acid anion that is the conjugate base of methoxy mycolic acid,\allowbreak{} obtained by deprotonation of the carboxy group;\allowbreak{} major species at pH 7.3. It is a hydroxy fatty acid anion,\allowbreak{} a mycolate,\allowbreak{} a branched-\allowbreak{}chain fatty acid anion and an ultra-\allowbreak{}long-\allowbreak{}chain fatty acid anion. It is a conjugate acid of a methoxy mycolic acid.\par}
\end{minipage}
\end{minipage}\hfill\begin{minipage}[t]{0.485\textwidth}
\vspace{0pt}
\noindent\begin{minipage}[t]{0.285\linewidth}
\vspace{0pt}
\fcolorbox{SAGAGreen}{white}{\includegraphics[width=\dimexpr\linewidth-2\fboxsep-2\fboxrule\relax]{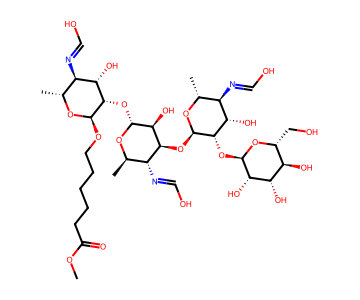}}
\end{minipage}\hfill
\begin{minipage}[t]{0.685\linewidth}
\vspace{0pt}
{\fontsize{8.0}{8.8}\selectfont\bfseries\textcolor{SAGAGreen}{\#3 score=0.292 \quad T=0.82}}\\[-1pt]
{\fontsize{7.55}{8.35}\selectfont\raggedright\sloppy The molecule is a \textcolor{SAGARed}{glycoside} that consists of an alpha-\allowbreak{}D-\allowbreak{}mannose residue and three \textcolor{SAGARed}{N-\allowbreak{}formyl-\allowbreak{}alpha-\allowbreak{}D-\allowbreak{}perosamine} residues linked sequentially \textcolor{SAGARed}{(1-\allowbreak{}\textgreater{}2)\allowbreak{},\allowbreak{}} \textcolor{SAGARed}{(1-\allowbreak{}\textgreater{}3)\allowbreak{}} and (1-\allowbreak{}\textgreater{}2)\allowbreak{} and linked at the reducing end \textcolor{SAGARed}{glycosidically} to a \textcolor{SAGARed}{5-\allowbreak{}(methoxycarbonyl)\allowbreak{}pentyl} group. It is a \textcolor{SAGARed}{methyl} \textcolor{SAGARed}{ester,\allowbreak{}} a \textcolor{SAGARed}{glycoside} and a tetrasaccharide derivative.\par}
\end{minipage}
\end{minipage}\par\vspace{1.2mm}
\par\vspace{1.0mm}
{\fontsize{7.6}{8.3}\selectfont\textcolor{SAGARed}{red} = structural-class term shared with the query \quad
score = text-\allowbreak{}-\allowbreak{}SMILES cosine similarity \quad
T = Morgan-fingerprint Tanimoto similarity to target}
\endgroup

\caption{\textbf{Supplementary Fig.~\ref{fig:supp_casestudy_text_overlap_atomas_b} $|$ Description-level structural-class overlap for Atomas-B retrieval examples.} Text$\rightarrow$Molecule retrieval examples for Atomas-B, shown with the same annotation scheme as Supplementary Fig.~\ref{fig:supp_casestudy_text_overlap_molca}. Scores are native Atomas text--SMILES cosine similarities, and red text marks structural-class terms shared with the query description. The \methodname\ neighbours have higher structural similarity to the target and frequently share chemotype terms with the query descriptions.}
\label{fig:supp_casestudy_text_overlap_atomas_b}
\end{figure}

\begin{figure}[p]
\ContinuedFloat
\centering

\begingroup
\setlength{\fboxsep}{2pt}
\setlength{\parindent}{0pt}
\footnotesize
{\fontsize{8.8}{9.6}\selectfont\bfseries Query 2 \quad Text$\rightarrow$Molecule \quad target CID 12306712}\par\vspace{1.2mm}
\begin{minipage}[t]{0.14\textwidth}
\vspace{0pt}
\fcolorbox{SAGADark}{white}{\includegraphics[width=\dimexpr\linewidth-2\fboxsep-2\fboxrule\relax]{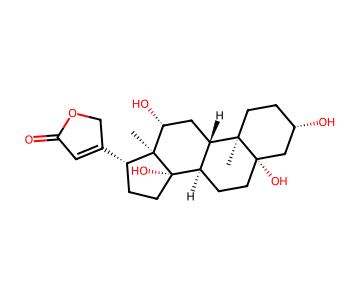}}
\end{minipage}\hfill
\begin{minipage}[t]{0.84\textwidth}
\vspace{0pt}
{\fontsize{8.0}{8.8}\selectfont\raggedright\sloppy The molecule is a 3beta-\allowbreak{}hydroxy steroid,\allowbreak{} a 5beta-\allowbreak{}hydroxy steroid,\allowbreak{} a 12beta-\allowbreak{}hydroxy steroid and a 14beta-\allowbreak{}hydroxy steroid. It derives from a hydride of a 5beta-\allowbreak{}cardanolide.\par}
\end{minipage}\par\vspace{2.4mm}
\noindent\begin{minipage}[t]{0.485\textwidth}{\fontsize{8.4}{9.2}\selectfont\textcolor{SAGABlue}{\textbf{Atomas-\allowbreak{}B (baseline): top-3 retrieved}}}\end{minipage}\hfill\begin{minipage}[t]{0.485\textwidth}{\fontsize{8.4}{9.2}\selectfont\textcolor{SAGAGreen}{\textbf{Atomas-\allowbreak{}B + \methodname: top-3 retrieved}}}\end{minipage}
\noindent\begin{minipage}[t]{0.485\textwidth}
\vspace{0pt}
\noindent\begin{minipage}[t]{0.285\linewidth}
\vspace{0pt}
\fcolorbox{SAGAGray}{white}{\includegraphics[width=\dimexpr\linewidth-2\fboxsep-2\fboxrule\relax]{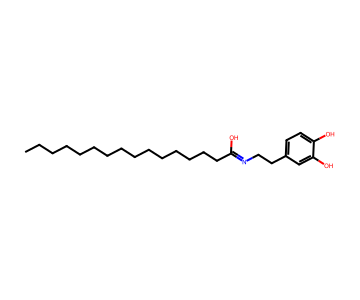}}
\end{minipage}\hfill
\begin{minipage}[t]{0.685\linewidth}
\vspace{0pt}
{\fontsize{8.0}{8.8}\selectfont\bfseries\textcolor{SAGABlue}{\#1 score=0.334 \quad T=0.05}}\\[-1pt]
{\fontsize{7.55}{8.35}\selectfont\raggedright\sloppy The molecule is a fatty amide resulting from the formal condensation of the carboxy group of hexadecanoic acid with the amino group of dopamine. It is present as an endogenous compound in the mammalian brain. It is a monocarboxylic acid amide,\allowbreak{} a fatty amide and a member of catechols. It derives from a hexadecanoic acid and a dopamine.\par}
\end{minipage}
\end{minipage}\hfill\begin{minipage}[t]{0.485\textwidth}
\vspace{0pt}
\noindent\begin{minipage}[t]{0.285\linewidth}
\vspace{0pt}
\fcolorbox{SAGAGreen}{white}{\includegraphics[width=\dimexpr\linewidth-2\fboxsep-2\fboxrule\relax]{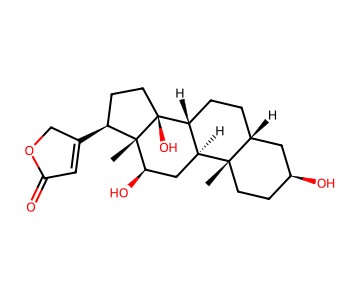}}
\end{minipage}\hfill
\begin{minipage}[t]{0.685\linewidth}
\vspace{0pt}
{\fontsize{8.0}{8.8}\selectfont\bfseries\textcolor{SAGAGreen}{\#1 score=0.321 \quad T=0.66}}\\[-1pt]
{\fontsize{7.55}{8.35}\selectfont\raggedright\sloppy The molecule is a \textcolor{SAGARed}{hydroxy} \textcolor{SAGARed}{steroid} that consists of \textcolor{SAGARed}{5beta-\allowbreak{}cardanolide} having a double bond at the 20(22)\allowbreak{}-\allowbreak{}position as well as \textcolor{SAGARed}{hydroxy} groups at the \textcolor{SAGARed}{3beta-\allowbreak{},\allowbreak{}} \textcolor{SAGARed}{12beta-\allowbreak{}} and \textcolor{SAGARed}{14beta-\allowbreak{}positions.} It has been isolated from the plant species of the genus Digitalis. It has a role as a hapten and a plant metabolite. It is a \textcolor{SAGARed}{3beta-\allowbreak{}sterol,\allowbreak{}} a \textcolor{SAGARed}{12beta-\allowbreak{}hydroxy} \textcolor{SAGARed}{steroid,\allowbreak{}} a \textcolor{SAGARed}{3beta-\allowbreak{}hydroxy} \textcolor{SAGARed}{steroid} and a \textcolor{SAGARed}{14beta-\allowbreak{}hydroxy} \textcolor{SAGARed}{steroid.} It is a conjugate acid of a digoxigenin(1-\allowbreak{})\allowbreak{}. It derives from a hydride of a \textcolor{SAGARed}{5beta-\allowbreak{}cardanolide.}\par}
\end{minipage}
\end{minipage}\par\vspace{1.2mm}
\noindent\begin{minipage}[t]{0.485\textwidth}
\vspace{0pt}
\noindent\begin{minipage}[t]{0.285\linewidth}
\vspace{0pt}
\fcolorbox{SAGAGray}{white}{\includegraphics[width=\dimexpr\linewidth-2\fboxsep-2\fboxrule\relax]{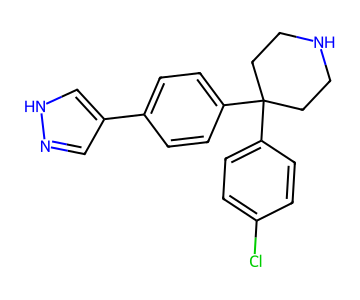}}
\end{minipage}\hfill
\begin{minipage}[t]{0.685\linewidth}
\vspace{0pt}
{\fontsize{8.0}{8.8}\selectfont\bfseries\textcolor{SAGABlue}{\#2 score=0.319 \quad T=0.07}}\\[-1pt]
{\fontsize{7.55}{8.35}\selectfont\raggedright\sloppy The molecule is a member of the class of piperidines carrying two aryl substituents (4-\allowbreak{}chlorophenyl and 4-\allowbreak{}(pyrazol-\allowbreak{}4-\allowbreak{}yl)\allowbreak{}phenyl)\allowbreak{} at position 4. It has a role as an EC 2.7.11.1 (non-\allowbreak{}specific serine/\allowbreak{}threonine protein kinase)\allowbreak{} inhibitor and an antineoplastic agent. It is a member of pyrazoles,\allowbreak{} a member of piperidines and a member of monochlorobenzenes.\par}
\end{minipage}
\end{minipage}\hfill\begin{minipage}[t]{0.485\textwidth}
\vspace{0pt}
\noindent\begin{minipage}[t]{0.285\linewidth}
\vspace{0pt}
\fcolorbox{SAGAGreen}{white}{\includegraphics[width=\dimexpr\linewidth-2\fboxsep-2\fboxrule\relax]{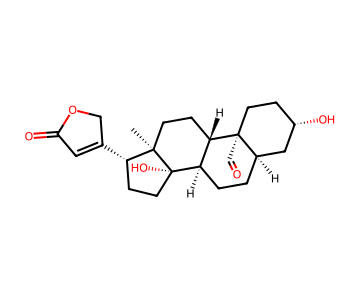}}
\end{minipage}\hfill
\begin{minipage}[t]{0.685\linewidth}
\vspace{0pt}
{\fontsize{8.0}{8.8}\selectfont\bfseries\textcolor{SAGAGreen}{\#2 score=0.309 \quad T=0.54}}\\[-1pt]
{\fontsize{7.55}{8.35}\selectfont\raggedright\sloppy The molecule is a \textcolor{SAGARed}{3beta-\allowbreak{}hydroxy} \textcolor{SAGARed}{steroid,\allowbreak{}} a \textcolor{SAGARed}{14beta-\allowbreak{}hydroxy} \textcolor{SAGARed}{steroid,\allowbreak{}} a 19-\allowbreak{}oxo \textcolor{SAGARed}{steroid} and a \textcolor{SAGARed}{steroid} aldehyde. It derives from a hydride of a \textcolor{SAGARed}{5beta-\allowbreak{}cardanolide.}\par}
\end{minipage}
\end{minipage}\par\vspace{1.2mm}
\noindent\begin{minipage}[t]{0.485\textwidth}
\vspace{0pt}
\noindent\begin{minipage}[t]{0.285\linewidth}
\vspace{0pt}
\fcolorbox{SAGAGray}{white}{\includegraphics[width=\dimexpr\linewidth-2\fboxsep-2\fboxrule\relax]{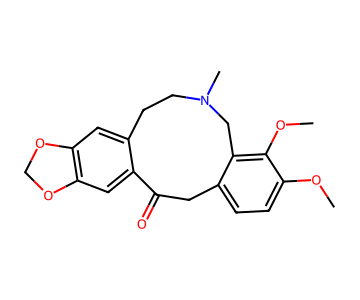}}
\end{minipage}\hfill
\begin{minipage}[t]{0.685\linewidth}
\vspace{0pt}
{\fontsize{8.0}{8.8}\selectfont\bfseries\textcolor{SAGABlue}{\#3 score=0.313 \quad T=0.09}}\\[-1pt]
{\fontsize{7.55}{8.35}\selectfont\raggedright\sloppy The molecule is a dibenzazecine alkaloid,\allowbreak{} an organic heterotetracyclic compound,\allowbreak{} a tertiary amino compound,\allowbreak{} a cyclic ketone,\allowbreak{} a cyclic acetal and an aromatic ether.\par}
\end{minipage}
\end{minipage}\hfill\begin{minipage}[t]{0.485\textwidth}
\vspace{0pt}
\noindent\begin{minipage}[t]{0.285\linewidth}
\vspace{0pt}
\fcolorbox{SAGAGreen}{white}{\includegraphics[width=\dimexpr\linewidth-2\fboxsep-2\fboxrule\relax]{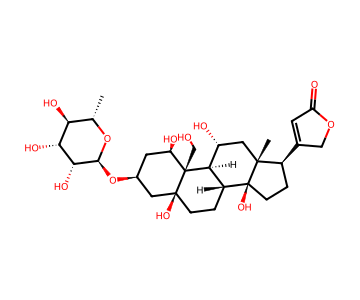}}
\end{minipage}\hfill
\begin{minipage}[t]{0.685\linewidth}
\vspace{0pt}
{\fontsize{8.0}{8.8}\selectfont\bfseries\textcolor{SAGAGreen}{\#3 score=0.279 \quad T=0.43}}\\[-1pt]
{\fontsize{7.55}{8.35}\selectfont\raggedright\sloppy The molecule is a \textcolor{SAGARed}{steroid} hormone that is a multi-\allowbreak{}hydroxylated alpha-\allowbreak{}L-\allowbreak{}rhamnosyl \textcolor{SAGARed}{cardenoloide.} It binds to and inhibits the plasma membrane Na(+)\allowbreak{}/\allowbreak{}K(+)\allowbreak{}-\allowbreak{}ATPase (sodium pump)\allowbreak{}. It has been isolated naturally from Strophanthus gratus. It has a role as an EC 3.6.3.9 (Na(+)\allowbreak{}/\allowbreak{}K(+)\allowbreak{}-\allowbreak{}transporting ATPase)\allowbreak{} inhibitor,\allowbreak{} an EC 3.6.3.10 (H(+)\allowbreak{}/\allowbreak{}K(+)\allowbreak{}-\allowbreak{}exchanging ATPase)\allowbreak{} inhibitor,\allowbreak{} an EC 2.3.3.1 [citrate (Si)\allowbreak{}-\allowbreak{}synthase]\allowbreak{} inhibitor,\allowbreak{} an EC 3.1.3.41 (4-\allowbreak{}nitrophenylphosphatase)\allowbreak{} inhibitor,\allowbreak{} a plant metabolite,\allowbreak{} a cardiotonic drug,\allowbreak{} an ion transport inhibitor and an anti-\allowbreak{}arrhythmia drug. It is a \textcolor{SAGARed}{cardenolide} glycoside,\allowbreak{} a \textcolor{SAGARed}{steroid} hormone,\allowbreak{} an alpha-\allowbreak{}L-\allowbreak{}rhamnoside,\allowbreak{} a \textcolor{SAGARed}{14beta-\allowbreak{}hydroxy} \textcolor{SAGARed}{steroid,\allowbreak{}} a \textcolor{SAGARed}{5beta-\allowbreak{}hydroxy} \textcolor{SAGARed}{steroid} and an 11alpha-\allowbreak{}hydroxy \textcolor{SAGARed}{steroid.} It is a conjugate acid of an ouabain(1-\allowbreak{})\allowbreak{}.\par}
\end{minipage}
\end{minipage}\par\vspace{1.2mm}
\par\vspace{1.0mm}
{\fontsize{7.6}{8.3}\selectfont\textcolor{SAGARed}{red} = structural-class term shared with the query \quad
score = text-\allowbreak{}-\allowbreak{}SMILES cosine similarity \quad
T = Morgan-fingerprint Tanimoto similarity to target}
\endgroup

\caption[]{\textbf{Supplementary Fig.~\ref{fig:supp_casestudy_text_overlap_atomas_b} $|$ Description-level structural-class overlap for Atomas-B retrieval examples, continued.} Same layout and annotation scheme as the first panel; red text marks structural-class terms shared with the query description.}
\end{figure}

\begin{figure}[p]
\ContinuedFloat
\centering

\begingroup
\setlength{\fboxsep}{2pt}
\setlength{\parindent}{0pt}
\footnotesize
{\fontsize{8.8}{9.6}\selectfont\bfseries Query 3 \quad Text$\rightarrow$Molecule \quad target CID 25202222}\par\vspace{1.2mm}
\begin{minipage}[t]{0.14\textwidth}
\vspace{0pt}
\fcolorbox{SAGADark}{white}{\includegraphics[width=\dimexpr\linewidth-2\fboxsep-2\fboxrule\relax]{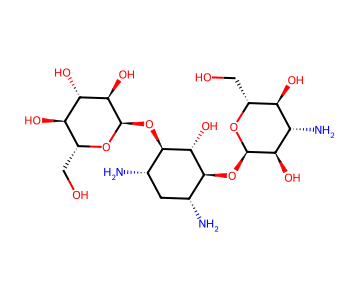}}
\end{minipage}\hfill
\begin{minipage}[t]{0.84\textwidth}
\vspace{0pt}
{\fontsize{8.0}{8.8}\selectfont\raggedright\sloppy The molecule is a kanamycin that is kanamycin A in which the 6'-\allowbreak{}amino group is replaced by a hydroxy group. It derives from a kanamycin A. It is a conjugate acid of a kanamycin X(3+)\allowbreak{}.\par}
\end{minipage}\par\vspace{2.4mm}
\noindent\begin{minipage}[t]{0.485\textwidth}{\fontsize{8.4}{9.2}\selectfont\textcolor{SAGABlue}{\textbf{Atomas-\allowbreak{}B (baseline): top-3 retrieved}}}\end{minipage}\hfill\begin{minipage}[t]{0.485\textwidth}{\fontsize{8.4}{9.2}\selectfont\textcolor{SAGAGreen}{\textbf{Atomas-\allowbreak{}B + \methodname: top-3 retrieved}}}\end{minipage}
\noindent\begin{minipage}[t]{0.485\textwidth}
\vspace{0pt}
\noindent\begin{minipage}[t]{0.285\linewidth}
\vspace{0pt}
\fcolorbox{SAGAGray}{white}{\includegraphics[width=\dimexpr\linewidth-2\fboxsep-2\fboxrule\relax]{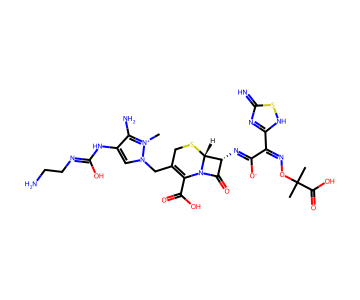}}
\end{minipage}\hfill
\begin{minipage}[t]{0.685\linewidth}
\vspace{0pt}
{\fontsize{8.0}{8.8}\selectfont\bfseries\textcolor{SAGABlue}{\#1 score=0.337 \quad T=0.06}}\\[-1pt]
{\fontsize{7.55}{8.35}\selectfont\raggedright\sloppy The molecule is a fifth-\allowbreak{}generation cephalosporin antibiotic having (5-\allowbreak{}amino-\allowbreak{}4-\allowbreak{}\{[(2-\allowbreak{}aminoethyl)\allowbreak{}carbamoyl]\allowbreak{}amino\}-\allowbreak{}1-\allowbreak{}methyl-\allowbreak{}1H-\allowbreak{}pyrazol-\allowbreak{}2-\allowbreak{}ium-\allowbreak{}2-\allowbreak{}yl)\allowbreak{}methyl and [(2Z)\allowbreak{}-\allowbreak{}2-\allowbreak{}(5-\allowbreak{}amino-\allowbreak{}1,\allowbreak{}2,\allowbreak{}4-\allowbreak{}thiadiazol-\allowbreak{}3-\allowbreak{}yl)\allowbreak{}-\allowbreak{}2-\allowbreak{}\{[(2-\allowbreak{}carboxypropan-\allowbreak{}2-\allowbreak{}yl)\allowbreak{}oxy]\allowbreak{}imino\}acetyl]\allowbreak{}amino side groups located at positions 3 and 7 respectively;\allowbreak{} developed for the treatment of infections with gram-\allowbreak{}negative bacteria that have become resistant to conventional antibiotics. It is a cephalosporin and a member of thiadiazoles.\par}
\end{minipage}
\end{minipage}\hfill\begin{minipage}[t]{0.485\textwidth}
\vspace{0pt}
\noindent\begin{minipage}[t]{0.285\linewidth}
\vspace{0pt}
\fcolorbox{SAGAGreen}{white}{\includegraphics[width=\dimexpr\linewidth-2\fboxsep-2\fboxrule\relax]{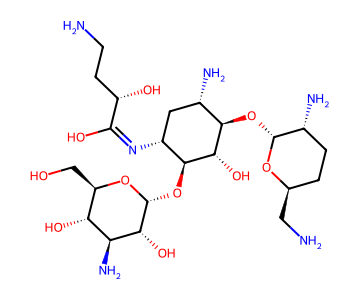}}
\end{minipage}\hfill
\begin{minipage}[t]{0.685\linewidth}
\vspace{0pt}
{\fontsize{8.0}{8.8}\selectfont\bfseries\textcolor{SAGAGreen}{\#1 score=0.297 \quad T=0.48}}\\[-1pt]
{\fontsize{7.55}{8.35}\selectfont\raggedright\sloppy The molecule is a \textcolor{SAGARed}{kanamycin} that is \textcolor{SAGARed}{kanamycin} B bearing an N-\allowbreak{}(2S)\allowbreak{}-\allowbreak{}4-\allowbreak{}amino-\allowbreak{}2-\allowbreak{}hydroxybutyryl group on the aminocyclitol ring. It has a role as an antimicrobial agent,\allowbreak{} a protein synthesis inhibitor,\allowbreak{} an antibacterial agent and an antibacterial drug. It is a member of \textcolor{SAGARed}{kanamycins} and an aminoglycoside. It derives from a \textcolor{SAGARed}{kanamycin} B.\par}
\end{minipage}
\end{minipage}\par\vspace{1.2mm}
\noindent\begin{minipage}[t]{0.485\textwidth}
\vspace{0pt}
\noindent\begin{minipage}[t]{0.285\linewidth}
\vspace{0pt}
\fcolorbox{SAGAGray}{white}{\includegraphics[width=\dimexpr\linewidth-2\fboxsep-2\fboxrule\relax]{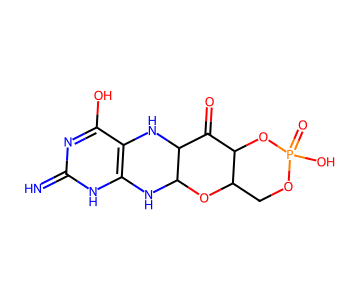}}
\end{minipage}\hfill
\begin{minipage}[t]{0.685\linewidth}
\vspace{0pt}
{\fontsize{8.0}{8.8}\selectfont\bfseries\textcolor{SAGABlue}{\#2 score=0.330 \quad T=0.08}}\\[-1pt]
{\fontsize{7.55}{8.35}\selectfont\raggedright\sloppy The molecule is a linear-\allowbreak{}fused organic heterotetracyclic compound consisting of a [1,\allowbreak{}3,\allowbreak{}2]\allowbreak{}dioxaphosphinane fused to a pyran-\allowbreak{}4-\allowbreak{}one ring which is in turn fused to a pteridine ring system. It is an organic heterotetracyclic compound,\allowbreak{} an oxacycle and an organonitrogen heterocyclic compound. It is a conjugate acid of a precursor Z(1-\allowbreak{})\allowbreak{}.\par}
\end{minipage}
\end{minipage}\hfill\begin{minipage}[t]{0.485\textwidth}
\vspace{0pt}
\noindent\begin{minipage}[t]{0.285\linewidth}
\vspace{0pt}
\fcolorbox{SAGAGreen}{white}{\includegraphics[width=\dimexpr\linewidth-2\fboxsep-2\fboxrule\relax]{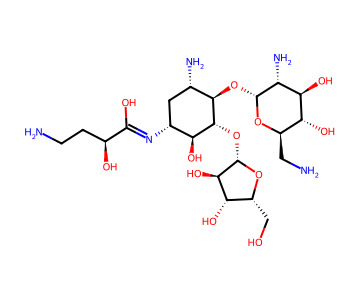}}
\end{minipage}\hfill
\begin{minipage}[t]{0.685\linewidth}
\vspace{0pt}
{\fontsize{8.0}{8.8}\selectfont\bfseries\textcolor{SAGAGreen}{\#2 score=0.279 \quad T=0.45}}\\[-1pt]
{\fontsize{7.55}{8.35}\selectfont\raggedright\sloppy The molecule is a butirosin that consists of neamine in which is substituted at position 2 by a beta-\allowbreak{}D-\allowbreak{}xylofuranosyl and at position 4 by an (S)\allowbreak{}-\allowbreak{}2-\allowbreak{}hydroxy-\allowbreak{}4-\allowbreak{}aminobutyryl group. It has a role as an antimicrobial agent. It derives from a neamine.\par}
\end{minipage}
\end{minipage}\par\vspace{1.2mm}
\noindent\begin{minipage}[t]{0.485\textwidth}
\vspace{0pt}
\noindent\begin{minipage}[t]{0.285\linewidth}
\vspace{0pt}
\fcolorbox{SAGAGray}{white}{\includegraphics[width=\dimexpr\linewidth-2\fboxsep-2\fboxrule\relax]{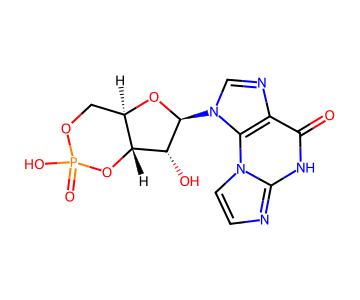}}
\end{minipage}\hfill
\begin{minipage}[t]{0.685\linewidth}
\vspace{0pt}
{\fontsize{8.0}{8.8}\selectfont\bfseries\textcolor{SAGABlue}{\#3 score=0.314 \quad T=0.08}}\\[-1pt]
{\fontsize{7.55}{8.35}\selectfont\raggedright\sloppy The molecule is a nucleoside 3',\allowbreak{}5'-\allowbreak{}cyclic monophosphate analogue having imidazo[2,\allowbreak{}1-\allowbreak{}b]\allowbreak{}purin-\allowbreak{}4-\allowbreak{}one as the modified nucleobase. It is a nucleoside monophosphate analogue and an imidazopurine. It derives from a 3',\allowbreak{}5'-\allowbreak{}cyclic GMP.\par}
\end{minipage}
\end{minipage}\hfill\begin{minipage}[t]{0.485\textwidth}
\vspace{0pt}
\noindent\begin{minipage}[t]{0.285\linewidth}
\vspace{0pt}
\fcolorbox{SAGAGreen}{white}{\includegraphics[width=\dimexpr\linewidth-2\fboxsep-2\fboxrule\relax]{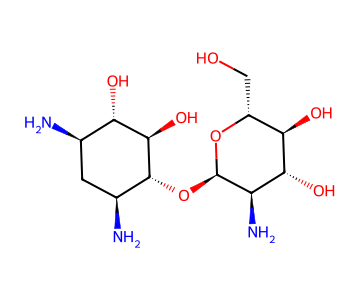}}
\end{minipage}\hfill
\begin{minipage}[t]{0.685\linewidth}
\vspace{0pt}
{\fontsize{8.0}{8.8}\selectfont\bfseries\textcolor{SAGAGreen}{\#3 score=0.275 \quad T=0.68}}\\[-1pt]
{\fontsize{7.55}{8.35}\selectfont\raggedright\sloppy The molecule is an \textcolor{SAGARed}{aminoglycoside} that is 4alpha,\allowbreak{}6alpha-\allowbreak{}diaminocyclohexane-\allowbreak{}1beta,\allowbreak{}2alpha,\allowbreak{}3beta-\allowbreak{}triol in which the pro-\allowbreak{}R hydroxy group has been converted into its 2-\allowbreak{}amino-\allowbreak{}alpha-\allowbreak{}D-\allowbreak{}glucoside derivative. It is an \textcolor{SAGARed}{aminoglycoside,\allowbreak{}} a primary amino compound and a triamine. It derives from an alpha-\allowbreak{}D-\allowbreak{}glucose and a \textcolor{SAGARed}{2-\allowbreak{}deoxystreptamine.} It is a conjugate base of a paromamine(3+)\allowbreak{}.\par}
\end{minipage}
\end{minipage}\par\vspace{1.2mm}
\par\vspace{1.0mm}
{\fontsize{7.6}{8.3}\selectfont\textcolor{SAGARed}{red} = structural-class term shared with the query \quad
score = text-\allowbreak{}-\allowbreak{}SMILES cosine similarity \quad
T = Morgan-fingerprint Tanimoto similarity to target}
\endgroup

\caption[]{\textbf{Supplementary Fig.~\ref{fig:supp_casestudy_text_overlap_atomas_b} $|$ Description-level structural-class overlap for Atomas-B retrieval examples, continued.} Same layout and annotation scheme as the first panel; red text marks structural-class terms shared with the query description.}
\end{figure}

\begin{figure}[p]
\centering

\begingroup
\setlength{\fboxsep}{2pt}
\setlength{\parindent}{0pt}
\footnotesize
{\fontsize{8.8}{9.6}\selectfont\bfseries Query 1 \quad Text$\rightarrow$Molecule \quad target CID 25201044}\par\vspace{1.2mm}
\begin{minipage}[t]{0.14\textwidth}
\vspace{0pt}
\fcolorbox{SAGADark}{white}{\includegraphics[width=\dimexpr\linewidth-2\fboxsep-2\fboxrule\relax]{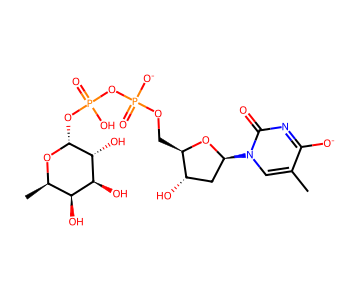}}
\end{minipage}\hfill
\begin{minipage}[t]{0.84\textwidth}
\vspace{0pt}
{\fontsize{8.0}{8.8}\selectfont\raggedright\sloppy The molecule is a dTDP-\allowbreak{}D-\allowbreak{}fucose(2-\allowbreak{})\allowbreak{} in which the anomeric centre of the fucose fragment has alpha-\allowbreak{}configuration. It is a conjugate base of a dTDP-\allowbreak{}alpha-\allowbreak{}D-\allowbreak{}fucose.\par}
\end{minipage}\par\vspace{2.4mm}
\noindent\begin{minipage}[t]{0.485\textwidth}{\fontsize{8.4}{9.2}\selectfont\textcolor{SAGABlue}{\textbf{Atomas-\allowbreak{}L (baseline): top-3 retrieved}}}\end{minipage}\hfill\begin{minipage}[t]{0.485\textwidth}{\fontsize{8.4}{9.2}\selectfont\textcolor{SAGAGreen}{\textbf{Atomas-\allowbreak{}L + \methodname: top-3 retrieved}}}\end{minipage}
\noindent\begin{minipage}[t]{0.485\textwidth}
\vspace{0pt}
\noindent\begin{minipage}[t]{0.285\linewidth}
\vspace{0pt}
\fcolorbox{SAGAGray}{white}{\includegraphics[width=\dimexpr\linewidth-2\fboxsep-2\fboxrule\relax]{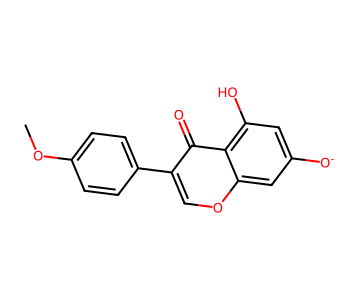}}
\end{minipage}\hfill
\begin{minipage}[t]{0.685\linewidth}
\vspace{0pt}
{\fontsize{8.0}{8.8}\selectfont\bfseries\textcolor{SAGABlue}{\#1 score=0.340 \quad T=0.12}}\\[-1pt]
{\fontsize{7.55}{8.35}\selectfont\raggedright\sloppy The molecule is conjugate base of biochanin A arising from selective deprotonation of the 7-\allowbreak{}hydroxy group;\allowbreak{} major species at pH 7.3. It is a conjugate base of a biochanin A.\par}
\end{minipage}
\end{minipage}\hfill\begin{minipage}[t]{0.485\textwidth}
\vspace{0pt}
\noindent\begin{minipage}[t]{0.285\linewidth}
\vspace{0pt}
\fcolorbox{SAGAGreen}{white}{\includegraphics[width=\dimexpr\linewidth-2\fboxsep-2\fboxrule\relax]{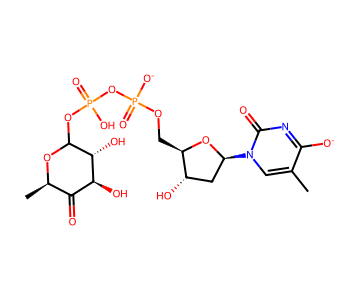}}
\end{minipage}\hfill
\begin{minipage}[t]{0.685\linewidth}
\vspace{0pt}
{\fontsize{8.0}{8.8}\selectfont\bfseries\textcolor{SAGAGreen}{\#1 score=0.270 \quad T=0.81}}\\[-1pt]
{\fontsize{7.55}{8.35}\selectfont\raggedright\sloppy The molecule is dianion of \textcolor{SAGARed}{dTDP-\allowbreak{}4-\allowbreak{}dehydro-\allowbreak{}6-\allowbreak{}deoxy-\allowbreak{}D-\allowbreak{}galactose} arising from deprotonation of the \textcolor{SAGARed}{diphosphate} OH groups;\allowbreak{} major species at pH 7.3. It is a conjugate base of a \textcolor{SAGARed}{dTDP-\allowbreak{}4-\allowbreak{}dehydro-\allowbreak{}6-\allowbreak{}deoxy-\allowbreak{}D-\allowbreak{}galactose.}\par}
\end{minipage}
\end{minipage}\par\vspace{1.2mm}
\noindent\begin{minipage}[t]{0.485\textwidth}
\vspace{0pt}
\noindent\begin{minipage}[t]{0.285\linewidth}
\vspace{0pt}
\fcolorbox{SAGAGray}{white}{\includegraphics[width=\dimexpr\linewidth-2\fboxsep-2\fboxrule\relax]{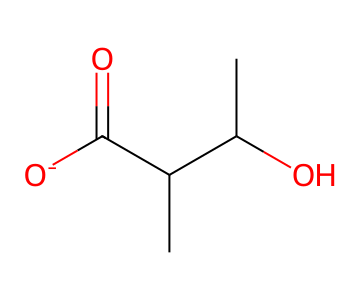}}
\end{minipage}\hfill
\begin{minipage}[t]{0.685\linewidth}
\vspace{0pt}
{\fontsize{8.0}{8.8}\selectfont\bfseries\textcolor{SAGABlue}{\#2 score=0.338 \quad T=0.05}}\\[-1pt]
{\fontsize{7.55}{8.35}\selectfont\raggedright\sloppy The molecule is a hydroxy fatty acid anion that is the conjugate base of 2-\allowbreak{}methyl-\allowbreak{}3-\allowbreak{}hydroxybutyric acid,\allowbreak{} obtained by deprotonation of the carboxy group;\allowbreak{} major species at pH 7.3. It derives from a butyrate. It is a conjugate base of a 3-\allowbreak{}hydroxy-\allowbreak{}2-\allowbreak{}methylbutanoic acid.\par}
\end{minipage}
\end{minipage}\hfill\begin{minipage}[t]{0.485\textwidth}
\vspace{0pt}
\noindent\begin{minipage}[t]{0.285\linewidth}
\vspace{0pt}
\fcolorbox{SAGAGreen}{white}{\includegraphics[width=\dimexpr\linewidth-2\fboxsep-2\fboxrule\relax]{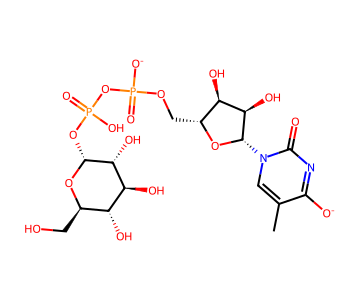}}
\end{minipage}\hfill
\begin{minipage}[t]{0.685\linewidth}
\vspace{0pt}
{\fontsize{8.0}{8.8}\selectfont\bfseries\textcolor{SAGAGreen}{\#2 score=0.246 \quad T=0.65}}\\[-1pt]
{\fontsize{7.55}{8.35}\selectfont\raggedright\sloppy The molecule is a \textcolor{SAGARed}{nucleotide-\allowbreak{}sugar} \textcolor{SAGARed}{oxoanion} arising from deprotonation of the \textcolor{SAGARed}{diphosphate} OH groups of \textcolor{SAGARed}{TDP-\allowbreak{}alpha-\allowbreak{}D-\allowbreak{}glucose.} Major microspecies at pH 7.3. It is a \textcolor{SAGARed}{nucleotide-\allowbreak{}sugar} \textcolor{SAGARed}{oxoanion} and a ribonucleoside \textcolor{SAGARed}{5'-\allowbreak{}diphosphate-\allowbreak{}alpha-\allowbreak{}D-\allowbreak{}glucose(2-\allowbreak{})\allowbreak{}.} It is a conjugate base of a \textcolor{SAGARed}{TDP-\allowbreak{}alpha-\allowbreak{}D-\allowbreak{}glucose.}\par}
\end{minipage}
\end{minipage}\par\vspace{1.2mm}
\noindent\begin{minipage}[t]{0.485\textwidth}
\vspace{0pt}
\noindent\begin{minipage}[t]{0.285\linewidth}
\vspace{0pt}
\fcolorbox{SAGAGray}{white}{\includegraphics[width=\dimexpr\linewidth-2\fboxsep-2\fboxrule\relax]{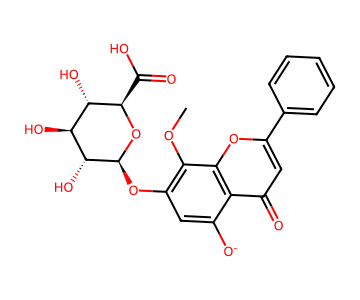}}
\end{minipage}\hfill
\begin{minipage}[t]{0.685\linewidth}
\vspace{0pt}
{\fontsize{8.0}{8.8}\selectfont\bfseries\textcolor{SAGABlue}{\#3 score=0.336 \quad T=0.16}}\\[-1pt]
{\fontsize{7.55}{8.35}\selectfont\raggedright\sloppy The molecule is the monocarboxylic acid anion that is the conjugate base of wogonin 7-\allowbreak{}O-\allowbreak{}beta-\allowbreak{}D-\allowbreak{}glucuronide. It is a \textcolor{SAGARed}{carbohydrate} acid derivative anion and a monocarboxylic acid anion. It is a conjugate base of a wogonin 7-\allowbreak{}O-\allowbreak{}beta-\allowbreak{}D-\allowbreak{}glucuronide.\par}
\end{minipage}
\end{minipage}\hfill\begin{minipage}[t]{0.485\textwidth}
\vspace{0pt}
\noindent\begin{minipage}[t]{0.285\linewidth}
\vspace{0pt}
\fcolorbox{SAGAGreen}{white}{\includegraphics[width=\dimexpr\linewidth-2\fboxsep-2\fboxrule\relax]{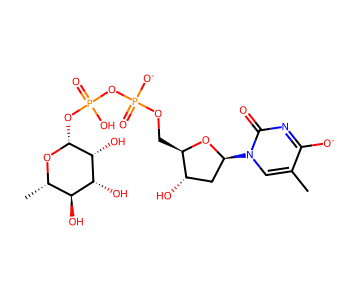}}
\end{minipage}\hfill
\begin{minipage}[t]{0.685\linewidth}
\vspace{0pt}
{\fontsize{8.0}{8.8}\selectfont\bfseries\textcolor{SAGAGreen}{\#3 score=0.241 \quad T=1.00}}\\[-1pt]
{\fontsize{7.55}{8.35}\selectfont\raggedright\sloppy The molecule is dianion of \textcolor{SAGARed}{dTDP-\allowbreak{}6-\allowbreak{}deoxy-\allowbreak{}beta-\allowbreak{}L-\allowbreak{}mannose} arising from deprotonation of both free OH groups of the \textcolor{SAGARed}{diphosphate.} It is a conjugate base of a \textcolor{SAGARed}{dTDP-\allowbreak{}beta-\allowbreak{}L-\allowbreak{}rhamnose.}\par}
\end{minipage}
\end{minipage}\par\vspace{1.2mm}
\par\vspace{1.0mm}
{\fontsize{7.6}{8.3}\selectfont\textcolor{SAGARed}{red} = structural-class term shared with the query \quad
score = text-\allowbreak{}-\allowbreak{}SMILES cosine similarity \quad
T = Morgan-fingerprint Tanimoto similarity to target}
\endgroup

\caption{\textbf{Supplementary Fig.~\ref{fig:supp_casestudy_text_overlap_atomas_l} $|$ Description-level structural-class overlap for Atomas-L retrieval examples.} Text$\rightarrow$Molecule retrieval examples for Atomas-L, shown with the same annotation scheme as Supplementary Fig.~\ref{fig:supp_casestudy_text_overlap_molca}. Scores are native Atomas text--SMILES cosine similarities, and red text marks structural-class terms shared with the query description. These examples provide a text-side view of the same structural-neighbour effect measured by Tanimoto similarity.}
\label{fig:supp_casestudy_text_overlap_atomas_l}
\end{figure}

\begin{figure}[p]
\ContinuedFloat
\centering

\begingroup
\setlength{\fboxsep}{2pt}
\setlength{\parindent}{0pt}
\footnotesize
{\fontsize{8.8}{9.6}\selectfont\bfseries Query 2 \quad Text$\rightarrow$Molecule \quad target CID 13786989}\par\vspace{1.2mm}
\begin{minipage}[t]{0.14\textwidth}
\vspace{0pt}
\fcolorbox{SAGADark}{white}{\includegraphics[width=\dimexpr\linewidth-2\fboxsep-2\fboxrule\relax]{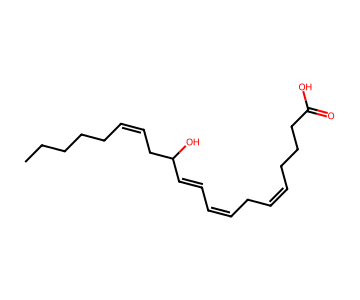}}
\end{minipage}\hfill
\begin{minipage}[t]{0.84\textwidth}
\vspace{0pt}
{\fontsize{8.0}{8.8}\selectfont\raggedright\sloppy The molecule is the (5Z,\allowbreak{}8Z,\allowbreak{}10E,\allowbreak{}14Z)\allowbreak{}-\allowbreak{}stereoisomer of 12-\allowbreak{}HETE. It has a role as a mouse metabolite. It is a conjugate acid of a 12-\allowbreak{}HETE(1-\allowbreak{})\allowbreak{}.\par}
\end{minipage}\par\vspace{2.4mm}
\noindent\begin{minipage}[t]{0.485\textwidth}{\fontsize{8.4}{9.2}\selectfont\textcolor{SAGABlue}{\textbf{Atomas-\allowbreak{}L (baseline): top-3 retrieved}}}\end{minipage}\hfill\begin{minipage}[t]{0.485\textwidth}{\fontsize{8.4}{9.2}\selectfont\textcolor{SAGAGreen}{\textbf{Atomas-\allowbreak{}L + \methodname: top-3 retrieved}}}\end{minipage}
\noindent\begin{minipage}[t]{0.485\textwidth}
\vspace{0pt}
\noindent\begin{minipage}[t]{0.285\linewidth}
\vspace{0pt}
\fcolorbox{SAGAGray}{white}{\includegraphics[width=\dimexpr\linewidth-2\fboxsep-2\fboxrule\relax]{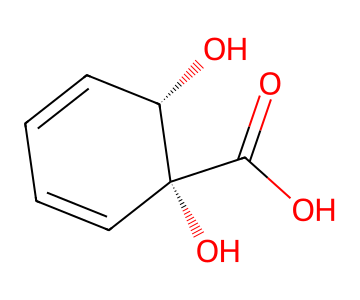}}
\end{minipage}\hfill
\begin{minipage}[t]{0.685\linewidth}
\vspace{0pt}
{\fontsize{8.0}{8.8}\selectfont\bfseries\textcolor{SAGABlue}{\#1 score=0.409 \quad T=0.09}}\\[-1pt]
{\fontsize{7.55}{8.35}\selectfont\raggedright\sloppy The molecule is a cis-\allowbreak{}1,\allowbreak{}6-\allowbreak{}dihydroxycyclohexa-\allowbreak{}2,\allowbreak{}4-\allowbreak{}dienecarboxylic acid. It is a conjugate acid of a (1R,\allowbreak{}6S)\allowbreak{}-\allowbreak{}1,\allowbreak{}6-\allowbreak{}dihydroxycyclohexa-\allowbreak{}2,\allowbreak{}4-\allowbreak{}diene-\allowbreak{}1-\allowbreak{}carboxylate. It is an enantiomer of a (1S,\allowbreak{}6R)\allowbreak{}-\allowbreak{}1,\allowbreak{}6-\allowbreak{}dihydroxycyclohexa-\allowbreak{}2,\allowbreak{}4-\allowbreak{}dienecarboxylic acid.\par}
\end{minipage}
\end{minipage}\hfill\begin{minipage}[t]{0.485\textwidth}
\vspace{0pt}
\noindent\begin{minipage}[t]{0.285\linewidth}
\vspace{0pt}
\fcolorbox{SAGAGreen}{white}{\includegraphics[width=\dimexpr\linewidth-2\fboxsep-2\fboxrule\relax]{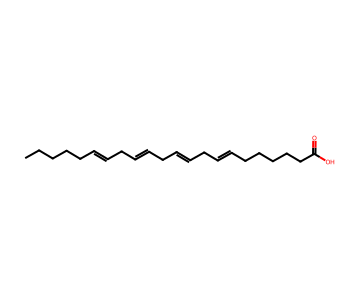}}
\end{minipage}\hfill
\begin{minipage}[t]{0.685\linewidth}
\vspace{0pt}
{\fontsize{8.0}{8.8}\selectfont\bfseries\textcolor{SAGAGreen}{\#1 score=0.279 \quad T=0.62}}\\[-1pt]
{\fontsize{7.55}{8.35}\selectfont\raggedright\sloppy The molecule is a \textcolor{SAGARed}{docosatetraenoic} acid in which the \textcolor{SAGARed}{four} \textcolor{SAGARed}{double} \textcolor{SAGARed}{bonds} are located at positions 7,\allowbreak{} 10,\allowbreak{} 13 and 16 (geometry unspecified)\allowbreak{}. It has a role as a mouse metabolite.\par}
\end{minipage}
\end{minipage}\par\vspace{1.2mm}
\noindent\begin{minipage}[t]{0.485\textwidth}
\vspace{0pt}
\noindent\begin{minipage}[t]{0.285\linewidth}
\vspace{0pt}
\fcolorbox{SAGAGray}{white}{\includegraphics[width=\dimexpr\linewidth-2\fboxsep-2\fboxrule\relax]{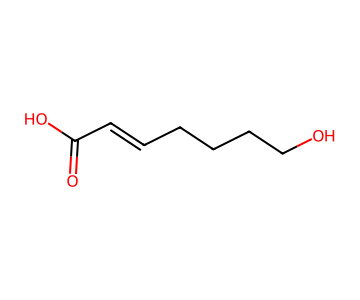}}
\end{minipage}\hfill
\begin{minipage}[t]{0.685\linewidth}
\vspace{0pt}
{\fontsize{8.0}{8.8}\selectfont\bfseries\textcolor{SAGABlue}{\#2 score=0.370 \quad T=0.24}}\\[-1pt]
{\fontsize{7.55}{8.35}\selectfont\raggedright\sloppy The molecule is an omega-\allowbreak{}hydroxy \textcolor{SAGARed}{fatty} \textcolor{SAGARed}{acid} that is trans-\allowbreak{}2-\allowbreak{}heptenoic acid in which one of the hydrogens of the terminal methyl group is replaced by a hydroxy group. It is an omega-\allowbreak{}hydroxy \textcolor{SAGARed}{fatty} \textcolor{SAGARed}{acid,\allowbreak{}} an alpha,\allowbreak{}beta-\allowbreak{}unsaturated monocarboxylic acid,\allowbreak{} a medium-\allowbreak{}chain \textcolor{SAGARed}{fatty} \textcolor{SAGARed}{acid,\allowbreak{}} a straight-\allowbreak{}chain \textcolor{SAGARed}{fatty} \textcolor{SAGARed}{acid} and a \textcolor{SAGARed}{hydroxy} monounsaturated \textcolor{SAGARed}{fatty} \textcolor{SAGARed}{acid.}\par}
\end{minipage}
\end{minipage}\hfill\begin{minipage}[t]{0.485\textwidth}
\vspace{0pt}
\noindent\begin{minipage}[t]{0.285\linewidth}
\vspace{0pt}
\fcolorbox{SAGAGreen}{white}{\includegraphics[width=\dimexpr\linewidth-2\fboxsep-2\fboxrule\relax]{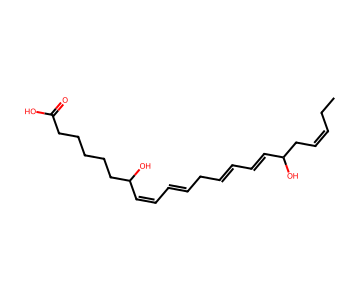}}
\end{minipage}\hfill
\begin{minipage}[t]{0.685\linewidth}
\vspace{0pt}
{\fontsize{8.0}{8.8}\selectfont\bfseries\textcolor{SAGAGreen}{\#2 score=0.248 \quad T=0.61}}\\[-1pt]
{\fontsize{7.55}{8.35}\selectfont\raggedright\sloppy The molecule is a dihydroxydocosapentaenoic acid (DiHDPE)\allowbreak{} that is (8Z,\allowbreak{}15E,\allowbreak{}19Z)\allowbreak{}-\allowbreak{}docosa-\allowbreak{}8,\allowbreak{}10,\allowbreak{}13,\allowbreak{}15,\allowbreak{}19-\allowbreak{}pentaenoic acid carrying two hydroxy substituents at positions 7 and 17. An intermediate of specialized proresolving mediators It has a role as a specialised pro-\allowbreak{}resolving mediator and a human xenobiotic metabolite. It is a conjugate acid of a 7,\allowbreak{}17-\allowbreak{}dihydroxy-\allowbreak{}(8Z,\allowbreak{}15E,\allowbreak{}19Z)\allowbreak{}-\allowbreak{}docosa-\allowbreak{}8,\allowbreak{}10,\allowbreak{}13,\allowbreak{}15,\allowbreak{}19-\allowbreak{}pentaenoate.\par}
\end{minipage}
\end{minipage}\par\vspace{1.2mm}
\noindent\begin{minipage}[t]{0.485\textwidth}
\vspace{0pt}
\noindent\begin{minipage}[t]{0.285\linewidth}
\vspace{0pt}
\fcolorbox{SAGAGray}{white}{\includegraphics[width=\dimexpr\linewidth-2\fboxsep-2\fboxrule\relax]{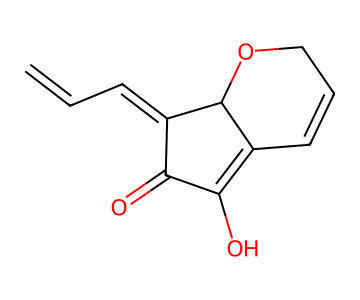}}
\end{minipage}\hfill
\begin{minipage}[t]{0.685\linewidth}
\vspace{0pt}
{\fontsize{8.0}{8.8}\selectfont\bfseries\textcolor{SAGABlue}{\#3 score=0.363 \quad T=0.06}}\\[-1pt]
{\fontsize{7.55}{8.35}\selectfont\raggedright\sloppy The molecule is an organic heterobicyclic compound that is 7,\allowbreak{}7a-\allowbreak{}dihydrocyclopenta[b]\allowbreak{}pyran-\allowbreak{}6(2H)\allowbreak{}-\allowbreak{}one substituted by a hydroxy group at position 5 and a prop-\allowbreak{}2-\allowbreak{}en-\allowbreak{}1-\allowbreak{}ylidene group at position 7 (the Z isomer)\allowbreak{}. Isolated from the sponge Ulosa and ascidian Diplosoma virens,\allowbreak{} it exhibits antimicrobial activity and toxicity against HCT116 cells (human colorectal cancer cells)\allowbreak{} by triggering apoptotic cell death. It has a role as a metabolite,\allowbreak{} an antimicrobial agent and an antineoplastic agent. It is a cyclic ether,\allowbreak{} an enol,\allowbreak{} an enone and an organic heterobicyclic compound.\par}
\end{minipage}
\end{minipage}\hfill\begin{minipage}[t]{0.485\textwidth}
\vspace{0pt}
\noindent\begin{minipage}[t]{0.285\linewidth}
\vspace{0pt}
\fcolorbox{SAGAGreen}{white}{\includegraphics[width=\dimexpr\linewidth-2\fboxsep-2\fboxrule\relax]{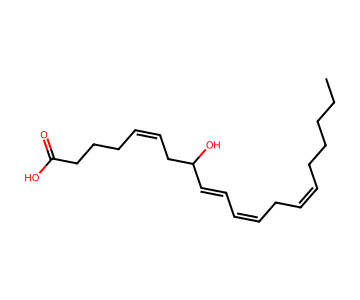}}
\end{minipage}\hfill
\begin{minipage}[t]{0.685\linewidth}
\vspace{0pt}
{\fontsize{8.0}{8.8}\selectfont\bfseries\textcolor{SAGAGreen}{\#3 score=0.228 \quad T=1.00}}\\[-1pt]
{\fontsize{7.55}{8.35}\selectfont\raggedright\sloppy The molecule is an \textcolor{SAGARed}{HETE} having a 8-\allowbreak{}hydroxy group and (5Z)\allowbreak{}-\allowbreak{},\allowbreak{} (9E)\allowbreak{}-\allowbreak{},\allowbreak{} (11Z)\allowbreak{}-\allowbreak{} and (14Z)\allowbreak{}-\allowbreak{}double bonds. It has a role as a mouse metabolite. It is a conjugate acid of an \textcolor{SAGARed}{8-\allowbreak{}HETE(1-\allowbreak{})\allowbreak{}.}\par}
\end{minipage}
\end{minipage}\par\vspace{1.2mm}
\par\vspace{1.0mm}
{\fontsize{7.6}{8.3}\selectfont\textcolor{SAGARed}{red} = structural-class term shared with the query \quad
score = text-\allowbreak{}-\allowbreak{}SMILES cosine similarity \quad
T = Morgan-fingerprint Tanimoto similarity to target}
\endgroup

\caption[]{\textbf{Supplementary Fig.~\ref{fig:supp_casestudy_text_overlap_atomas_l} $|$ Description-level structural-class overlap for Atomas-L retrieval examples, continued.} Same layout and annotation scheme as the first panel; red text marks structural-class terms shared with the query description.}
\end{figure}

\begin{figure}[p]
\ContinuedFloat
\centering

\begingroup
\setlength{\fboxsep}{2pt}
\setlength{\parindent}{0pt}
\footnotesize
{\fontsize{8.8}{9.6}\selectfont\bfseries Query 3 \quad Text$\rightarrow$Molecule \quad target CID 129011046}\par\vspace{1.2mm}
\begin{minipage}[t]{0.14\textwidth}
\vspace{0pt}
\fcolorbox{SAGADark}{white}{\includegraphics[width=\dimexpr\linewidth-2\fboxsep-2\fboxrule\relax]{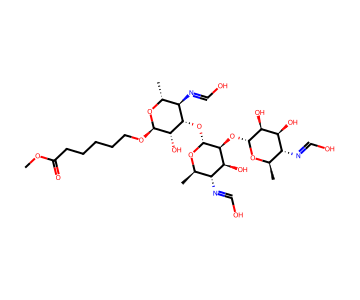}}
\end{minipage}\hfill
\begin{minipage}[t]{0.84\textwidth}
\vspace{0pt}
{\fontsize{8.0}{8.8}\selectfont\raggedright\sloppy The molecule is a glycoside that consists of three N-\allowbreak{}formyl-\allowbreak{}alpha-\allowbreak{}D-\allowbreak{}perosamine residues linked sequentially (1-\allowbreak{}\textgreater{}2)\allowbreak{} and (1-\allowbreak{}\textgreater{}3)\allowbreak{} and linked at the reducing end glycosidically to a 5-\allowbreak{}(methoxycarbonyl)\allowbreak{}pentyl group. It is a glycoside,\allowbreak{} a methyl ester and a trisaccharide derivative. It derives from an alpha-\allowbreak{}D-\allowbreak{}Rhap4NFo-\allowbreak{}(1-\allowbreak{}\textgreater{}2)\allowbreak{}-\allowbreak{}alpha-\allowbreak{}D-\allowbreak{}Rhap4NFo-\allowbreak{}(1-\allowbreak{}\textgreater{}3)\allowbreak{}-\allowbreak{}alpha-\allowbreak{}D-\allowbreak{}Rhap4NFo.\par}
\end{minipage}\par\vspace{2.4mm}
\noindent\begin{minipage}[t]{0.485\textwidth}{\fontsize{8.4}{9.2}\selectfont\textcolor{SAGABlue}{\textbf{Atomas-\allowbreak{}L (baseline): top-3 retrieved}}}\end{minipage}\hfill\begin{minipage}[t]{0.485\textwidth}{\fontsize{8.4}{9.2}\selectfont\textcolor{SAGAGreen}{\textbf{Atomas-\allowbreak{}L + \methodname: top-3 retrieved}}}\end{minipage}
\noindent\begin{minipage}[t]{0.485\textwidth}
\vspace{0pt}
\noindent\begin{minipage}[t]{0.285\linewidth}
\vspace{0pt}
\fcolorbox{SAGAGray}{white}{\includegraphics[width=\dimexpr\linewidth-2\fboxsep-2\fboxrule\relax]{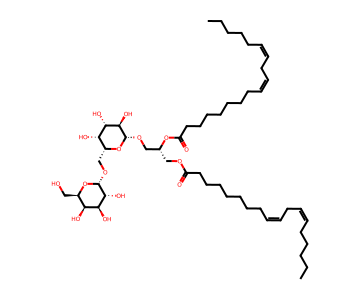}}
\end{minipage}\hfill
\begin{minipage}[t]{0.685\linewidth}
\vspace{0pt}
{\fontsize{8.0}{8.8}\selectfont\bfseries\textcolor{SAGABlue}{\#1 score=0.385 \quad T=0.24}}\\[-1pt]
{\fontsize{7.55}{8.35}\selectfont\raggedright\sloppy The molecule is a \textcolor{SAGARed}{3-\allowbreak{}[alpha-\allowbreak{}D-\allowbreak{}galactosyl-\allowbreak{}(1-\allowbreak{}\textgreater{}6)\allowbreak{}-\allowbreak{}beta-\allowbreak{}D-\allowbreak{}galactosyl]\allowbreak{}-\allowbreak{}1,\allowbreak{}2-\allowbreak{}diacyl-\allowbreak{}sn-\allowbreak{}glycerol} in which the 1-\allowbreak{} and 2-\allowbreak{}acyl groups are both specified as linoleoyl. It has a role as a Brassica napus metabolite. It derives from a linoleic acid.\par}
\end{minipage}
\end{minipage}\hfill\begin{minipage}[t]{0.485\textwidth}
\vspace{0pt}
\noindent\begin{minipage}[t]{0.285\linewidth}
\vspace{0pt}
\fcolorbox{SAGAGray}{white}{\includegraphics[width=\dimexpr\linewidth-2\fboxsep-2\fboxrule\relax]{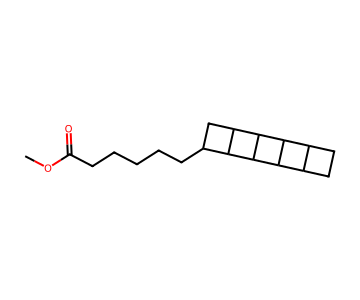}}
\end{minipage}\hfill
\begin{minipage}[t]{0.685\linewidth}
\vspace{0pt}
{\fontsize{8.0}{8.8}\selectfont\bfseries\textcolor{SAGAGreen}{\#1 score=0.306 \quad T=0.24}}\\[-1pt]
{\fontsize{7.55}{8.35}\selectfont\raggedright\sloppy The molecule is a fatty acid \textcolor{SAGARed}{methyl} \textcolor{SAGARed}{ester} consisting of methyl hexanoate,\allowbreak{} which is terminally substituted by five fused butane-\allowbreak{}rings ([5]\allowbreak{}-\allowbreak{}ladderane)\allowbreak{}. Ladderane fatty acid \textcolor{SAGARed}{methyl} \textcolor{SAGARed}{esters} are core lipids of anammox bacteria. It is a ladderane and a fatty acid \textcolor{SAGARed}{methyl} \textcolor{SAGARed}{ester.} It derives from a methyl hexanoate.\par}
\end{minipage}
\end{minipage}\par\vspace{1.2mm}
\noindent\begin{minipage}[t]{0.485\textwidth}
\vspace{0pt}
\noindent\begin{minipage}[t]{0.285\linewidth}
\vspace{0pt}
\fcolorbox{SAGAGray}{white}{\includegraphics[width=\dimexpr\linewidth-2\fboxsep-2\fboxrule\relax]{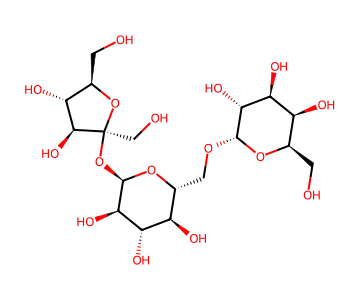}}
\end{minipage}\hfill
\begin{minipage}[t]{0.685\linewidth}
\vspace{0pt}
{\fontsize{8.0}{8.8}\selectfont\bfseries\textcolor{SAGABlue}{\#2 score=0.367 \quad T=0.15}}\\[-1pt]
{\fontsize{7.55}{8.35}\selectfont\raggedright\sloppy The molecule is a \textcolor{SAGARed}{trisaccharide} composed of alpha-\allowbreak{}D-\allowbreak{}galactopyranose,\allowbreak{} alpha-\allowbreak{}D-\allowbreak{}glucopyranose and beta-\allowbreak{}D-\allowbreak{}fructofuranose joined in sequence by 1-\allowbreak{}\textgreater{}6 and 1\textless{}-\allowbreak{}\textgreater{}2 \textcolor{SAGARed}{glycosidic} \textcolor{SAGARed}{linkages,\allowbreak{}} respectively. It has a role as a plant metabolite,\allowbreak{} a Saccharomyces cerevisiae metabolite and a mouse metabolite. It is a raffinose family oligosaccharide and a \textcolor{SAGARed}{trisaccharide.}\par}
\end{minipage}
\end{minipage}\hfill\begin{minipage}[t]{0.485\textwidth}
\vspace{0pt}
\noindent\begin{minipage}[t]{0.285\linewidth}
\vspace{0pt}
\fcolorbox{SAGAGreen}{white}{\includegraphics[width=\dimexpr\linewidth-2\fboxsep-2\fboxrule\relax]{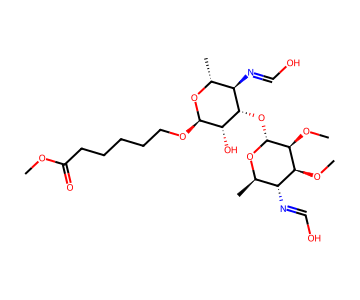}}
\end{minipage}\hfill
\begin{minipage}[t]{0.685\linewidth}
\vspace{0pt}
{\fontsize{8.0}{8.8}\selectfont\bfseries\textcolor{SAGAGreen}{\#2 score=0.299 \quad T=0.82}}\\[-1pt]
{\fontsize{7.55}{8.35}\selectfont\raggedright\sloppy The molecule is a \textcolor{SAGARed}{glycoside} that consists of an \textcolor{SAGARed}{N-\allowbreak{}formyl-\allowbreak{}2,\allowbreak{}3-\allowbreak{}di-\allowbreak{}O-\allowbreak{}methyl-\allowbreak{}alpha-\allowbreak{}D-\allowbreak{}perosamine} residue linked \textcolor{SAGARed}{(1-\allowbreak{}\textgreater{}3)\allowbreak{}} to an \textcolor{SAGARed}{N-\allowbreak{}formyl-\allowbreak{}alpha-\allowbreak{}D-\allowbreak{}perosamine} residue which is linked glycosidically to a \textcolor{SAGARed}{5-\allowbreak{}(methoxycarbonyl)\allowbreak{}pentyl} group. It is a disaccharide derivative,\allowbreak{} a \textcolor{SAGARed}{glycoside} and a \textcolor{SAGARed}{methyl} \textcolor{SAGARed}{ester.}\par}
\end{minipage}
\end{minipage}\par\vspace{1.2mm}
\noindent\begin{minipage}[t]{0.485\textwidth}
\vspace{0pt}
\noindent\begin{minipage}[t]{0.285\linewidth}
\vspace{0pt}
\fcolorbox{SAGAGray}{white}{\includegraphics[width=\dimexpr\linewidth-2\fboxsep-2\fboxrule\relax]{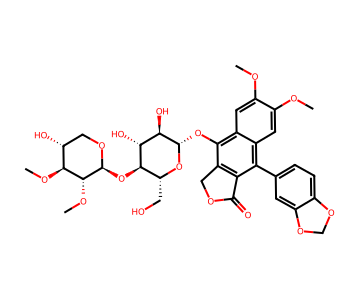}}
\end{minipage}\hfill
\begin{minipage}[t]{0.685\linewidth}
\vspace{0pt}
{\fontsize{8.0}{8.8}\selectfont\bfseries\textcolor{SAGABlue}{\#3 score=0.360 \quad T=0.14}}\\[-1pt]
{\fontsize{7.55}{8.35}\selectfont\raggedright\sloppy The molecule is a member of the class of cleistanthins that is cleistanthin A in which the hydroxy group at position 4 of the \textcolor{SAGARed}{glucoside} moiety has been converted to the corresponding \textcolor{SAGARed}{2,\allowbreak{}3-\allowbreak{}di-\allowbreak{}O-\allowbreak{}methyl-\allowbreak{}beta-\allowbreak{}D-\allowbreak{}xylopyranoside.} It is a member of cleistanthins and a disaccharide derivative. It derives from a cleistanthin B.\par}
\end{minipage}
\end{minipage}\hfill\begin{minipage}[t]{0.485\textwidth}
\vspace{0pt}
\noindent\begin{minipage}[t]{0.285\linewidth}
\vspace{0pt}
\fcolorbox{SAGAGreen}{white}{\includegraphics[width=\dimexpr\linewidth-2\fboxsep-2\fboxrule\relax]{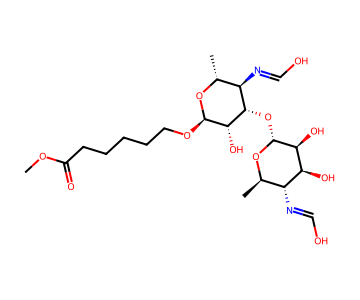}}
\end{minipage}\hfill
\begin{minipage}[t]{0.685\linewidth}
\vspace{0pt}
{\fontsize{8.0}{8.8}\selectfont\bfseries\textcolor{SAGAGreen}{\#3 score=0.272 \quad T=0.94}}\\[-1pt]
{\fontsize{7.55}{8.35}\selectfont\raggedright\sloppy The molecule is a \textcolor{SAGARed}{glycoside} that consists of two \textcolor{SAGARed}{N-\allowbreak{}formyl-\allowbreak{}alpha-\allowbreak{}D-\allowbreak{}perosamine} residues linked \textcolor{SAGARed}{(1-\allowbreak{}\textgreater{}3)\allowbreak{}} and linked at the reducing end glycosidically to a \textcolor{SAGARed}{5-\allowbreak{}(methoxycarbonyl)\allowbreak{}pentyl} group. It is a \textcolor{SAGARed}{glycoside,\allowbreak{}} a \textcolor{SAGARed}{methyl} \textcolor{SAGARed}{ester} and a disaccharide derivative. It derives from an \textcolor{SAGARed}{alpha-\allowbreak{}D-\allowbreak{}Rhap4NFo-\allowbreak{}(1-\allowbreak{}\textgreater{}3)\allowbreak{}-\allowbreak{}alpha-\allowbreak{}D-\allowbreak{}Rhap4NFo.}\par}
\end{minipage}
\end{minipage}\par\vspace{1.2mm}
\par\vspace{1.0mm}
{\fontsize{7.6}{8.3}\selectfont\textcolor{SAGARed}{red} = structural-class term shared with the query \quad
score = text-\allowbreak{}-\allowbreak{}SMILES cosine similarity \quad
T = Morgan-fingerprint Tanimoto similarity to target}
\endgroup

\caption[]{\textbf{Supplementary Fig.~\ref{fig:supp_casestudy_text_overlap_atomas_l} $|$ Description-level structural-class overlap for Atomas-L retrieval examples, continued.} Same layout and annotation scheme as the first panel; red text marks structural-class terms shared with the query description.}
\end{figure}

\end{document}